# AI Centered on Scene Fitting and Dynamic Cognitive Network

CHEN Feng[1]

**Abstract**

This paper briefly analyzes the advantages and problems of AI mainstream technology and puts forward: To achieve stronger Artificial Intelligence, the end-to-end function calculation must be changed and adopt the technology system centered on scene fitting. It also discusses the concrete scheme named Dynamic Cognitive Network model (DC Net).

Discussions : The knowledge and data in the comprehensive domain are uniformly represented by using the rich connection heterogeneous Dynamic Cognitive Network constructed by conceptualized elements; A network structure of two dimensions and multi layers is designed to achieve unified implementation of AI core processing such as combination and generalization; This paper analyzes the implementation differences of computer systems in different scenes, such as open domain, closed domain, significant probability and non-significant probability, and points out that the implementation in open domain and significant probability scene is the key of AI, and a cognitive probability model combining bidirectional conditional probability, probability passing and superposition, probability collapse is designed; An omnidirectional network matching-growth algorithm system driven by target and probability is designed to realize the integration of parsing, generating, reasoning, querying, learning and so on; The principle of cognitive network optimization is proposed, and the basic framework of Cognitive Network Learning algorithm (CNL) is designed that structure learning is the primary method and parameter learning is the auxiliary.

The logical similarity of implementation between DC Net model and human intelligence is analyzed in this paper.

## 1. Introduction

In recent years, a series of artificial intelligence technologies represented by deep learning have made significant progress. However, people recognize that the realization of stronger artificial intelligence still requires long-term efforts, and are exploring different technological routes actively.

In this article, we analyze some key issues of artificial intelligence and put forward the key points to problem-solving.

1 CHEN Feng (1973-), Male, Independent AI researcher, Research Direction: Basic Theory of AI, Natural Language Processing, Email: chenfeng@nengsike.com

We believe that nowadays data and hardware performance are no longer the shortest planks on AI, and simply piled up of data amount and hardware computing capabilities will be extremely limited for improving the level of intelligence. The key is to make a breakthrough in basic model and algorithm theory. Among them, "combinatorial generalization must be a top priority for AI to achieve human-like abilities, and that structured representations and computations are key to realizing this objective[1] ".

The core idea that we put forward and strongly recommend is: **to achieve more powerful AI, we can't just rely on end-to-end function computing with fixed input and output as the goal, but should mainly adopt the calculation, which scene-fitting centered on the basis of cognitive nature.** For this target, it is necessary to design a new cognitive structure with comprehensive domain scene description capabilities, and to perform scene fitting and realize various intelligent tasks by dynamically growing the structure.

Based on this idea, we agree that "the three concerns of classic artificial intelligence-knowledge, internal models, and reasoning need to be rethinked, and we hope to use modern technology to solve these problems in a new way[2]", and had done a lot of research work.

Obviously, building a core basic knowledge system is more important than building peripheral data knowledge, the former is the basis for explaining the latter. This kind of knowledge should be conceptualized and interpretable general knowledge, capable of omnidirectional and flexible application, rather than black box or special knowledge that is only suitable for single task processing.

The more basic question before building knowledge is how to define the internal model that carries this knowledge? This most classic AI problem must now be solved, which is the basis for other work. We believe that network-based knowledge representation is the right direction[3], but there are still many equally important principles that need to be clarified and implemented. Among them: following the basic principles of conceptualization, a unitary basic structure definition, cross-domain representation ability, the ability to flexibly combine unlimited applications, the ability to dynamically change in real time, a reasonable probability model, and the integration of learning and application are all important features.

We also agree that the knowledge representation of causality is more important than the representation of correlation[4]. But at the same time, we must see that the division of causality and correlation is relative, and causality is only one part of the many relations needed to describe the world. We believe that a more complete relation system should be constructed and used flexibly in different scenes, and follow the principle that essence relations take precedence over correlation relations[1].

Reasoning is an important ability to dynamically apply knowledge. But it should also be noted that reasoning is only a part of a complete intelligent algorithm system. This algorithm system should closely focus on the above-mentioned knowledge system to implement algorithms such as analysis, generation, reasoning, query, and learning in a unified way and integrate them.

In short, more effective knowledge is the key to intelligent systems. The new generation of intelligent systems must design a more powerful knowledge representation structure, build a more effective knowledge system, and implement an algorithm system that uses knowledge more fully and flexibly. Only when an important breakthrough is made on the issue can it be capable of interpretability, stability, and versatility, and evolve to a stronger artificial intelligence[5].

---

1 A complete relation system includes various relations such as space and causality. The distinction between essentiality and correlation is relative rather than absolute. Essential relations are more direct than correlation relations. A correlation relation can usually be decomposed into many more essential relations.

We also believe that we have not yet reached the stage where automated machine learning can be fully pursued. Automatic machine learning is a very attractive and ultimate goal, but it must be based on a powerful basic model to learn effective general knowledge. At this stage, this basic model still needs manual design to build, and then it will be able to achieve more powerful automatic machine learning than existing methods. this kind of learning must be based on structural learning, not limited to adjust parameters such as weights, and can achieve unsupervised learning and real-time incremental learning.

## 2. Basic analysis

### 2.1. Current situation analysis

The following first analyzes the main advantages and problems of several mainstream technologies, and then discusses more reasonable solutions.

#### 2.1.1. Symbolicism method

The correctness and advantages of the symbolicism method are mainly reflected in:
- The principle of using refined and summarized conceptual symbols to describe knowledge is undoubtedly correct and conforms to the nature of human cognitive thinking and knowledge processing. Conceptual knowledge has the characteristics of being interpretable, easy to maintain, and flexible in use. It can fully reflect the value of knowledge. The large amount of knowledge accumulated by humans is mainly expressed and passed on through highly conceptual symbols.

But for a long time, symbolicism method has made slow progress in realizing strong artificial intelligence, reflecting its problems to be improved:
- The description ability of the of the existing symbol system' basic structure is inadequate. The representative knowledge graph reflects the correct direction of expressing knowledge with a network structure, but the structure is still too simple, not complete and flexible, and it is difficult to realize a more complete cognitive system.
- The existing knowledge system usually more focus on definition and lacks explanation. Emphasis on the management of large amounts of data knowledge, lack of in-depth representation of basic knowledge and common sense, unable to support multi-level intelligent computing.
- The design and representation of many knowledge systems are too simplified and solidified, losing the dynamic and flexibility that intelligent computing must have.
- Lack of a reasonable probability system. The design of various symbol systems prefer to handle deterministic knowledge, and the system of describing and dealing uncertain knowledge is not constructed.
- The basic concepts of classification, knowledge, data, instance, rule, generalization, reasoning, application, and learning, as well as their interrelations, adopt different representations and implementation methods, which cannot be summarized under one system for processing.
- Lack of a general intelligent computing model closely integrated with the symbolic

knowledge system.

### 2.1.2. Connectionism method

The typical representative method of connectionism[1], deep learning has achieved remarkable results in some fields in recent years, mainly reflecting the following correct directions:

- The description ability of the network structure is powerful, and knowledge representation needs to be realized by a network with sufficient connection density.
- The use of continuous values such as weights to realize the description and calculation of uncertainty proves the importance of uncertainty processing for intelligent systems.
- Demonstrates the great value and potential of automatic machine learning.

Similarly, the existing connectionism and deep learning technology also have many problems. There are obvious defects in aspects of interpretability, stability, and flexibility. The key reasons are as follows:

- The non-conceptualization of internal network elements of connectionism is the fundamental reason for its lack of interpretability. The knowledge contained in this kind of network is embodied as a highly specialized structure and parameters for fitting specific input and output. It does not have the general description ability that conceptual knowledge should have, and it is difficult to realize the flexible use of knowledge-such as the advanced form of organization Inference calculation.
- Because the elements are not conceptualized, the weight of pragmatism is used to handle with uncertainty, which cannot reflect the interpretability and independent adjustment ability of a true probability system.
- Only rely on an over-simplified structure to describe complex knowledge, such as relying on only one M-P structure. Realizing basic relationships like xor requires a complex multi-level structure to force simulation and can not completely solve the problem. A more reasonable solution is to introduce a special xor relationship to solve the problem. The source of the connectionism theoretical, the human brain does just this. Of course, this means that the heterogeneous network with multiple relationship types will replace the single structure network.
- Pursuing end-to-end superficial fitting and extremely simplifying the design of the intermediate model, which is suitable for unidirectional function conversion processing, but cannot achieve flexible intelligent processing that requires omnidirectional calculation. For different tasks and even refined subtasks, different networks need to be trained. Therefore, many types of networks are derived but it is difficult to flexibly integrate together to solve complex problems faced by real scenes.
- The machine learning implemented on the above basis is difficult to learn structural

---

[1] The connectionism methods, deep neural networks, and deep learning referred to in this article usually have these characteristics: use non-conceptual elements to organize into an internal multi-level network; do not define precise types and semantics for each element, and are mainly reflected by parameters such as weights to show the differences between each other; facing end-to-end fitting and calculation for fixed input and output; using a single structural model such as M-P model to express and calculate; training and learning methods are mainly to adjust the weight parameters.

knowledge, and mainly uses the adjustment of weights and other parameters to simulate the structure (so usually only the differentiable knowledge can be learned). In fact, structure is far more important than parameters for cognitive representation, and cognitive learning must take structural learning as the basic goal.

### 2.1.3. Probabilistic graph

Probabilistic graph models combine probability theory and graph theory, apply them to specific systems to solve uncertainty problems, and uphold the idea of using causality to replace correlation. These theories and methods are very correct and important, and they have produced tremendous value in many fields. However, it is not enough to achieve strong AI. The key problem is that the various theories of existing probability graphs are mainly for closed scenes and systems, and cannot be applied to the complexity of open domains that include multiple relation types and dynamic structures. It requires a completely new design of the basic assumptions and description calculation system of the probability model in such domains and scenes.

## 2.2. Target System

Based on the analysis of the above-mentioned key issues, we propose a solution to achieve stronger Artificial Intelligence.

The realization of a Strong AI system requires a lot of work in many fields, but in aspects of basic software structure and algorithms, the target system should be composed of "intelligent system + external system". Among them, the intelligent system is at the core and realizes general intelligent processing for the open domain; the external system is used as a professional tool to assist in the realization of various closed domains, and is called and directed by the intelligent system.

Obviously, the realization of the intelligent system is a difficult point. As mentioned earlier, it should adopt calculations centered on scene fitting.

Most of the traditional systems are unidirectional function calculations. The characteristics of this calculation are:
- Calculate the output of the other end according to a function mapping for the input at one end. For example: in the injective function $f(x,y)=x+y=>z$, x and y are known and input, z is unknown and output, the calculation process is to calculate z for x, y, x,y is unidirectional influence to z, z does not have a reverse influence to x and y. The target of calculation is only the output x rather than the entire system. The system is just a fixed tool to complete the calculation.
- Input and output are fixed, and will not change during system design and calculation.
- The intermediate data and network structure are designed for the unidirectional mapping calculation of this function, and cannot support reverse calculation and describe the entire system.

Most of the existing neural network systems are unidirectional function calculations. The internal network itself is a unidirectional static conversion structure. Data is flowed from the network and output after calculation. The network itself does not dynamically build and fit the scene.

The unidirectional function computing performs well in implementation tool computing tasks, supporting the rapid development of information technology for decades.

From the appearance point of view, it seems that the human brain is also a unidirectional function calculation-the information transmission of each nerve cell is basically unidirectional. The M-P model is designed on this theoretical basis, and the derived multi-layer neural network system and other related technologies have achieved remarkable results in some fields. This creates an illusion: it seems that the unidirectional flow of data can produce intelligence.

But it is not the case, unidirectional function computing lacks comprehensiveness and flexibility, and it difficult to achieve more powerful intelligent computing.

Let's review the calculation of the equation. It is more flexible than the function calculation. If the function we mentioned above is converted to the equation x+y=z, then we can see:
- The equation itself does not distinguish between independent variables and dependent variables. All variables are equal. Therefore, the known, unknown and the calculation direction are not limited. In the specific calculation, we can calculate z based on x and y, or calculate y based on x and z, or calculate x based on y, z.
- In essence, a function formula is one of multiple variations of an equation. The equation is a description, and the function is a calculation. **"Description is more important than calculation, and calculation serves for description."**

Calculation centered on scene fitting is a systematic structural calculation, which has the essence and flexibility of equation calculation, and has more powerful description and calculation capabilities:
- Treat the structure composed of all variables as a complete system, describing not only the relation between variables but also the relation between variables and the system. It can be understood as: the structure of the equation itself is only a tool for describing the relation between variables, and the structure of the scene fitting calculation itself is also a target variable (root variable).
- **The target of calculation is the entire system!** The calculation is not limited to directly calculating unknown variables based on known variables, but more importantly, calculating the entire system based on known variables and calculating unknown variables based on the entire system. For example, a scene fitting structure of 4 elements can be defined as S(x,y,z), where x, y, z are directly related to S, respectively. S and other variables can be solved based on many or only one variable, and the feasibility of all calculations and result evaluation are determined entirely based on probability.
- **The calculation method is to select different structures and dynamically combine them to build a system that fits the scene.**

In order to realize this kind of scene fitting ability, it is necessary to design a newly model, namely the Dynamic Cognitive Network model.

We believe that the essence of intelligence is scene fitting (a scene is a fragment of the world), that is cognition.

**Cognition**: The cognition of the human brain is to fit the target scene with the logical structure represented by the specific state of a group of neurons; the cognition of the computer is to build a data structure to simulate the logical structure of the human brain to fit the target scenes. This logical structure can be called cognitive structure, which is embodied as a kind of network structure, so it is also called cognitive network structure. This structure is a comprehensive and essential description structure (similar to an equation) rather than a unidirectional conversion calculation structure (similar to a function formula). It is independent of calculation and beyond calculation. Various calculations and input and output are all based on different uses on same cognitive structure.

The calculation centered on scene fitting aims at constructing a cognitive network that fully fits the target scene, in which the input and output are also the cognitive network components that constitute the target scene. Therefore, the calculation from input to output is always divided into three steps:
- Obtain input information, that is known components of the target cognitive network;
- Calculate and dynamically build a complete cognitive network for scene fitting based on input information;
- Extract the required informational component output from the constructed cognitive network;

A simple example: a cognitive network is constructed for "cat" [cat: cat has a head, cat has 4 cat paws, cat has a tail...]. This is an essential description network, itself has no fixed input and output, and it does not make meaning to directly ask "What is the input and output of the cat?". Only in the specific calculations, one part of the information is set as the known and the other part as the unknown for calculation and solution: at a moment when the head of the cat is seen from the front, the head is used as input to calculate and construct the essential cognitive structure of the entire cat. After the construction of this structure is completed, not only recognition and classification tasks can be realized, but also many tasks such as invisible paws and tails can be inferred; at another moment, the cat's tail may be seen from the back, and the tail is used as input to calculate and construct the same structure and achieve various targets and tasks equally.

Compared with the end-to-end "direct" calculation, this "indirect" calculation that always goes through the scene construction process is the final solution to solve the essential problem, which can achieve stronger AI, especially in the following aspects:

1. **Comprehensiveness**: On the basis of complete scene fitting, cognitive representations and different computing tasks in various domains can be realized uniformly. For example, for image recognition, after completing the construction of the spatial scene, different tasks such as image classification, semantic segmentation, and entity segmentation can be realized at the same time, it needn't to be decomposed into different systems to implement them separately. Moreover, comprehensive scene fitting is not limited to things like cat and spatial scenes. It can also construct cognitive information in various domains such as space, time, causality, thinking, and communication as a cognitive network with the same structure to fit a comprehensive scene. This kind of fitting includes restoring the events and states that have occurred in the scene, as well as calculating the past of the scene and predicting or even manipulating the future. In this way, it is no longer limited to only performing a single preset task, but has comprehensive intelligent computing capabilities.

2. **Flexibility**: Complex scene can be composed of a combination of scenes in different domains. The specific implementation is to select cognitive network modules on demand to dynamically combine them into a new cognitive network. This combination is carried out at runtime, ac-cording to the needs of the task and the calculation process to adjust in real time to reflect the powerful integration capability. Models and systems with fixed structures are difficult to achieve this capability.

3. **Continuity**: Centering around a cognitive network structure that is independent of computing and persists, it can realize the continuous execution of multiple tasks.

4. **Interpretability**: The formed cognitive network structure is completely interpretable, even the process of intermediate calculation and reasoning can be described and explained.

5. **Essentiality**: This calculation is more conform the nature of human intelligence. We believe that **the main working mode of the human brain is scene fitting!** Although the unidirectional information flow between nerve cells observed locally look like unidirectional functional calculations, **from a macro perspective, the overall effect of the combination of many unidirectional information flows (combined with xor and other mechanisms) is to construct the correct scene fitting structure!** By contrast, the M-P model only describes a part of the whole process rather than the whole picture.

Here is an example to illustrate that the calculation centered on scene fitting has the ability to integrate information and tasks across domains, and these capabilities are unified by sharing the same set of cognitive knowledge and algorithms.

Suppose Tom and Jerry met, and the following conversation occurred:

*Tom said, "Hello."*

*Jerry asked: "Hello, are you French?"*

*Tom replied: "No, I am an American."*

*Jerry said: "Great, so am I."*

The top level of this example is a [Dialogue] scene, which contains different types and sizes of concepts and scenes such as [Causality], [Ask], [Answer], [Country], [Person], and their external forms(For example, video, sound, language) are also part of the scene and a component of the cognitive network.

However, the known information that can be obtained at the beginning of each task is only a part of the complete scene, and it is a different part in each task. The core of the task always calculates and completes the **same** complete scene based on these **different** known information.

Firstly, let's look at an understanding task. Assume that the robot as a third party see and understand the entire process of the conversation between Tom and Jerry. It can directly obtain information such as images and sounds. The task is to complete the entire scene based on the known partial information. The specific process involves: completing entities such as Tom and Jerry based on image recognition; completing [Tom speaks to Jerry] and other actions based on video recognition; obtain the spoken text through speech recognition and perform semantic understanding to obtain the content of the text; and construct the causality of the whole process..., Once the entire scene can be correctly completed, various specific outputs become very simple: For example, for asking "What is Jerry doing?" and "Which country is Tom from?", we can use questions to match information in the constructed scene and extract the correct answer; we can also make further reasoning, such as asking "Which country is Jerry?". At this time, we need to expand the scene, because reasoning is also a composition structure of the scene. We can add reasoning components to expand the scene until include the information "Jerry is an American" and extract the output.

Secondly, let's look at another dialogue task, let the robot play the role of Tom for dialogue task. In this different task, only the known and unknown settings have changed. The entire scene has not changed, the cognitive knowledge it is based on has not changed, the core principle of completing the entire complete scene based on partial information will not change.

To analyze a piece of information more specifically, suppose there is a piece of causality knowledge [[A asks B question] cause [B answers A answer]] in the cognitive knowledge base. According to this piece of knowledge, [Jerry asks: "Hello, Are you French?"] and [Tom answers: "No, I am an American."] these two action scenes are combined as a larger scene [[Jerry asks: "Hello, you Is it French?"]] cause [Jerry answers: "No, I am American."]] through middle causality relation. When the robot as a third party understands the dialogue, both actions are known, and only the unknown part [cause] in the middle needs to be made up; while the robot plays Tom for the dialogue task, according to [Jerry asks: "Hello, are you French?"] the action that has already taken place to calculate Tom's reaction, what still needs to be done is still based on the known completion scene. The only difference is: only [Ask] is known, so it can only be completed based on this known item to complement [cause] and [Answer] the two unknown parts.

The point here is: each task and the role of the observation are different, so the input and output are different, **but the calculation is based on the same knowledge, the constructed scene is same.** For the above understanding task, there are known two components [ask] and [answer], so the result will be relatively single and with higher probability, similar to solving S and z based on x and y. While for the dialogue task described later, it is known that there is only one component [ask]. There are many results and the probability of each result is low. It is similar to solving S and y, z according to x. It can even be set [Tom's answer: "No, I am the United States People."] as known to solve the unknown actions and relations in the front, at this time, based on the same knowledge to calculate [Jerry asks Tom: "Are you from X country?"] the result has a relatively high probability.

Thirdly, let's look at another translation task. If Tom and Jerry speak different languages and the robot is required to intervene to provide translation for both parties, in this task the robot will still follow the same knowledge and algorithm to fit the same scene.

In short, because the scene contains all the information, various tasks such as article classification, sentiment analysis, semantic search, information extraction, article summarization, and article generation can be implemented around the same scene fitting without being broken down into different tasks. It can't be broken down! Because all information is entangled with each other.

In addition, the following points should be noted from the above examples:
- Dynamic structure and static structure are processed in the same way. infer the information that has not yet occurred (for example, Jerry answered Tom's question above) and infer the information that has occurred but cannot be directly obtained (for example, see the cat's head and then infer the invisible tail), there is no essential difference.
- In the causal structure, there is no essential difference between the backward calculation and the forward calculation along the time.
- Subjective information is objectified: there is no essential difference between the pro-

cessing of human thinking and communication and the processing of objective information.
- There is no essential difference between action and understanding. Performing actions by yourself and understanding the actions of others follow the same knowledge.
- **In brief, in the system centered scene fitting, there are no essential differences between various computing tasks, and the differences are mainly reflected in basic aspects such as probability!**
- Scene fitting is oriented to comprehensive domains and not limited to specific domains such as natural language. In the real scene, the words spoken by Tom and Jerry are processed with natural language, and the speak actions are processed with videos. In this paper, since the video cannot be inserted, the speak actions represented and processed with natural language too. Either way, the goal of the construction is the same scene and internal cognitive structure, which can simultaneously contain different external components such as images, videos and natural language.

The principles of design and implementation to intelligent system centered on scene fitting are as follows:

### 2.2.1. Uniform structure definition

The basis of an intelligent system is a network structure composed of conceptualization elements combined with built-in probability models.

**Conceptualization elements:** Symbols are used to describe and calculate information. The nodes and connections of the hidden layer of neural networks are also essentially symbols. The essential difference between symbolism and connectionism is reflected in the different principles for the definition and realization of symbols. Specifically, conceptualized and non-conceptualized symbols are used respectively.

Conceptualized symbols have been optimized by generalize, regularization, and orthogonalization. Non-conceptualized symbols are an representation of vague, redundant, and non-orthogonalized.

To give an example to metaphor, as shown in Fig. 1, to represent the information of a point on a two-dimensional plane, using two orthogonal dimensions X and Y to represent can be regarded as a conceptual symbolic represent; while using the multiple dimensions $V_1 \sim V_n$ without optimization and orthogonal processing to represent is a non-conceptual symbolic represent.

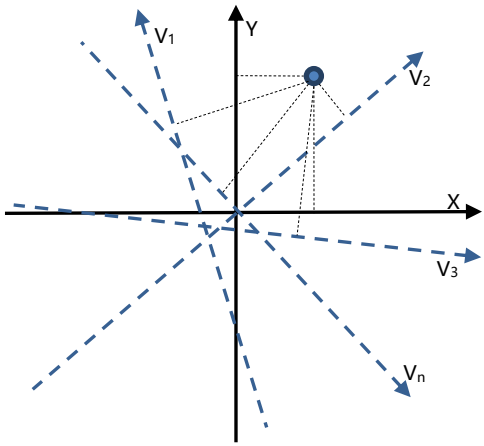

X,Y: Conceptualized symbols

$V_1$~$V_n$: Non-conceptualized symbols

Fig. 1: The Contrast Diagram of Conceptualized and Non-conceptualized symbols

Obviously, the conceptualized system has a clearer and more efficient description of the goal. Even if the non-conceptualized system can accomplish tasks equivalent to the former, it will be much worse of flexibility and performance. There is also a key interpretable issue: human cognitive thinking and language are highly conceptualized, for ambiguous and vague information, it is difficult to make the name and explanation which humans can understand. If X and Y correspond to a person's height and weight, they can be given language names such as "height" and "weight" for representation and communication. $V_1$~$V_n$ all are mixed with height and weight information and cannot be meaningfully named. There is no explanation and communication about them.

Therefore, the conceptualized symbol system is the goal pursued by the system. Even if it is possible to use non-conceptual symbols to initialize the construction system and automatically learn and train as a method, the ultimate goal should also evolve to the conceptual system to achieve better optimization (a key point of machine learning is also to use compression, tailoring and other networks Simplify processing to try to achieve similar goals). If a machine learning method is good enough, the result of automatic learning and optimization has the same capabilities as the conceptual system and has almost the same performance, there is reason to believe that its internal is already a conceptualized system[6].

Now the question is: how to build a conceptualized structural system? Obviously, if automatic machine learning can do it, of course it will be done by the machine. If it can't, it can only be done by manual design. We believe that the automatic machine learning technology at this stage does not have the ability to learn to obtain a conceptual model that can comprehensively describe the world. Therefore, this work needs more manual methods to achieve. A suitable technical framework can combine the two: Manual construction is main and machine learning is assist, and both share the same knowledge base. Machine learning does a large amount of data correlation analysis to construct the prototype knowledge to be determined; manual construction sets a priori constraints for machine learning and confirms and corrects the learned knowledge.

Therefore, it different from the neural network which is mainly constructed by non- conceptualized symbols and adjusted by weight parameters, the new intelligent system should be a conceptualized symbol system, and each node and relation is more accurate, efficient and interpretable conceptualized symbol.

*(For the human brain, perhaps each neuron at the micro level does not directly correspond to a conceptualized symbol, but there is reason to believe that the macroscopic effect formed by these microstructures is conceptualized. The basic principles of computers are very different from the human brain Therefore, it is not necessary to completely simulate the microscopic details of the*

*human brain. It is more efficient to directly simulate the macroscopic effects of its conceptualization. Therefore, the model proposed in this article is not called neural network, but cognitive network.)*

**Network structure:** The network structure has flexible and powerful information description capabilities. Models and systems that hold different ideas such as symbolicism, connectionism, probability graphs, graph neural networks, etc., are all embodied in a certain network structure.

The basic structure of the new intelligent system should be a multi-level heterogeneous network structure composed of conceptual nodes and relations, with sufficient connection density and structural complexity (measured by this standard, the structure of the existing knowledge graph is too Simple) in order to realize the in-depth, complete and flexible representation of rich knowledge. Among them: the heterogeneous structure requires that each element of the network can have different types and semantics. This multi-level structure is a dynamic structure that is finely organized on-demand, and is significantly different from a fixed multi-level structure.

**Probability model:** Human cognition of the world is closely combined with the description and calculation of uncertainty. Dealing with uncertainty is the basic requirement of an AI system, and it can be used as one of the key features to distinguish between AI systems and non-AI systems. The intelligent system should assume that the uncertainty probability (0~1) is the basic property of most information, and the certainty probability (0 or 1) is only a specific value of the uncertainty probability.

Looking back at the existing systems, failure to incorporate a reasonable probability model is one of the important reasons why many systems are difficult to achieve powerful intelligence. Systems such as symbolism, connectionism, and probability graphs have all proved the importance of uncertainty calculation in the AI field and have different implementations, but these implementation methods still have problems that need to be improved.

Some systems (such as prolog language, many knowledge base systems) selectively only represent and process deterministic knowledge, while ignoring or forcing the removal of uncertainty. It makes knowledge representation and calculation lose the universality and practicability and has a large initial deviation and accumulated deviation, resulting in unsolvable problems of knowledge conflict[1] and lack of scalability.

Other systems use qualitative quantifiers such as [all] and [exist] to qualitatively represent the uncertainty of knowledge, which should be unified into quantitative probabilities for description and calculation.

As in the previous analysis, the internal structure of most neural networks is non-conceptual, so weights and other mechanisms are used instead of a probability system to solve the uncertainty problem. The neural network system does not have the complete description ability of the conceptualized probability system for cognitive representation and calculation. This is also the reason why it is difficult to explain and difficult to accurately debug and adjust.

The probabilistic graph model usually is not combined with the semantically rich heterogeneous cognitive structure. The basic design of the probabilistic model is to face a closed domain and a static network system with a simple structure type. It is not suitable for the dynamic cognition

---

1 The essential of knowledge conflict is transforming uncertain knowledge into certainty and restoring the uncertainty of each knowledge is the basis for solving conflict problems.

intelligent system that face an open domain and complex type structure.

The new intelligent system must solve the above problems and design a probability model which matches the cognitive network and suits for cognitive representation and calculation.

**Unified structure**: This basic structure has high unity. It is not only a description structure, but also a calculation structure and a memory structure.

### 2.2.2. Building a cognitive knowledge base

On the basis of the aforementioned unified structure model, to build a cognitive knowledge for different domains, the following goals should be achieved:

**Full domain description**: Cognitive knowledge in different domains (for example: things, events, actions, attributes, relations, causality, thinking, forms, images, video, natural language, and even computational processes) are represented in the aforementioned unified structure. They can be integrated into one to achieve a comprehensive fitting of complex worlds and scenes.

**Essential description**: Cognitive knowledge should describe the essential structure of the target scene, reducing the need to construct and train isolated knowledge for different tasks.

**Data-based representation**: All kinds of knowledge should be represented in Dataization form as far as possible to reduce hard-coded and other fixed forms of representation.

**Homogeneous of knowledge and data**: The same representation and calculation are used for knowledge and data, and the difference is only reflected in the level of abstraction.

**Comprehensive representation and select to use**: Try to build a more comprehensive cross-domain knowledge to realize cognitive description, and select the most effective part for use in specific scenes according to actual needs. This is an important feature and advantage of human intelligence, and it is also the basic principle of an open domain-oriented artificial intelligence system. It can fundamentally solve the problems of independent and identical distribution assumptions and models with fixed parameter structures.

### 2.2.3. Dynamic Cognitive Network algorithm system

Realizing a Dynamic Cognitive Network algorithm system based on the aforementioned model structure should have the following characteristics:

**Dynamic network growth**: The processing of various algorithms is always organized around the cognitive network, and **the calculation process and results are reflected in the same cognitive network**. This cognitive network is created, grown, and changed with different algorithm processing, and exists in the whole process of the entire task or even longer.

**Algorithm fusion**: fusion of various parsing, classification, generation, query, reasoning, and even learning algorithms into one, they are all reflected in the same dynamic network growth and share the same set of cognitive knowledge bases, various algorithms can be dynamically combined in real time, And can realize the incremental continuous calculation of multi domains integration.

**Structure-oriented machine learning:** The new knowledge obtained by machine learning must have the same structure as the existing knowledge, and follow the principle that the learning structure is better than the learning parameters. The structure is qualitative and the parameters are quantitative. The learning algorithm can first learn the new knowledge Know the network structure, and on this basis, adjust and optimize the parameters of the structure as needed.

Learning structure is more difficult than learning parameters. Many machine learning is to adjust the parameters after a certain fixed structure is preset. This is effective for a single task that this preset structure can meet. For more complex intelligent computing tasks, it is impossible to adjust parameters to solve the fundamental problem of insufficient structure description ability.

There are two main difficulties in learning structure: one is the definition of the target structure of learning. If the target structure is not clear, it is impossible to learn. The another one is the differentiability of knowledge. Nowadays mainstream machine learning methods usually require knowledge to be differentiable.

The new intelligent system must solve the two problems fundamentally: First, the core idea of the intelligent system is to build a unified conceptual cognitive structure, which defines the specific target structure for machine learning. Secondly, the essence of structure learning is learning the knowledge of discretization. Discrete value is more basic than continuous value and the two can be merged. The intelligent system will combine discretized structure learning and continuous value parameter learning to form complete cognitive learning.

## 3. Dynamic Cognitive Network

After analyzing the basic principle of the target intelligent system, we worked out a concrete implementation scheme.

The basic model of this implementation scheme is called "Dynamic Cognitive Network" or "Structure Network", or "DC Net" or "S Net" for short. It faces open domains. And it conceptualizes all kinds of cognitive knowledge into cognitive network elements and integrates these elements with the cognitive probabilistic model based on bidirectional conditional probabilities to form the cognitive network structure of two dimensions multiple levels to describe and fit the scene. It does various calculations by taking the omnidirectional network matching—growth algorithm system driven by task targets and probabilities.

The following provides details on the model of DC Net in terms of structural model, probabilistic model and algorithm system.

### 3.1. Structural Model

**Basic elements:** DC Net, as a network structure, is also composed of nodes and edges, and a DC Net can be regarded as a triple:

S=(C,R,P)

C: Concepts set, each concept is a node of the network.

R:　Relations set, each relation is an edge of the network.

P: Parameters set, including global parameters and private parameters of concepts and relations.

Concepts and relations are collectively called elements.

#### 3.1.1. Concept

**Concept:** DC Net is a multilevel cognitive model and it is used to representes the cognition of the world through successive generalization and combination of concepts, which can be summarized as "Everything is a concept".

The definition and description of concept have the features of comprehensiveness and fundamentality.

Comprehensiveness: It does conceptual representation of the cognition in comprehensive domains. It brings various basic concepts, parameter concepts, entity concepts, relation concepts, event concepts, causality concepts and scene concepts into the same concept system, and fuses them to realize comprehensive cognitive representation and calculation.

Fundamentality: It gives priority to the construction of the most basic concepts (such as quantity, degree, collection, space, time, existence, comparison, relation, derivation, etc.), and use them to construct the core semantics and computing systems to support the realization of various specific concepts in comprehensive domains.

How to build a detailed system of the concept and relation is a special topic which is not the focus of this paper.

### 3.1.2. Relation

**Relation**: The relation is a special concept. A relation can connect two concepts or two relations. The two elements are called the A end and the B end of the relation respectively.

In DC Net, relations are bidirectional and a relation can be regarded as the combination of two connections in opposite directions: One is from A to B which is called A=> B connection; The other one is from B to A which is called B=> A connection.

Relations are more basic than concepts. The semantics and computing rules of the concept and network are mainly carried by relations. A concept alone has no meaning and each concept interprets itself by means of the relation with other concepts. Concepts interpret each other.

**Parameters of relation:** The relation has different parameters. For example, the [has component relation] and [adjacent relation] which represent the spatial structure has the parameters [angle], [distance] and so on.

**Relational deriving:** One of the important design in DC Net is that the relation is also a multilevel derived system same as the concept. The relation and the related concepts derive synchronously together and realize the derivation of the whole network.

**Basic relation:** The relation at the top of the derived system is called basic relation, and different basic relations have different semantics, parameters and computing rules.

The following lists some of the most important basic relations to illustrate and they are classified as the set-dimension relation and the domain-dimension relation.

#### 1. Set-dimension Relation

The set-dimension relation representes the set relation of the concepts with the same type and the semantic inheritance relation of derived concepts to base concepts.

**Belong to relation:** The belong to relation is represented as [A belongs to B] or $A \subseteq B$. A and B are the two ends of the relation which can be two concepts or two relations. A is called derived concept (or derived relation) and B is called base concept (or base relation).

The conditional probability of the belong to relation meets $P(B|A)=1$, $0<P(A|B)<=1$, which quantitatively expresses the semantics of "all derived concepts are base concepts and partial base concepts are derived concepts" with uniform conditional probabilities.

The belong to relation is of transitivity, that is, if $A \subseteq B$ and $B \subseteq C$, then $A \subseteq C$.

A concept can have multiple basis concepts which can be dynamically added and removed.

**Equivalence relation:** The equivalence relation is represented as [A equal B] or A=B.

The equivalence relation is a particular case of belong to relations and its conditional probabil-ity meets P(A|B)=1, P(B|A)=1. The belong to relations mentioned in this paper all include equivalence relations.

The equivalence relation is also of transitivity and symmetry as well.

**The belong to and equivalence of the value concept:** The continuous value is discretized into value concepts, And the set relation between value concepts is represented by the belong to relation and equivalence relation.

例如： 25 ⊆ (20,30) = (20,30) ⊆ (10,50) ⊆ (0,100) = (0,100)

e.g., 25 ⊆ (20,30) = (20,30) ⊆ (10,50) ⊆ (0,100) = (0,100)

The belong to relation and equivalence relation of the value concept are reflected by real-time calculation, and they don't need formal representation.

The relations and derived system in the set dimension are very important which is the basis to support the whole cognitive description and computing system. The derived system has the following characteristics:

**Derived overloading:** The derived concept(derived relation) inherit all the semantics and parameters of the base concept (base relation) and can overload them. The new values overloaded cannot deviate from the scope defined by the base concept and should be more concrete values.

**The normalization of concepts and instances:** One very important idea of DC Net is that it does not distinguish the basic definitions of concepts and instances. In essence, describing all concepts is for the purpose of mapping specific denotations, and the instance is also a concept -whose denotational amount from the concept to the mapping is equal to 1.

To emphasize this, this paper analyzes such a derived chain: sun⊆star⊆heavenly body ⊆ thing⋯. Among them, the star, heavenly body and thing are obviously classes, because the denotational amount they map is greater than 1. However, there are different schemes for whether the sun should be defined as an object or a class: In tradition, the sun, which is different from the other three, is defined as an instance of objects instead of a class with different implementation methods for the structure and algorithm; However, in DC Net, they are the same and they are all concepts except that the denotational amount of sun is 1. But that the denotational amount is 1 itself is difficult to guarantee the information eternal correct, so it will cause a lot of problems to divide concepts absolutely based on such information.

Therefore, everything is a concept in DC Net which does not distinguish the basic types of knowledge, data, classes, objects, instances, etc.. And it implements the generalization relation between concepts(including the relations between knowledge and data, basis classes and derived classes, and classes and instances of objects) with the same belong to relation. There is no need for additional instancing relation. Adopting the terms facing objects can be called the unification of objects and classes—"An instance is a concept and a object is a class", which is critical to achieving a highly consistent cognitive model.

**The formalization of variable bindings:** Variable bindings (variable assignments) can also be formalized as needed, for example, the variable binding operation of "Tom. Age=15" can be represented as the relation of "Tom.age equal 15". The unidirectional procedural operation can be converted to a two-way relation and we can also describe and track the causal scenes of time, versions and other processes, for example, [[Tom.age equal 15] has time 2010] and [[Tom.age equal 16] has time 2011] are used to describe two relations.

## 2. Domain-dimension Relation

The semantics of the domain-dimension relation is rich and can be extended according to needs. The following is a brief description of some typical basic relations of domain dimension, which can be divided into two categories the longitudinal relation and lateral relation.

- **Longitudinal Relation**

The concepts of the two ends of a longitudinal relation has different levels. Typical longitudinal relations are as follows:

**Owning components relation:** It is represented as [A has component B] in which A is a high-level concept and B is a low-level concept, representing that A is composed of B.

The conditional probability of owning components relation meets $0<P(A|B)<=1$, $0< P(B|A)<=1$.

e.g., [person has face], [face has eye], [face has nose], [ask has asker ] and [ask has listener].

**Owning parts relation:** It is represented as [A has part B] which is a particular case of owning components relation. A and B are the concepts of same type. The granularity of A is greater than that of B, which reflects the decomposition and combination of different granularities of concepts.

The owning parts relation can represent the concept with different precision. According to the needs, it can roughly represent the high-level concept with large granularity, or represent the subordinate low-level concept with small granularity more precisely.

e.g., [curve has part curve] [surface has part surface] [time has part time]

**Owning attributes relation:** It is represented as [A has attribute B] in which A is a high-level concept and B is a low-level concept called attribute.

Owning attributes relation is similar to owning components relation except that the concept of attributes embodied as a feature parameter is more basic than the concept of components.

e.g., [eye has color], [person has number], [face has angle], [ask has time], [[face has eye relation] has angle], [[adjoining relation] has distance]

**Owning forms relation:** It is represented as [A has form B] in which A is a high-level concept and B is a low-level concept called formal concept.

In DC Net, the form is also a concept which has no essential difference from the normal concept except that it is at a relatively peripheral and low-level position in the whole cognitive description system, which can usually be directly perceived and measured by senses or sensors. The definition of the formal concept is related to specific domains, such as: language forms (strings and language roles) in natural language processingand image forms in the image processing domain.

e.g., [concept has language form] , [concept has image form].

**Owning contents relation:** It is represented as [A has content B] in which A is a high-level container concept and B is a low-level content concept.

e.g., [think has content] , [represent has content].

- **Lateral Relation**

There is no obvious difference of levels between the two ends of the lateral relation. Typical lateral relations are as follows:

**Adjoining relation:** It is represented as [A adjoin B ] which describes the positional relation between the equal-level concepts A and B.

Adjoining relations have the parameters of [angle], [distance] and so on.

e.g., [eye adjoin nose] and [nose adjoin mouth].

**Comparison relation:** The comparison relation which representes quantification among equal-level concepts can be regarded as a special adjoining relation.

The comparison relation has the parameter of [comparison difference ].

e.g., [5>3] and [2001 earlier than 2002].

**Conversion relation:** It describes the conversion of two tree networks and is a tree-network relation. The elements of the tree-network relation include not only the relation itself and the nodes at both ends, but also the tree network structure expanded by the nodes and the additional relation in the middle.

e.g., [addition convert subtraction], [buy convert sell],[question convert answer].

**Causality relation:** It describes the conversion relations of events in chronological order.

The conditional probability of the causality relation meets $0<P(A|B)<=1$, $0< P(B|A)<=1$,and both directions can be calculated by direct reasoning.

e.g., [takeoff cause fly], [like cause buy] and [ask cause answer].

**Changing relation:** It representes the causality relation of the state of the same concept changing over time and both ends are regarded as different versions of the same concept.

**Moving relation:** It representes the causality relation of the position of the same concept changing over time and both ends are regarded as different versions of the same concept.

**Equivalence relation:** Equivalence relation is both the set-dimension relation and the domain-dimension relation. In the domain dimension, it is reflected as a lateral relation. The two ends can be the same or different versions of the same concept.

**Mutually exclusive relation:** It is represented as [A xor B] which represents a mutually exclusive relation between two elements.

The conditional probability of the mutually exclusive relation meets $P(A|B)=0$, $P(B|A)=0$.

### 3.1.3. Two dimension multi level structure(TDML)

In DC Net, concepts and relations are connected into the cognitive network , which is represented as the two-dimension multilevel structure (TDML) in the set dimension and domain dimension. It representes and calculates the abstractness of cognitive model by means of **derived network** structure in the set dimension, which embodies the ability of **generalization**; it representes and calculates various cognitive models in various domains with **tree network** structure in the domain dimension, which embodies the ability of **combination**. The two multilevel structures are described independently and closely integrated in terms of semantic meaning and computing rules.

**1. Derived Network Structure**

The derived network structure is a multilevel structure in the set dimension.

Definition: For two networks $S_d$ and $S_b$, if there is a sub-network $S_d'$ in $S_d$ which meets the topological isomorphism of $S_d'$ and $S_b$, and meanwhile every element $E_{bi}$ of in $S_b$ and the element $E_{di}$ corresponding to the topology in $S_d'$ meets $E_{di} \subseteq E_{bi}$, the two networks form a derived network structure which can be represented as $S_d \subseteq S_b$. We say $S_d$ is $S_b$'s derived network and $S_b$ is $S_d$'s base network.

This definition is injective rather than bijective which requires that all elements of the base network have the elements corresponding to the topology in the derived network, but not vice versa. This means that the derived network can extend or derive and overload its own structure and parameters after meeting the base network, so as to realize the flexible combination and extension of networks. On the other hand, a base network represents multiple derived networks with the same basic structure but different details extended, which embodies the ability to realize the generalized representation of cognitive network. **The calculation from the base network to the derived network is the deduction and application, and the calculation from the derived network to the base network is the induction and learning.**

Fig.2-a represents a derived network structure of binary relation, i.e. $S_d(A,R,B) \subseteq S_b(A_b,R_b,B_b)$, Fig.2-b represents the derived network structure of tree network.

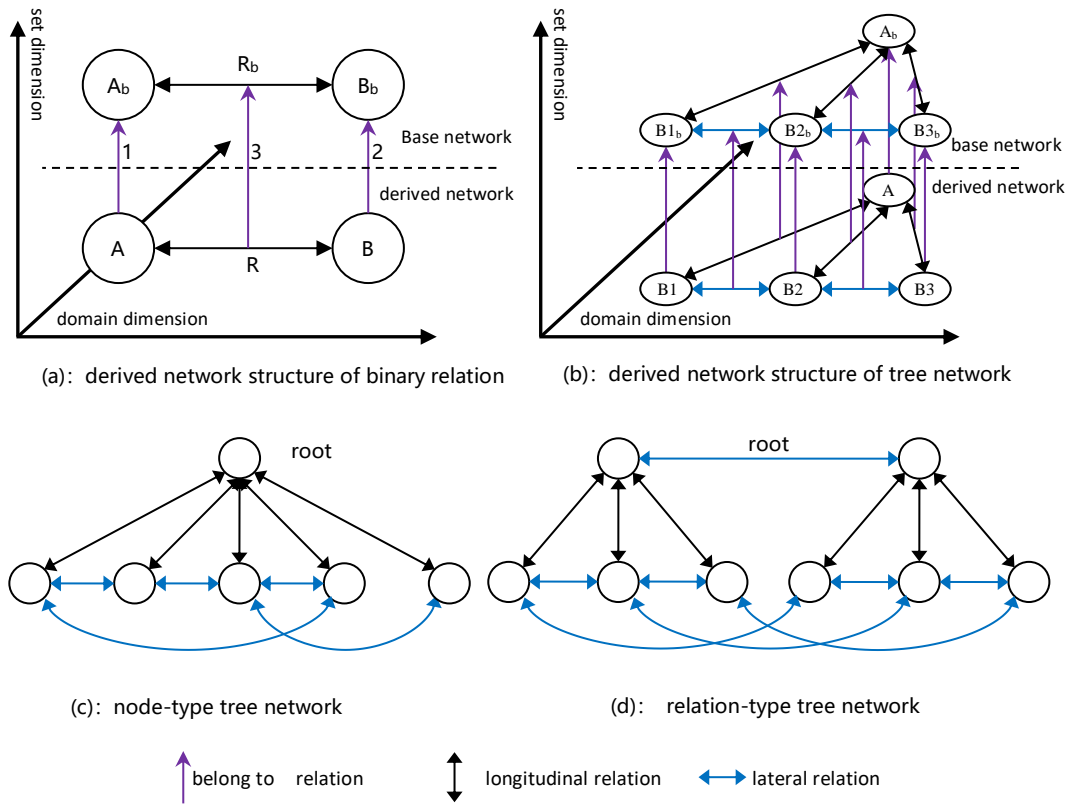

Fig. 2: The diagram of the basic structure of DC Net

**2. Tree-network Structure**

The tree-network structure is a multilevel structure in the domain dimension.

Definition:

1. A tree network is a connected network.

2. There is only one root element $E_r$ which can be a concept or a relation and it is the top-level element in the tree network.

3. Relations in a tree network can be divided into two categories based on their types and their connections to the root element:

  a. The longitudinal relation of the tree network: Multiple longitudinal relations are connected from top to bottom and take the root concept $E_r$ or the concepts of the two ends of root relation as the top-level concepts. These longitudinal relations are called the longitudinal relations of the tree network.

    b. The additional relation of the tree network: The other relations are called the additional relations of tree networks. Among them, the one whose basic type is a lateral relation is called the lateral relation of the tree network.

Fig.2-c represents a node-type tree network with the concept as the root element, while Figure.2-d represents a relational tree network with the relation as the root element.

The tree network can be understood from the main aspect and the secondary aspect.

1. The main structure is a tree structure with the root element as the topmost element and multilevel longitudinal relations extending from top to bottom, which reflects the hierarchical feature of the tree.

2. The additional relations added to the above tree structure forms the secondary structure, which makes the whole structure upgrade from the "pure" tree structure to the network structure.

Each element in the tree network is a part of the entire structure in which the root element can be used as a handle to represent the entire structure.

**The nesting of tree networks:** The tree network is the basic organizational structure of cognitive network. Every DC Net can be composed by various predefined tree networks. A tree network can shrink to a root element, and the connection between the root element and the external elements is regarded as the connection between the entire tree network and the external elements. So the relational structure inside the tree network is transparent to the outside and the tree network can be combined and nested with the root element as the joint point.

The definition of the basic tree network should follow the principle of minimization. Each independent tree network only defines its own indispensable components. Complex networks are always composed by the combining and nesting of basic tree networks, so that the tasks of cognitive description and computation can be fully decomposed and solved independently.

Fig.3 shows an example of a scene combined and nested by multiple tree networks.

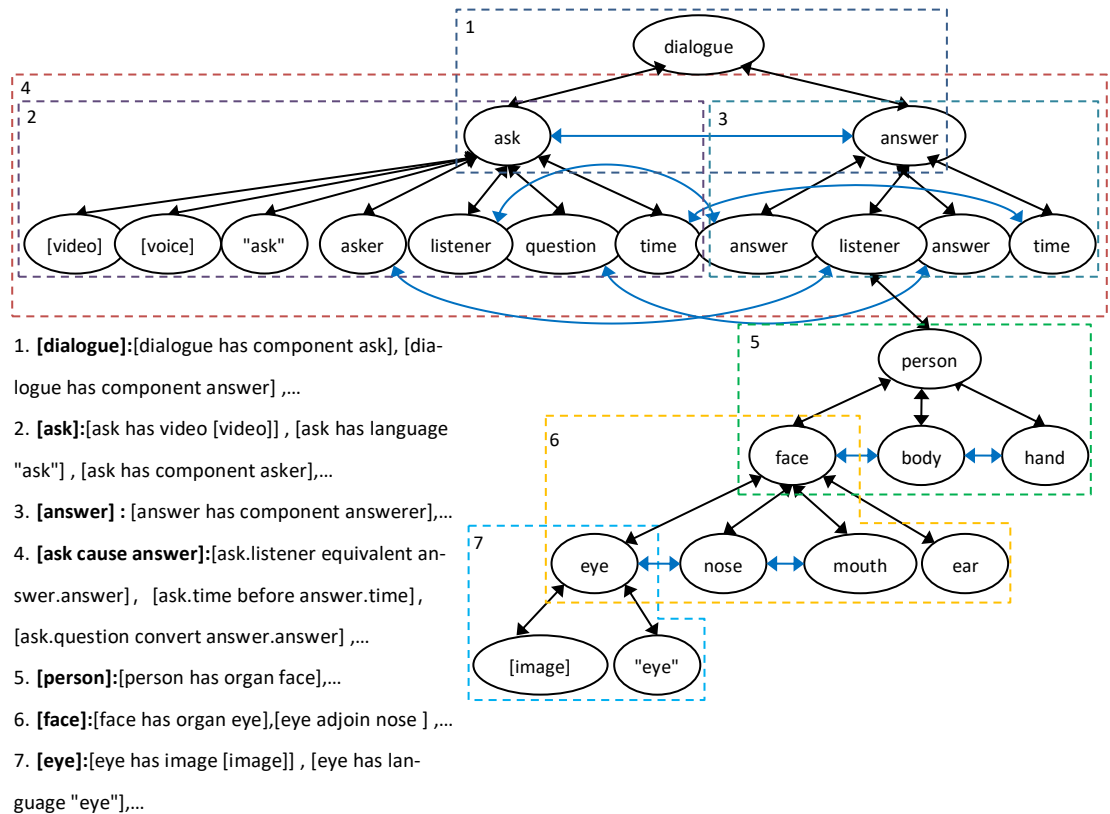

1. **[dialogue]**:[dialogue has component ask], [dialogue has component answer] ,…

2. **[ask]**:[ask has video [video]] , [ask has language "ask"] , [ask has component asker],…

3. **[answer]** : [answer has component answerer],…

4. **[ask cause answer]**:[ask.listener equivalent answer.answer], [ask.time before answer.time], [ask.question convert answer.answer] ,…

5. **[person]**:[person has organ face],…

6. **[face]**:[face has organ eye],[eye adjoin nose ] ,…

7. **[eye]**:[eye has image [image]] , [eye has language "eye"],…

Fig.3: The diagram of the combining and nesting of tree network

**The analysis of the meaning of the tree network:** Neither the pure tree structure nor the normal network structure can effectively represent the cognitive information. The normal network structure lacks the hierarchical information and is difficult to decompose the problem, while the pure tree structure lacks the ability to fully represent complex models. The structure of tree network combines the hierarchical structure and problem decomposition ability of the tree with the overall information representation ability of the network, and decomposes the complex cognitive representation into relatively simple partial problems to solve independently, which is of great significance to the development of AI. We think that the human brain also adopts a large number of logical structures similar to tree networks in its operation.

### 3. The Equivalence Principle of Derived Networks

The combination of derived network structure and tree network structure forms the equivalence principle of derived network. It is described as follows.

Assume that $S_d$ and $S_b$ are two tree networks whose root elements are $E_{dr}$ and $E_{br}$ respectively, and meanwhile, $E_{ds}$ and $E_{bs}$ are other elements except for the root element, then, $E_{dr} \subseteq E_{br} \Leftrightarrow S_d \subseteq S_b \Leftrightarrow E_{ds} \subseteq E_{bs}$.

The infer of the above formula is bidirectional and its meaning is that if the root element of a derived network belongs to the root element of a basis network, its elements except for the root element meet the element requirements of the basis network, and vice versa.

**The inference of partial meet:** The above formula shows that if the all elements of basis network are satisfied by the derived network, the satisfaction of the derived network to the basis network can be fully deduced. However, not all elements of the derived network can be known in practice, and it is often necessary to extrapolate only from some known elements. The result of this calculation is reflected as a membership degree. As long as the membership degree reaches a certain threshold value, it can be judged that the whole tree network will probably satisfy, and the whole tree network and unknown elements can be calculated and supplemented. This quantitative calculation is described in detail in the following probabilistic model.

The equivalence principle of derived network is one of the most fundamental principles by which human beings cognize the world and the thinking methods such as facing objects, deduction and induction all reflect this principle. The DC Net theory gives a definition in the form of a network for it and forms a mathematical quantized system combined with probabilistic calculation, which is used as the basis of cognitive description and computation.

### 3.2. Probabilistic Model

Before introducing the probabilistic model of DC Net, different computing scenes are compared and analyzed firstly based on the scope of available knowledge and the probabilities of computed results.

**The closed-domain scene:** Set relatively closed boundary conditions to limit the knowledge and information available to a small range and adopt a fixed structure to simplify the calculation.

**The open-domain scene:** Reduce limits on boundary conditions to extend the scope of available knowledge and information, adopt dynamic structure and pursue the computing which uses information more comprehensively.

**The scene of non-significance probability:** Based on known models, knowledge and information, it has a relatively low probability of the calculating result of unknown targets without significant value (e.g., $P<0.9$), and the goal of the computing is to select the relative optimal result as the final result.

**The scene of significance probability:** Based on known models, knowledge and information, it has a relatively high probability of the calculating result of unknown targets with obvious value (e.g., $P>0.9$), and the goal of the computing can pursue the absolute optimal result reaching the obviousness threshold as the final result.

It is very important to classify the computational scenes based on open domains and closed domains. The key of the evolution from the traditional system and weak AI to the strong AI is to change the assumption condition from the closed domain to open domain. There will be difficulties generated in this process which need to be solved, but the additional advantages can be gained at the same time.

To the closed-domain system, computing can become easy by objectively or subjectively constraining the known conditions and knowledge to the closed scope, which will generate two kinds of systems:

System A: In the scene with the closed domain and significance probability, the task target itself is relatively simple and clear, and the limited information of closed domain is enough to carry out accurate calculation. This is the job of many traditional computer systems, and these systems are mature and well functioning which are far more efficient than human beings.

System B: In the scene with the closed domain but non-significance probability, the limited information is not enough to deduce the absolutely correct result and only the "relatively excellent" result can be selected, which lacks reliability and stability. This is the current state of many weak

AI systems.

To the open-domain system, it become more complex because of being added more useful information. However, more excellent results or even absolute optimal results can be obtained if the added information can be fully utilized. It can be divided into two kinds of systems in detail.

System C: In the open-domain scene, computing significance probability can be realized after obtaining more information.

Here is an example: Human being's recognizing a person from an image is usually characterized by a high degree of confidence and stability, and that means the significant probability is obtained($P \approx 1$). In essence, the reason is that it makes full use of a lot of information in the image—nose, eyes, skin, clothing, environment···, which are all effective information to recognize person. the information is excessive—it exceeds the amount of information needed to compute the significance probability. In fact, even if lacking half of the information when half of a person is blocked out, we can still get the significance probability.

If significant probabilistic results can be obtained, then reliability, stability and interpretability will not be difficulty. When obtaining a significant result with the probability greater than 99% after computing, compared with choosing the more excellent result 75% from two results 75% and 25%, the difference is not measured by the difference 22% between 99% and 75%, but 25 times the difference between 1% and 25%. It also makes computing more easy since there is no need to enumerate different results and make comparison if significant results are obtained.

At present, the fundamental problem of the lack of reliability and stability of the system in this scene is that the rich information in the open domain is underutilized. Of course, this can be solved in theory and is also the keypoint to be broken through in work. What's more, once the rich information can be fully utilized, the way and scope of obtaining information can be expanded more actively to obtain more abundant information and further improve the reliability of the system, which is not a problem over current techniques.

System D: In the open-domain scene, after all available information is fully utilized, the result of significance probability still cannot be obtained after computing for the task target. Such a situation is not only a concern for AI systems, but also a permanent problem that humans themselves are constantly encountering and trying to solve. To AI or human intelligence, there is not a simple method to solve problems once and for all. Instead, many methods and steps need to be taken according to the situation to solve problems step by step. The general principle is followed: First, decompose the problems that can be classified as belong to system C to solve, and determine as much information as possible, and then surround and limit the uncertain system D with the determined information to reduce the scope and difficulty of the latter, and then solve it step by step.

After reexamining system B, we get two cases: In one case, the closed limitation of information is restricted by objective conditions which is difficult to change. Then, it is considered to belong to the open domain but impossible to obtain enough information, which can be classified into system D. In the other case, considering the false limitation or even active self-limitation caused by the lack of technical capacity at present, the technology should be improved to free this limitation and transform it into system C for effective solution.

Based on the above analysis, now the key is to implement system C well, which will be an important breakthrough towards strong AI.

The probabilistic model of DC Net is a probabilistic representation and computing model which is designed around DC cognitive network structure closely. With facing the scenes of open domains, significance probabilities, heterogeneous and dynamic network structure, it is called the cognitive probabilistic model. Its goal is to make full use of the large amount of knowledge and information that can be obtained in the open domain, and to calculate the result close to the significance probability for the task target.

Based on the above assumptions and design objectives, the cognitive probabilistic model has the following characteristics:

**Unifying probabilistic assessment systems**: In DC Net, multiple structures across domains are integrated into one, and all cognitive concepts and structures are described and computed by the same probabilistic system.

**Computing dynamic change with probability**: A large amount of knowledge and information can be obtained because of facing open domains, and they can be used as needed. The probability calculation is not for the static model with fixed structures, parameters and weights, but for the dynamic model with real-time changes. The calculation process and results follow dynamic consistency rather than static consistency. This characteristic is significantly different from most existing probabilistic theories and methods, which is actually closer to how the human brain works.

**The main method is probability addition**: Based on the assumptions facing open scenes and selecting as needed, the probability addition is the main method to calculate the probability of cognitive probabilistic models.

The following is a detailed description of the implementation of cognitive probabilistic model.

### 3.2.1. Probability Representation

**Bidirectional conditional probability:** DC Net adopts bidirectional conditional probabilistic model in which each relation has two conditional probabilities. The specific definition and value are determined by the semantics of the relation.

P(A|B): The probability of the A-end concept when the B-end concept in the relation exists.
P(B|A): The probability of the B-end concept when the A-end concept in the relation exists.

The conditional probability of the discrete value is represented as a probability value (0~1). The conditional probability of continuous values is represented by the probability distribution function and the corresponding membership function is used to calculate the membership degree. For example, the conditional probability of [adult has height] can be defined as the Gaussian distribution function N(μ=1.7m, σ=0.1m). The P(A|B) is calculated based on a height value x and the Gaussian membership function $f(x, \sigma, u) = e^{-\frac{(x-\mu)^2}{2\sigma^2}}$.

Partial examples of conditional probabilities are shown in Tab.1:

Tab.1: Sample table of conditional probabilities

| Basic Relation | Base Relation | Relation | P(B|A) | P(A|B) | Explanation |
|---|---|---|---|---|---|
| Belong to | Belong to | Apple belong to fruit | 1 | 0.00001 | |
| Equal | Equal | Apple equal apple | 1 | 1 | |
| Has component | Face has organ | Face has nose | 1 | 1 | |

| Has component | Face has organ | Face has eye | 1 | 1 | |
|---|---|---|---|---|---|
| Has component | Face has eye | European face has black eye | 0.3 | 0.05 | |
| Has form | Has shape | Face has oval | 1 | 0.00001 | |
| Has form | Has language form | Face has English form "face" | 0.9 | 0.6 | |
| Has form | Has language form | Apple has English form "apple" | 0.9 | 0.3 | "apple" can be parsed as [fruit], [company]… |
| Has attribute | Has attribute | Person has height | N(1.0,0.5) | 0.00001 | |
| Has attribute | Person has height | Adult has height | N(1.7,0.1) | 0.00001 | |
| Lateral relation | Conversion relation | Addition convert subtraction | 1 | 1 | |
| Lateral relation | Causality relation | Ask cause answer | 0.95 | 1 | |

**Input probability and result probability**：Each concept in DC Net has probability, and there are two probabilities of a concept needed to be recorded when computing a network. The probability which is set directly based on initial conditions or external calculations is called the input probability. The result probability of each concept can be calculated with the input probability and conditional probability. It's called probability for short.

**Probabilistic overloading:** When overloading a relation, the conditional probabilistic value which is different from the base relation can be set for the derived relation.

For example, the probability of base relation [face has eye] is $P(B|A)=1$, $P(A|B)=1$ and the probability of the derived relation [European face has black eye] is $P(B|A)=0.3$, $P(A|B)=0.05$. Based on these two relations, using the known information that the B end is black eye, it can be inferred that the probability of the A end being face is 1 and being European face is 0.05.

### 3.2.2. Probabilistic State

The cognitive probabilistic model is different from traditional probabilistic graphical model. It does not calculate a probability value for a cognitive network, such as joint probability or potential function value, but describes the probabilistic state of the cognitive network with the probability of each element. To calculate the probability of DC Net is to continuously calculate the probability of each element. There is no clear boundary between the process and end of the calculation. It can be finished or restarted at any time. The ideal result is each element achieves probabilistic collapse, which is equivalent to the whole cognitive network getting a completely definite result.

Obviously, the probabilistic state of the cognitive network is more in line with human cognitive model. Take the image recognition, for example. Human recognizing an image is not to give a probability value for the whole image, but to give different recognition probabilities for different parts.

Some computing tasks do need to give an evaluation value for the whole network, which should not be the joint probability, but a weighted average value calculated by combining probabilities with some evaluation indexes set based on task objectives.

### 3.2.3. Probability Calculation

**Conditional probabilistic calculation:**

Assume that the conditional probabilities of the relation $R_{AB}$ in both directions is $P(A|B)$ and $P(B|A)$, then:

calculate the probability of A based on the known B's probability $P(B)$:

*Formula 1*: *P(A)=P(B)\*P(A|B);*

calculate the probability of B based on the known A's probability P(A):

*Formula 2*: *P(B)=P(A)\*P(B|A);*

**The formula of probabilistic superposition:**

The definition of conditional probability follows the independent sampling hypothesis. To calculate the probability of a concept X, we can figure out n probabilities based on the n concepts $(Y_1, Y_2, \ldots, Y_n)$ that have direct relations and the conditional probability, and then add them together with probability addition.

The binary probabilistic superposition is calculated by the following formula:

*Formula 3: $P(X)=P(X_1 \cup X_2)=P(X_1)+ P(X_2)-P(X_1)P(X_2)$*

Undoing a binary probabilistic superposition is calculated by the following formula:

*Formula 4: $P(X_1)=(P(X)-P(X_2))/(1-P(X_2))$*

The n-element probabilistic superposition is calculated by the following formula:

*Formula 5: $P(X)=P(X_1 \cup X_2 \cup \ldots X_n)$*

$$= \sum_{i=1}^{n} P(X_i) - \sum_{1 \leq i \leq j \leq n} P(X_i X_j) + \sum_{1 \leq i \leq j \leq k \leq n} P(X_i X_j X_k) - \cdots + (-1)^{n-1} P(X_1 X_2 \ldots X_n)$$

In the above formula:

$P(X_i) = P(Y_i) R(XY_i) P(X|Y_i)$

$R(XY_i)=0\sim1$: Relational membership degree. For example, the relation of [face has eye] includes the parameters such as [angle], [distance] and so on. The relational membership degree is calculated based on the actual values of these parameters on the derived relation and the set values on the base relation.

For $P(X_1 \cup X_2 \cup X_3) = P(X_1 \cup X_2) \cup P(X_3) = P(X_1 \cup X_2) \cup P(X_3) = P(X_1) \cup P(X_2 \cup X_3)$, the n-element probabilistic superposition can be decomposed into binary probabilistic superposition in any order, which is convenient for multi-directional computation and parallel processing.

**Probabilistic passing and superposition(PPS):** After the input probability P(Y) is added to a concept Y, P(Y) takes Y as the starting point for probabilistic passing and superposition to the adjacent concept $X_i$ based on the conditional probabilities of related relations, so that the probability of each $X_i$ can be improved. Since the conditional probability is less than or equal to 1, this passing is decaying in principle. When the passing reaches to a certain range, it will decay to the point where it can be ignored and terminated.

The specific algorithm adopts breadth-first network traversal algorithm. Once the PPS algorithm is executed, the probabilistic passing and superposition of P(Y) to each $X_i$ is completed. The algorithmic flow is as follows:

1. Initial conditions: With the concept Y adding the input probability increment P(Y), mark Y as [traversed], taking Y as the starting point to start computing.
2. Query for all relations $R_{Xi-Y}$ with Y on one end and $X_i$ on the other end.
3. Each relation $R_{Xi-Y}$ and the end concept $X_i$ are treated as follows:
   a. Return if any of the recursive termination conditions is satisfied;
   b. Otherwise, turn to calculate the probability of P(Y) acting on $X_i$, $P(X_i)= P(Y)R(X_iY)P(X_i|Y)$, and superimpose the probability of $X_i$ according to Formula 3; mark the concept $X_i$ as [traversed] and set $Y=X_i$, $P(Y)=P(X_i)$, then go to 2 for recursive processing.

Recursive termination conditions:
1. If the opposite-end concept $X_i$ is marked as [traversed].
2. The probability of the opposite-end concept $X_i$ has collapsed.
3. The calculated value of the probability $P(X_i)$ is very small, indicating that the passed probabilistic influence has decayed to be negligible.
4. The cut-off limit set for the computing task has been reached.
5. Other recursive termination conditions based on task requirements and relational semantics.

**Multiple paths and rings**: DC Net is an omnidirectional describing network, which does not exclude the exist of rings. However, in each PPS calculation, the probabilistic influence of the starting point concept Y on the target concept $X_i$ is allowed to be calculated by only one path, so rings will not appear. When there are multiple paths, it is not guaranteed that the calculated probabilistic results of choosing different paths for the same target concept are completely consistent, but if the network's design and probabilistic values are reasonable, the final results should be consistent. The rule of choosing paths can be set as selecting the path with the highest probabilistic value calculated for the target concept.

On multiple input probabilities $P(Y_i)$, the PPS algorithm is invoked one by one to perform a probabilistic superposition on the target concept X. Therefore, the probabilistic passing and superposition can be divided into two aspects:
1. The input probability of a concept $Y_i$ passes the probabilistic influence of a target concept X by multilevel conditional probabilistic multiplication.
2. The influence of the input probability of multiple concepts $Y_i$ on the probability of a target concept X is reflected in the superposition of the above probabilistic multiplication results.

**Extended discussion:**

It has a clear mathematical explanation to calculate the probabilistic superposition according to Formula 5, and the calculation result is relatively accurate in theory, but the calculation process is very complex. A simplified model can be used in the actual calculation, and a feasible scheme is as follows:

Simplify Formula 5, only retain the separate probabilistic sub item in the front and remove the joint probabilistic part in the back, and set a compensation coefficient $K_i$ for each probabilistic sub item for compensation processing, and then the simplified probabilistic superposition formula is obtained.

*Formula 6: $P(X) = \sum_{i=1}^{n} P(X_i) = \sum_{i=1}^{n} k_i P(Y_i) R(XY_i) P(X|Y_i)$. $0 < k_i <= 1$.*

After the simplification, the value interval of the probability is changed from (0~1)) to (0~>1), and the probabilistic collapse threshold (see the below) can be set as P>=1.

The value of the compensation coefficient $K_i$ is determined according to the actual model and relation definition (usually 0.5~1). This coefficient acts as compensation, making the calculated P(X) in the interval (0~1) as close as possible to the theoretical value calculated by the normal probabilistic superposition formula and preventing P(X) from reaching collapse too fast. Once the probabilistic collapse is reached, the probability will not continue to increase according to the collapse rule below, and the subsequent processing is completely the same as the normal probabilistic superposition.

For example, assuming that the known probabilities of the nose, eye and mouth are all 1 and the conditional probabilities of them to the face are all close to 1, then any one of these parameters available is enough to make the probability of the face equal to 1. After the compensation coefficient 0.5 is added, the calculated probability of each parameter is reduced to 0.5, and the superposition of two parameters is needed to make the probability of the face equal to 1, which is not an essential problem under the assumption of parameter excessive and significance probability.

Compared with Formula 5, the simplification of Formula 6 reduces the theoretical accuracy, but its representation and computing process are simple, so it is widely used in many scenes.

It can be seen that Formula 6 is similar to the main part $\sum \omega_i \cdot x_i$ of the M-P model's basic formula $\sum \omega_i \cdot x_i + b$. In our opinion, this indicates that similar polynomials are the basic formulas for doing uncertainty calculation effectively and are more suitable for open-domain scenes than the complex joint probabilistic calculation used commonly in probabilistic graph models.

On the other hand, compared with the weights and thresholds in the M-P model, the parameters and calculation results in the cognitive probabilistic model are the probabilities with values of (0~1), which are of clear interpretability and independent adjustment ability. The computing process is carried out by the probability theory only, without the need to select and train different activation functions and bias terms. In addition, it will make the whole system more complete to add the relational membership degree $R(XY_i)$ calculated based on relational parameters to the probabilistic passing and superposition.

Neuroscience research reveals that one of the basic working modes of human brain neurons is close to the polynomial calculation and activation model, which is also the main theoretical basis of the M-P model. We think that this opinion is generally reasonable. The human brain tends to search for more effective information to figure out a result of significant probability, and try to avoid relying only on insufficient information to perform "accurate" probabilistic calculations and make choices among the multiple results of non-significant probabilities. The working mode similar to the above simplified probabilistic superposition model can effectively meet the needs.

### 3.2.4. Probabilistic Collapse

Based on the hypothesis of the open domain and significant probabilistic scene as well as the characteristics of probabilistic superposition and omnidirectional computing, The principle and method of probabilistic collapse were designed.

According to the idea of scene fitting, all computing is the process of computing the unknown to make it gradually become known. The true probability of each concept in the correct target network is 1, but only part of the information is known at the beginning of the calculation, which is reflected as the input probabilities(P<1) of some concepts. To calculate the probability of each concept with the known input probabilities according to the above probabilistic algorithm is to gradually change the unknown into known.

We stipulate that if the probability of a concept reaches a preset significant threshold (e.g., P>0.9) during the calculation, we think we obtain the true probability 1 of the concept, so the collapse processing of the concept is triggered. The collapsed concept is said to be in the "collapsed state", while the not yet collapsed concept is said to be in the "superposition state".

The collapse processing of a concept is executed according to the following order:
1. Cancel the PPS calculation that had an impact on X before collapse to avoid repeated superposition. Refer to Formula 4 for the specific algorithm.
2. Set the input probability of X equal to 1 and restart the PPS calculation.

The probability of a concept reaching collapse will have an impact on subsequent calculations: This probability does not need and cannot continue to increase, and the subsequent probabilistic superposition will be cut off and in no need of further passing when encountering this concept. The more important is that this concept becomes absolutely known, which changes the known and unknown state distribution of the whole network, and causes the reorganization and simplification of the subsequent calculation process and even the change of calculation targets and methods.

For example, when the probability of the face reaches collapse based on the mouth and nose, there is no need to use the eye and ear to continue the meaningless calculation of the face! What's more, after the face becomes absolutely known, it can extrapolate the eye, ear, etc., even if their images are not clear or even completely invisible such as covered by sunglasses.

Combined with the probabilistic collapse, the probabilistic calculation process of the whole cognitive network is shown as followings:

- Every concept in an initial DC Net is in the state of probability =0.
- With the addition of known information and input probabilities, the probability of each concept is gradually improved by PPS calculation, and at this time each concept is in a superposition state.
- The probability continues to be improved. When a concept first reaches the collapsed state, the complexity will decrease since the collapse of a concept will divide the entire superposition-state network into several sub-networks for solving separately.
- At the same time, the collapse of one concept will increase the input probability, which makes other concepts reach the collapse faster and pushes the whole network to collapse—" **One collapse, more collapse**".

The differences in processing compared with some traditional systems:

Some systems ignore the probability completely, which is equivalent to forcibly collapsing all concepts into the state of either 0 or 1 at the beginning. There is no problem for the system facing the certain domain, while for the system with a lot of uncertainties, it is equivalent to introducing the probability deviation at the beginning, and more deviations are added to each of subsequent steps, so the final result will not be ideal.

Some systems introduce the processing of uncertainty (0~1). However, their hypotheses and design principles don't contain the theory and method of probabilistic collapse, or its algorithm system cannot support selecting as needed on demand and multi-directional calculation. This makes the whole system keep probabilistic superposition state and cannot be simplified by local collapse in the calculation process, and only at the last step can the relative optimal result be selected for collapse. In the process, the scale and calculation quantity of the uncertain networks become very large, the performance is very low while the result may not be more accurate(The passing and calculation of uncertainty themselves have accumulated deviations, **while the correct local collapse will eliminate such deviations in the intermediate process!**). What's more, the more information and the bigger the network, the more obvious the problem becomes.

Probabilistic collapse is also the basic thinking mode of the human brain. Human beings are always confronted with unknown and uncertain information when observing and interpreting the world, and we need to determine and even manipulate the prioritized clear information as quickly as possible. Once part of the information is determined as known, we can change the focus and the direction of computing and reasoning, and calculate the other unknown based on the known. With the continuous transformation of conditions and computations, we can deal with the complex world effectively. If the information that should be determined cannot be determined as soon as possible, nothing can be done in the face of the "chaotic" system with more and more uncertain information.

### 3.2.5. Probability extension implementation

The above basic probabilistic processing model can realize most of the cognitive probabilistic description and calculation. If needed, the basic probabilistic calculations of some parts can be replaced by the more accurate extended probabilistic calculations (with the logical calculations such as "and" and "or", etc.). The partial optimization of computational accuracy does not affect the overall probabilistic system.

## 3.3. Algorithm system

The core algorithm system of the DC Net is discussed below.

### 3.3.1. Algorithm overview

The core algorithm system of the DC Net revolves around the cognitive network structure, and the task targets and probability states jointly drive the matching and growth of the cognitive network to achieve various task targets. It can be called **omnidirectional network matching-growth algorithm system (OMG)**. The basic principles and main features are as follows:

**Network growth**: According to the scene fitting theory, the core of each calculation task is to build a cognitive network to fit a specific target scene, and the construction process is reflected in the continuous growth process of the cognitive network.

**Dynamic combination**: The specific realization method of network growth is to dynamically combine cognitive network components to form a larger network. This combination can be nested infinitely and merged across domains with strong flexibility.

**select to use as need**: There are a large number of cognitive network components in the knowledge base, and the actual calculations select suitable components to use according to the needs.

**Algorithm fusion**: Various algorithms including parsing, classification, generation, query, reasoning, prediction, action, learning, etc. are fused into a unified network growth algorithm.

**Omnidirectional growth**: In the definition and description of the cognitive network, the input, output and calculation direction are not set. Only in the specific calculation, some parts of the network are set as known and some parts are calculated as unknown. **The division of known and unknown is also relative, which is completely distinguished by probability!** Therefore, the entire calculation process and network growth can be carried out in any direction and adjusted in real time. Model design such as bidirectional conditional probability provides basic support for this.

The overall algorithm can be described in the following form:

$S_k' + S_t' = P_s(T, S_k, S_t, S_I)$

$S_I = P_I(I)$, $O = P_O(S_k' + S_t')$

$P_s$: Internal calculation of DC Net.

T: Task information. including task objectives and correlative parameters.

$S_k$: Existing knowledge network.

$S_k'$: New knowledge network.

$S_t$: Existing instance network.

$S_t'$: New instance network.

$S_I$: Conceptualized input information.

I: Input information.

$P_I$: Input adaptation processing.

O: Output information.

$P_O$: Output adaptation processing.

The overall algorithm flow is as follows:

1. Input processing: the external input information I is conceptualized into the concept of DC Net structure $S_I$ through the input adaptation algorithm $P_I$;

2. $P_s$ calculation: Combine $S_I$ with the existing $S_t$ and $S_k$, and calculate according to the requirements of task T to generate new instance network $S_t'$ and new knowledge network $S_k'$;

3. Output processing: extract the required parts from the generated $S_k'$ and $S_t'$ and convert them into the output information O required by the task through the output adaptation algorithm $P_O$;

The basic process of $P_s$ calculation is as follows:

1. Initial conditions: The input $S_I$ is a collection of DC Net segments, each element has an input probability, and conflicts and ambiguities between elements are allowed. $S_t$ is existing instance network, there can be multiple $S_t$, and each $S_t$ is grown separately. As an optimization, $S_t$ with a higher probability will preferentially grow, and the growth of $S_t$ with a low probability can be suppressed.

2. Matching: select a network segment $S_{Ii}$ from $S_I$ and combine some elements of $S_t$ to form a network segment $S_d$ for matching in the knowledge base $S_k$. If it succeeds, it will return one or more knowledge networks $S_b$. The membership degree matching calculated with $S_d$ to $S_b$ should be reach the activation threshold (activation threshold < collapse threshold, set according to different tasks).

3. Growth: Create a derived network $S_D$ by the base network $S_b$ as a template and add it to $S_t$ for growth. $S_d$ will be absorbed into $S_D$.

4. Processing: Probabilistic superposition, probability collapse, network reduce, etc. are processed in the process. If necessary, the knowledge network $S_k$ will also be expanded or modified, that is learning.

5. Iteration: Repeat the above steps until all elements $S_{Ii}$ of $S_I$ are calculated.

6. End: If the overall task ends, select the optimal result from $S_t$ that has absorbed all $S_I$ for output processing, otherwise, wait for the new $S_I$ input for the next round of calculation. According to the theory of cognitive probability, it is preferred to select the absolute optimal result where all elements reach the collapsed state. If there is no absolute optimal result, it is also possible to compare each $S_I$ and select the relatively optimal result.

### 3.3.2. Match

The network matching is a very basic algorithm. The basic method is to calculate the membership degree of one or a group of network segments $S_d$ to a base network $S_b$ based on the equivalence principle of derived network and the probability superposition formula.

**Complete matching**: Refer to the equivalence principle of derived network. If $S_d$ is a complete derived network of $S_b$, that is, $S_d \subseteq S_b$, then $S_d$ matches $S_b$ completely, that is, the membership degree P=1. Some tasks require complete matching, such as question queries.

**Partial matching**: According to the principle of known information is incomplete, usually known $S_d$ is not a complete derivation network of $S_b$ but just a sub-network of the derivation network. In that way perform partial matching with $S_d$ to $S_b$ to calculate the membership degree P. The specific algorithm is as follows:

- **Single-concept matching:** A concept can be regarded as a smallest tree network, To calculate the membership degree P of one concept $S_d$ to another concept $S_b$. There are two cases:
  1. For the discrete concept, if $S_d \subseteq S_b$, then P=1, otherwise P=0.
  2. For the continuous value concept, according to the probability distribution function recorded in the base concept $S_b$ and use the corresponding membership function to calculate $P=N(S_d, S_b)$.

- **Matching of tree network:** The matching calculation of a tree network is divided into two steps:

  1. Structure matching: Based on the base network $S_b$, assuming that $S_d$ is a sub-network of a derived network $S_D$ of $S_b$, the structure matching is performed according to the definition of the derived network structure. If the derived structure matches successfully, execute the next step, otherwise it means that it does not match and returns 0.

  2. Membership degree calculation: According to the above structure matching results, substitute each element $S_{di}$ of $S_d$ as known information into the corresponding element $S_{bi}$ of the base network $S_b$, and calculate the membership degree $P_i$ of each element separately.

  Then $P_i$ is used as the input probability on $S_b$ and call the PPS algorithm to superimposed probabilistic. Finally, the probability of the root element of $S_b$ is the membership degree P of $S_d$ to $S_b$.

- **Matching of nested tree network:** Recursive algorithm is used to match each level of tree network level by level, and finally obtain the probability of root element of the top tree network is the membership degree P of the entire tree network.

### 3.3.3. Growth

The growth algorithm uses the base network in the knowledge base as a template, and grows the instance network and superimposes probability by creating, combining, and adding elements.

First of all, we define the three basic growth algorithms:

**1. Single concept growth**: Using the base concept $S_b$ in the base network as a template to copy a derived concept $S_d$, and set the belong to relation $S_d \subseteq S_b$. The ID of $S_d$ has a new unique value, and other parameters are the same as those of $S_b$.

**2. Bidirectional growth**: Add a relation between the two derived concepts $S_{di}$ and $S_{dj}$ to connect them, and at the same time, connect and grow the network where they are located and superimpose probability. The steps are: copy a derived relation $R_{di\text{-}dj}$ with the specified base relation $R_{bi\text{-}bj}$ as a template and set the relation $R_{di\text{-}dj} \subseteq R_{bi\text{-}bj}$, set the two ends of $R_{di\text{-}dj}$ for $S_{di}$ and $S_{dj}$ to connect them, and finally cancelled and then redone the previous PPS calculation which affects $S_{di}$ and $S_{dj}$ to complete the mutual probability passing and superposition of the two networks.

**3. Unidirectional growth**: similar to bidirectional growth, but only $S_{di}$ exists and $S_{dj}$ does not exist at this time. The steps are: first perform single concept growth to complete the growth of $S_{dj}$, then perform the connection growth and probability superposition of $S_{di}$ to $S_{dj}$ as in the above-mentioned bidirectional growth.

**The growth of tree network**: The growth of the derived network $S_D$ with $S_b$ as the base network and $S_d$ as the sub-network is divided into two steps:

> 1. Set the belong to relation between each element of $S_d$ and the corresponding element in $S_b$ to complete the growth of these existing elements.
> 2. Use $S_b$ as a template to find elements $S_{dn}$ that have not been created in $S_D$, and create them one by one to complete the growth of $S_D$. Each step of growth selects one of the three basic growth algorithms mentioned above according to the actual situation.

The result of growth is: an instance network $S_D$ derived from $S_b$ is formed. $S_D$ contains the known element $S_d$ and the newly grown element $S_{dn}$. $S_D$ and $S_b$ establish a complete derived network relation, and each element of $S_D$ has new probability.

**Note:**

1. As an optimization, if the probability of some elements in the newly grown $S_{dn}$ is lower than the activation threshold, doesn't create them immediately and the process is postponed. In addition, unnecessary the growth of elements can also be restricted according to the characteristics of the task. See the reduction algorithm below for details.

2. If the target object which needs to be created already exists, it will directly superimpose probability instead of creating it repeatedly.

3. The growth algorithm can be launched from any location in the network and can proceed in any direction. The specific growth direction is mainly affected by two factors: firstly, it is limited by the task target, which is not described in detail here; secondly, heuristic growth is driven by probability, and the highest input probability is usually selected to grow first.

4. The concept of reaching a collapsed state becomes a completely definite result, which will inhibit the growth of concepts which have conflict with it.

**Algorithms such as analysis, reasoning, and generation are essentially the same omnidirectional network growth algorithm!** All are based on some known concepts to complete the entire network. The difference is that the initial calculation direction is different due to the initial setting of the known and unknown. During the calculation process, the calculation direction will be dynamically changed with the change of known and unknown. The network growth is not limited to unidirection but omnidirection. Various algorithms can also be combined to complete the task together. A specific explanation is given below.

**parsing**: The network growth of calculate high-level concepts based on low-level concepts. also known as bottom-top parsing.

For example:
- In natural language processing, internal cognitive networks are generated based on language strings.

- In image processing, the nose, eyes, and face are generated based on super pixels.
- In image processing, generate faces based on nose or eyes

**Generative parsing**: The network growth of calculate low-level concepts based on high-level concepts. also known as top-bottom parsing.

The goal of the parsing task is also to complete the entire network, but the known information is usually some low-level concept. According to the principle of omnidirectional growth, the probability of high-level concepts in the execution of parsing tasks will gradually increase to become known and launch top-bottom growth to generate low-level concepts. The overall goal of this "generate" is still to complete the parsing task.

**Lateral growth**: Lateral growth is the growth based on lateral relations such as adjacent relations, conversion relations, and causal relations. It has no essential difference with the above-mentioned longitudinal parsing and generative parsing calculations, and it is also a bidirectional growth.

For example:
- Mutual growth according to the adjacent relation between eyes and nose.
- Mutual growth according to the conversion relation of addition and subtraction.
- Mutual growth according to the causality relation of the event.
- Perform movement calculations according to movement relations.

**Generation**: The generation algorithm is a top-bottom and lateral unidirectional growth, usually triggered by tasks such as language generation and image generation. It different from generative parsing, the high-level concepts in the generation algorithm are known but the low-level concepts are completely unknown (P=0). Top-bottom growth is a unidirectional "pure" generation.

The specific implementation of the generation algorithm is to generate low-level concepts based on known high-level concepts combined with downward and lateral connections. The result is determined by parameters such as the conditional probability of the relation and other conditions set by the task. Because there is no known concept at the low-level as constraint, the possible results of unidirectional generation will be more. For example: to draw a tree, 100 people will have 100 different painting methods, all of which are equally effective.

For example:
- In natural language processing, the low-level cognitive network is generated based on the expansion of the high-level cognitive network.
- In natural language processing, language strings are generated based on internal cognitive networks.
- In image processing, small parts are generated based on big concepts. such as noses and eyes based on faces.
- In image processing, super pixels are generated based on concepts such as face, nose, and eyes.
- In the action plan, an action plan is generated according to the big task target.

In theory, the structure and algorithm of the DC Net can support complete generation tasks. But in specific implementation, more optimized solution is that the intelligent system is responsible for the generation of higher-level scenes and a small number of large-granularity concepts, and the low-level detail generation and rendering processing are completed by professional systems. This reflects the significance of the aforementioned "intelligent system + external system" solution.

**Complement hidden or omission**: There are hidden objects which do not have a directly visible form in various task processing. For example, there are a lot of hidden parts in image processing, and there are also a lot of omission in natural language processing. The complement of these objects is also embodied as growth algorithm, which is based on known concepts to calculate unknown concepts which need to be completed. Similarly, this calculation can also be performed in multiple directions.

**Error correction**: There is no essential difference between error correction and complement hidden or omission, except that the target object has wrong known information. Therefore, the correct known concept needs to be based on a stronger (higher probability) for error correction to trigger the use of the calculated result as the correct result and forcibly discard the wrong known information.

See Fig.4 for schematic diagrams of several typical growth algorithms.

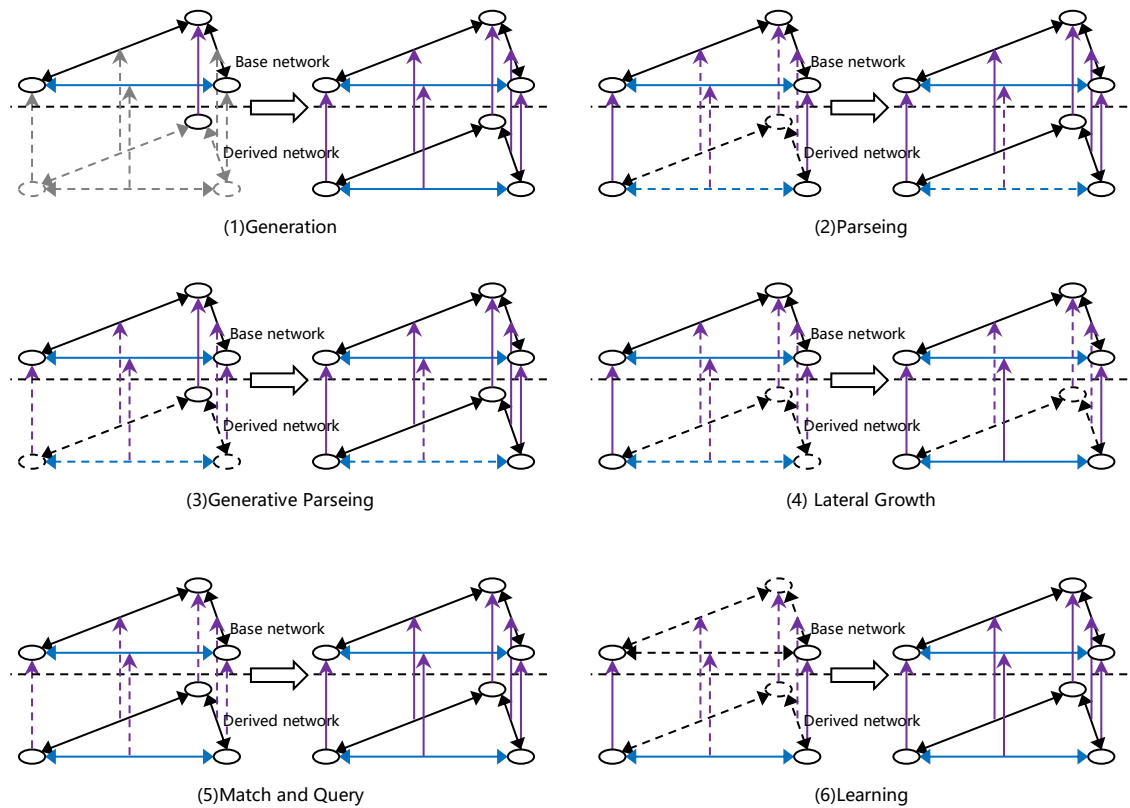

**Solid line:** elements that have grown; **dashed lines:** elements that objectively exist and have not grown; **gray dashed lines:** elements that objectively do not exist

**Black:** concept and longitudinal relation; **purple:** belong to relation; **blue**: lateral relation;

Fig.4：Typical Algorithm Schematic Diagram

Here is a simple example of a growth algorithm. The specific task is to recognize the face in

Fig.5. For simplicity, only one level of calculation is involved and only bottom-top parsing and top-bottom generative parsing processing are used.

Initial conditions:
- In the previous stage, the probability of eyes, nose, mouth, ears, face, eggs, ears, cup handle, etc. has been calculated according to the relation (green connection) such as [eye has image]. They are as the input probability in this round of calculation. among the face and eggs, ears and cup handles are conflicting and ambiguous interpretations, and there is mutually exclusive relation (red line).
- This round of calculation is based on the relations such as [face has nose][face has eye][face has mouth][face has ear][cup has cup handle], and their bidirectional condition probabilities are set as: P(B|A)=1 , P(A|B)=1.
- It is assumed that the relation membership degree calculated for each relation according to parameters such as distance and angle all satisfy R(AB)=1.
- The collapse threshold set to 0.9.

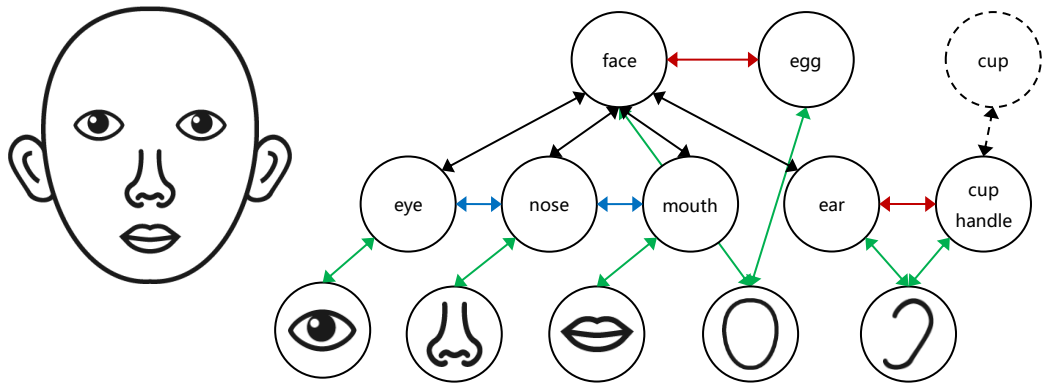

| Steps | eye | nose | mouth | ear | face | egg | cup handle | cup |
|---|---|---|---|---|---|---|---|---|
| 0. Initial conditions | **0.6**,0.6 | **0.5**,0.5 | **0.4**,0.4 | **0.1**,0.1 | **0.3**,0.3 | **0.5**,0.5 | **0.4**,0.4 | Not created |
| 1. Eyes launch calculation | **0.6**,0.6 | **0.5**,0.8 | **0.4**,0.76 | **0.1**,0.64 | **0.3**,0.72 | **0.5**,0.5 | **0.4**,0.4 | Not created |
| 2. Nose launch calculation | **0.6**,0.8 | **0.5**,0.8 | **0.4**,0.88 | **0.1**,0.82 | **0.3**,0.86 | **0.5**,0.5 | **0.4**,0.4 | Not created |
| 3. Mouth launch calculation | **0.6**,0.88 | **0.5**,0.88 | **0.4**,0.88 | **0.1**,0.892 | **0.3**,0.916 | **0.5**,0.5 | **0.4**,0.4 | Not created |
| 4. Face collapse calculation | **0.5**,1.0 | **0.5**,1.0 | **0.4**,1.0 | **0.1**,1.0 | 1.0 collapse | **0.5**,0.5 | **0.4**,0.4 | Not created |
| 5. Results | **1.0** collapse | **1.0** collapse | **1.0** collapse | **1.0** collapse | 1.0 collapse | **0.5**,0.5 | **0.4**,0.4 | Not created |

**Normal numbers**: result probability;

**numbers in bold**: the input probability of having performed PPS;

**numbers in bold underlined**: the input probability of not having performed PPS;

Fig.5: Example of Network Growth Calculation

The calculation steps performed according to the above initial conditions are expressed in the table of Fig.5. The calculation process is explained as follows:
- Step 1. Because the input probability of the eyes is 0.6 which is the highest, the calculation

is launched by the eyes first: the face is parsed with the input probability of the eyes 0.6 combined with the relation [face has eyes], and the probability of passing to the face=0.6×1=0.6, The probability of the superimposed face=0.3+0.6-0.3×0.6=0.72; then use the probability 0.6 that passed to the face to continue the generative parsing calculation of the nose according to the relation of [face has a nose], and the probability of passed and superimposed nose=0.5+ 0.6-0.5×0.6=0.8; in the same way, the probability of mouth=0.76 and the probability of ear=0.64.

- Step 2: Now the nose has the highest input probability. The calculation which launched by nose increases the probability of face, eyes, mouth, and ears.
- Step 3: Calculations launched by the mouth increases the probability of face, eyes, nose, and ears. The probability of the face reaches 0.916, which becomes a collapsed state.
- Step 4: Perform probability collapse processing for the face, set the input probability of the face to 1.0, then perform the probability passing and superposition calculation on the eyes, nose, mouth, and ears. Because the conditional probabilities are all set to 1.0, the probabilities of eyes, nose, mouth all reached 1.0 and became a collapsed state.
- Step 5: Eggs, cup handles and already collapsed faces and ears have a mutually exclusive relation. They are suppressed by probability without calculation. All other objects are collapsed. the calculation is over, the network constructed by collapsed objects becomes the final result.
- According to the principle of on-demand selection, the concepts, relations and calculation paths involved in each calculation do not need to be unique. The calculation path in this example only uses the longitudinal relation, and another calculation may use a different calculation path including the lateral relation. In any case, if the definition and probability settings of cognitive knowledge are reasonable, the final results obtained by different calculation paths will also tend to be stable.

According to this example, we will further explain and analyze the main features of the algorithm:

**Mutual calculation**: Each unidirectional calculation is based on a concept take PPS calculation other concept for increasing the probability of the latter, all concept are calculated mutually. For example, the face and the nose perform direct mutual calculations, while the nose and mouth perform indirect mutual calculations through the face. The ultimate goal of combining each concepts calculation is to calculate and complete the entire network.

**Omnidirectional growth**: The overall goal of the example task is to parsing, but the specific calculation and growth are not limited to the bottom-top, but also the top-bottom. Before the face collapses in step 2, the eyes parsed bottom-top the face according to the P(A|B) of the relation [face has eyes], while with the increased probability of the face, it is also performs a top-bottom generative parsing of the nose, etc. through the P(B|A) of relation [face has nose] etc.; in step 4, after the face reaches the collapse, a generative parsing of the eyes, nose, and ears is launched with the additional input probability due to the collapse; Moreover, it is also possible to carry out lateral bidirectional growth according to the lateral relation.

**Probability-driven**: The specific realization of omnidirectional growth is to select the highest priority relation and direction according to the task target and probability state, and dynamically adjust them. Ensure that each step of the calculation is more effective, and noneffective or insignificant calculations are postponed, eventually may be completely suppressed and cancelled.

**Probability collapse**: Let the probability of each concept increase rapidly and achieve that probability collapse is both a global and a local target of calculation. As a local target, the calculation task can be simplified by obtaining phased determination results. In step 4 of the example, after the face reaches the probability collapse, cause the nose, etc. also reach the probability collapse.

**Probability suppression**: The oval shape of the face is interpreted as an egg with a higher probability than the face, and the arc shape of the ear is interpreted as a cup handle with a higher probability than the ear. However, the preferential growth of eyes and other concepts postponed the growth of eggs and cup handles, while the growth of faces and ears went smoothly and reached a collapsed state, which completely suppressed the growth opportunities of eggs and cup handles. The potential calculation of creating a cup based on the cup handle was also not executed, which avoids the invalid calculations. On the other hand, even if the egg and the cup handle have the opportunity to grow first, but without the support of more information and so cannot continue to grow and reach the collapse, the growth of the face will still be executed and eventually collapse. If the definition of cognitive knowledge is reasonable, the final result will be stable even if the intermediate calculation process changes.

**Multi-level calculation**: This example only shows the two-level calculation of concepts such as face and eyes, and the principle of expansion to multi-level calculation and growth is exactly the same. Upward can level-by-level calculate and grow higher levels of concepts such as [person] and [ask] based on the relations such as [person has face] [ask has asker]. While downwards can alsoe from the eyes calculat and grow lower-level concepts such as eyelashes and pupils. In short, according to different known conditions and task targets, the entire cognitive network will select different cognitive components to combine and grow in different directions, and all growth is based on the same probability calculation model.

Traditional intelligent algorithms are usually unidirectional calculations, and the calculations such as parsing and generation are divided into completely different calculation tasks. Pattern recognition is usually defined as bottom-top unidirectional parsing calculations. Recently, the importance of top-bottom interpretation has also begun to gain attention[7], and the view that both interpretations should be taken into account has also been raised[8]. The omnidirectional growth in the DC Net is a more complete and specific solution. Now, we have an omnidirectional growth model that unifies various tasks[1].

We believe that the human brain uses a similar method: the calculation direction is always adjusted and optimized according to the probability changes of the known conditions and intermediate calculation results, and a large number of unnecessary calculations with low probability are suppressed. The reason why the human brain is so efficient for image recognition (obviously more efficient than AI at present) is reflected in these key points.

**Select to use as need**: In the example, the calculation of the face based on the input probability 0.1 of the ear combined with the [face has ear] relation was not executed, because the face has reached the probability collapse based on other information, and let the face top-bottom generate parsing just only according to P(B|A), so that the ears whose input probability should be only 0.1 to finish the probability collapse. The probability-driven omnidirectional growth method realizes the important principle of " comprehensive representation and select to use ".

---

1 In fact, the omnidirectional growth model unified also the discriminative model and the generative model.

Here is a more in-depth discussion of " comprehensive representation and select to use ". Suppose that dozens of concepts are used to describe the concept of cat: head, abdomen, back, left ear, right ear, left eye, right eye, nose, mouth, tail... However, according to the cognitive probability model, it is enough to make the cat reach the probability collapse just by getting the 4 or 5 components' known information. Therefore, only a part of components are needed for the calculation of recognition the cat.

And only a part of components can be used, because it is impossible to obtain all the information of the cat at the same time, and the information that can be obtained at different times is different. Therefore, we believe that it is unreasonable to train a set of weights according to the assumptions of independent and identical distribution and parameter sharing and use fixed network to deal with all situations. A large number of invalid calculations will be generated and the calculation results will be unstable.

The cognitive probability model and growth algorithm of the DC Net are suitable for this cognitive calculation of "select to use as need":
- The conditional probability is based on the assumption of independent sampling, and probability superposition is suitable for selecting part of the information for calculation;
- The dynamic growth algorithm ensures that the required knowledge will only be created and calculated, which is a very important optimization;
- Mechanisms such as probability collapse and probability suppression reduce invalid calculations and achieve more adequate optimization;

Human intelligence itself also follows this pattern. The human brain remembers a large number of knowledge modules, they have different perspectives and are suitable for different scenes. In different situations, the human brain will use different knowledge modules to solve specific problems: not only cat can be recognized based on different body parts; cat can also be recognized based on the sounds they hear; cat can be inferred based on footprints; and even knowledge modules like [ask others "Are you holding a cat?"] To solve the task of recognition cat.

Of course, although only part of the knowledge is needed in each specific task to solve the problem, it is still necessary to construct the entire knowledge system as complete as possible to ensure the efficient selection of correct knowledge in different scenes.

The paradox that "more knowledge has more ambiguity " is also a classic problem of artificial intelligence. Here is an explanation: every knowledge has a scope of application, and there is no knowledge's probability is absolutely equal to 1, and there will be a lot of conflicts the assumptions of absolute applicability and absolute probability of knowledge. Choosing the most appropriate knowledge to solve problems in different scenes is the key to human intelligence and the correct direction of strong AI. This requires each knowledge to have applicability scenes and reasonable probability values, and represent them in a clear structural form. Under such a system, the more knowledge the system has, the smarter it is, just like the performance of the human brain.

### 3.3.4. Reductions

Network growth cannot be carried out unlimitedly. The purpose of growth is to complete the task. Both the task target and actual conditions can define the growth boundary of the network, and the network that has already grown can also be cut and deleted for simplification.

**Cut-off**: Cut-off means grows to a boundary and terminates. Growth in different directions all can be cut-off. There are two situations: one is cut-off because the growth result can meet the task target; the other is insufficient known information to support continued growth, for example, the accuracy of the image defines the smallest concept that can be recognized (of course, if the camera has the ability to zoom in, it can also grow and improve the details incrementally-just like the human eye. Obviously, this must also be around the Dynamic Cognitive Network structure as the center can be realized.).

**Cut**: Cut some of the already grown network to save space. The basic principle is to retain more important information and cut out relatively unimportant information. Generally, the information closely related to the task target and larger-grained information are more important. For example: after the cat is recognized, the low-level concepts such as the head and tail created when the cat is recognized are cut and only the root concept is retained.

### 3.3.5. reasoning and iterative

Combining multi-level network growth and reduction can realize multi-level reasoning and iterative calculation.

In the DC Net, multi-level growth can be carried out in different directions such as upward, downward, and lateral. All growth is broad sense reasoning. Usually time-dependent multi-level lateral relation growth has a long growth sequence, which can be called narrow sense reasoning. All kinds of reasoning are bidirectional. The only influencing factor is probability. Reasoning in the direction with high probability is more effective and priority, and the direction with low probability cannot be reasoned.

Iteration is usually the growth of lateral relations such as change and movement. The tree network structure at both ends of these relations is exactly the same, which only reflects changes in parameters such as time and version. If only the latest state of the tree network is retained for each growth while the old tree network is cut, it will be reflected as a "pure" iteration without retaining the historical version.

### 3.3.6. network circulation

From a macro perspective, the combination of various algorithms forms a comprehensive effect of "network circulation".

- Complete a tree network: Every basic calculation unit completes a tree network. First of all, the tree network is constructed and in a state of probability superposition, and then combined with various information to increase the probability and complete the tree network complement. If finished successfully, the tree network will complete the growth and reach the probability collapse.
- Transferred the task to the new tree network: Starting from the completed tree network, according to the needs of the task, some higher-level tree networks can be grown upward from the root node, or more details tree networks can be grow downward from internal nodes, or other tree networks can be grow laterally.
- Delete unnecessary tree network: If the historical information is no longer needed, it can shrink or completely delete the useless tree network.

Observe the whole process: Driven by task targets and calculation results, new networks are continuously formed; networks are constantly growing; networks change from superimposed state to collapsed state; useless networks are reduced, thus forming a continuous network circulation calculation and fitting The dynamic change of the target scene.

We believe that such a model is relatively close to the way the human brain thinking. The construction of the tree network corresponds to the activation of a group of neurons in the human brain. The deletion of the tree network corresponds to the restoration of the normal state of the neuron, and the network circulation corresponds to the continuous transfer of the activation state of the neuron, that is, continuous thinking.

### 3.3.7. Query

In the DC Net, the realization of information query tasks can be divided into two parts:

**Query matching**: The basic processing of query is to use the query template generated by the problem task as the base network to execute the network matching algorithm in the knowledge network database, and the successful-matching result is output as the query result. This kind of matching usually requires a complete match rather than a partial match, that is, all parameters contained in the query template must be matched and satisfied.

**Conversion reasoning**: Complex query tasks need to process information such as reasoning or even calculation. Specifically, various growth algorithms are selected for multi-level reasoning. Each step of this conversion reasoning can be displayed and explained.

### 3.3.8. External implementation

The semantic implementation of different levels of concepts and relations is divided into two ways: internal implementation and external implementation.

**Internal implementation**: The semantics of various basic concepts and relations in the DC Net form a self-descriptive system. The higher-level internal concepts and relations are described each other, and mutual calculations are realized through built-in basic algorithms. After the system is mature and stable, just only to increase and adjust the cognitive network structure through manual or automatic learning to realize the expansion of the system, instead of changing the algorithm and code of the program.

**External implementation**: To implemented the most basic semantics requires specialized external models and external algorithms. The number of these models and algorithms is limited.

The internal and external implementations need to be connected. The latter provides the most basic semantic implementation support for the former. The boundary between the two is adjusted and optimized according to the development of technology. Examples are as follows:

**Natural language processing**: In DC Net, Natural language processing is not simply regarded as a string-to-string conversion process, but embodied in two parts: the main part is the growth and processing of the internal cognitive network, this part is actually not related to the specific language, it is the general cognitive calculation on each domain: language processing is only a small amount of secondary work, which reflects the parsing and generation conversion between the internal cognitive network and language forms. Among them, language forms such as strings and language roles are also components of the DC Net. Therefore, natural language processing is mainly implemented by the internal algorithm of the DC Net, and the external implementation is relatively simple.

**Spatial scene and image processing**: Spatial scene and image processing are very basic tasks. Among them, designing a scalable and efficient three-dimensional spatial cognition model according to the principle of DC Net is one of the most important tasks.

The basic implementation framework of image recognition and scene reconstruction is as follows: First of all, the external system uses basic algorithms such as convolution and SLIC to process the image to form the superpixels contain characteristic information as elementary elements, then through the network grow to form complete two-dimensional and three-dimensional space scene.

The elementary elements generated by the external system are not required to be absolutely correct (probabilistic collapse state), and overlap and ambiguity (probabilistic superposition state) are allowed. This ambiguity is resolved by subsequent network growth. Specifically, according to the principle of omnidirectional growth and probability priority, more reasonable elementary elements are selected to generate the final scene level by level, and the wrong elementary elements are naturally discarded. In natural language processing, word-level segmentation and ambiguity are solved by the same principle. The growth algorithm of the DC Net can avoid the problem of a unidirectional pipelined processing algorithm that errors in the previous stage are transmitted to the next stage and are amplified and spread.

**Action processing**: Action is also a kind of generation and output. output the action component generated by the network growth algorithm is the execution action. The main point is that the action components need to be decomposed and refined to the granularity that the external action system can accept and realize.

### 3.3.9. Machine Learning

First, make a basic definition of learning. The main point is the relation between knowledge and data.

In the set dimension, DC Net is also a multi-level structure, in which the higher-level network is more abstract and more approaching to knowledge, and the lower-level network is more concrete and more approaching to data.

The application task requires that the high-level knowledge network already exists and used it as a basis and template to grow the derived instance network, which is embodied as a top-bottom growth in the set dimension.

The learning task is to directly create instance network and knowledge network according to basic prior knowledge when there is no correct knowledge network, and to summarize and optimize the knowledge network, which is embodied as a bottom-top growth in the collection dimension.

Cognitive knowledge contains two aspects: the main aspect is the network structure composed of concepts and relations; the secondary aspect is the conditional probability and other parameters contained in the structure. Obviously, "**structure is more important than parameters**". The learning task must first produce an effective cognitive structure, and then adjust and optimize the parameters of the cognitive structure.

Machine learning algorithms are also embodied as network growth algorithms, and are closely

integrated with application tasks. The basic principles are:
1. Perform application tasks: build instance network based on the knowledge network to perform scene fitting.
2. The correctness of the result is judged by calculating the deviation of the instance network and the scene through algorithms such as network matching. The instance network with the deviation below the standard is regarded as the correct result, and the deviation exceeding the standard is regarded as the wrong result.
3. If the correct example network cannot be constructed based on the existing knowledge network, then it is considered that the necessary knowledge network is lacking, and a new knowledge network needs to be created to solve it, that is triggering the learning algorithm.

In the process of the task, if the produces multiple effective instance networks, which one should be used? This leads to an important basic theory.

**Cognitive network optimization principle**: When using cognitive network for information representation, always try to use the optimized network structure for information representation under the premise of ensuring that the deviation reaches the standard. The optimized network can be roughly defined as the network with the smallest dynamic structure, and the size of the network structure can be measured by the total number of concepts and relations contained in the network.

It can be considered that the evolution of human brain structure largely follows this optimization principle. Although the human brain is very powerful, its capacity (the number of neurons and connections) is ultimately limited. To process and memorize approximately infinite information, it is necessary to simplify processing and use the smallest possible structure to memorize as much useful information as possible. This simplified processing will lose some information, but as long as the key information is basically correct and complete, the requirements can be met.

In the DC Net, the tree network and the derived network structure are concrete manifestations of this principle. The tree network embodies the combination simplification of a single concept, and the derived network embodies the generalization simplification of multiple similar concepts.

The principle of cognitive network optimization involves a lot of content. We will not conduct an in-depth analysis here, but will simply illustrate with the aforementioned examples:

A cat is always composed of head, body, paws, tail and other parts. Therefore, the network of [head + body + paws + tail] can describe the information completely. If only consider the integrity of the information, then the concept of "cat" doesn't need to exist at all. However, the size of this network is 4, and every time describing the similar information needs to repeat the 4 elements, which results in waste of information storage and transmission resources. Therefore, the concept of a cat and a tree network [cat: has a head, has a body, has paws, has a tail] are constructed. The size of this network is 5, and define this network consumes extra space, but when encounter similar information in the future, it need only use the concept of "cat" with a size of 1 to represent it to save space. Parts such as head and body do not need to be represented explicitly. They can be calculated based on the knowledge and probability defined in the tree network, and the information is not lost.

The optimized representation comes at the cost of introducing deviations. The definition of "cat" is just a tree network. In order to achieve the optimization effect, it will not memorize the complete information of each cat but record the average information, such as remembering [Cat:

Have Average weight 3 kg]. When representing each cat, there is no problem if the deviation between its actual weight and average weight is within the allowable range. If the deviation exceeds the standard, for example, a cat weighing 50 kg appears, it violates the primary principle that deviation needs meet the standard, and should be dealt with, that is, to learn and create new knowledge, for example: to derive and create a new breed of "giant cat".

The optimization of human language communication on the basis of ensuring the effective transmission of information is one of the basic principles of linguistics. The principle of cognitive network optimization shows that the internal cognitive description also follows the similar optimization principle, and is more basic and essential.

The principle of cognitive network optimization is one of the design principles of cognitive network structure, and it serves as an important prior knowledge and goal constraint for machine learning. Cognitive network learning based on this principle is significantly different from end-to-end fitting learning:

- **End-to-end fitting learning**: set a relatively fixed initial structure; use algorithms such as the back propagation to adjust parameters to make the calculated residuals smaller and smaller; learning to pursue the minimum residuals as the goal;
- **Cognitive network learning**: it does not set a fixed structure, but learns different structures to fit the scene; the deviation is not used as the fitting target but as basic principle for each step of the calculation, that is, the deviation is not allowed to exceed the standard; learning to pursue the best optimal structure as the goal;

Among them, the specific definition and calculation of deviation are related to specific domains and tasks. For example: for natural language processing, the deviation is mainly reflected in the deviation between the actual character string and the target character string; for image recognition, the shape, size, texture, etc. are the main indicators of deviation calculation. All kinds of deviations are finally converted into membership degrees for unified evaluation and integrated with the probability calculation system.

As mentioned earlier, the main difficulties faced by structure learning is the problem of how to define the target structure and the requirement of differentiable knowledge. In the DC Net, these difficulties can be solved in basic theory:

1. The basic structure such as the tree network is defined, and the knowledge learned must conform to the basic structure. It self is effective prior knowledge and constraints.

2. The principle of cognitive network optimization provides an important evaluation basis and fitting direction for the learned cognitive structure.

3. Structural learning does not require differentiability. Differentiable parameter learning is only an extension of structural learning.

4. It can be combined with various prior knowledge in specific domains including human input knowledge.

The integration of the above principles lays the foundation for the realization of structure-oriented cognitive network machine learning.

**Cognitive network learning algorithm (CNL)**: The basic framework of a cognitive network learning algorithm is designed based on the above theory. The algorithm assumes that the known knowledge is mixed with the knowledge which need to be learned, and the known is used to learn the unknown. The typical implementation steps are as follows:

- The learning algorithm is centered on the Dynamic Cognitive Network and is carried out on the domain dimension and the set dimension at the same time, and realizes learning around the tree network and the derived network structure respectively.
- 1. Initial state: There are already multiple instance concepts $S_{di}$ (for example: super pixels in image processing) provided by the previous level processing on the lower i-layer in the domain dimension. At this time, it does not need to ensure that these instance concepts are absolutely correct (probability=1), but try to get all possible instance concepts, each with a certain probability.
- 2. Execute growth: Perform network growth according to the requirements of the application task. If the growth is successful, that is, match the base network $S_{bj}$ on the j layer and derive the instance network $S_{dj}$ to grow and absorb $S_{di}$, Which means that the existing knowledge can solve the encountered data without learning new knowledge.
- 3. Create hypothetical knowledge: If the above growth fails, the reason is that the j layer lacks the necessary base network, so try to create a new base network Sbj: The specific method is to select part of the Sdi and combine the necessary prior knowledge (adjacent relation in image recognition Is a very important prior knowledge) to create a new knowledge tree network $S_{bj}$ and instance tree network $S_{dj}$. Then includes necessary longitudinal and lateral relations. In the end, multiple $S_{bj}$ may be generated, all of which are hypothetical new knowledge.
- 4. Initialization parameters: firstly, mark the new knowledge of each hypothesis as [new knowledge]; then initialize its parameters and probabilities. The default setting is to use the parameters of each instance concept $S_{di}$ to assign values to the parameters of the corresponding concept of $S_{bj}$, and the conditions probability is initialized to 1.0, and the statistical quantity is initialized to 1. If there is applicable prior knowledge, the parameters and probabilities are calculated and assigned based on the prior knowledge.
- 5. Continuous growth: add the hypothetical new knowledge network $S_{bj}$ and its derived instance network $S_{dj}$ into the system, and then jump to 2 to continue network growth until all $S_{di}$ are absorbed and grown into one or more complete instance networks $S_d$.
- 6. Result select: Select the network that has reached the collapsed state as the final result from all the instance networks $S_d$. If multiple $S_d$ are collapsed, the smaller $S_d$ is selected as the final result according to the principle of cognitive network optimization. If the final result is determined, the new knowledge network it contains is also determined.
- 7. Statistics accumulation: The new knowledge created by learning can be applied immediately after being added to the system. The execution of more subsequent tasks will trigger the application of these knowledge, and the statistical number of each new knowledge successfully applied will be recorded. According to statistics, quantitative parameters such as conditional probability can be calculated, new knowledge with a larger conditional probability is finally determined to be the correct knowledge retention, and knowledge with a small bidirectional conditional probability is discarded.
- 8. Merge similarity: Compare the newly learned knowledge with the existing knowledge on the set dimension, merge the similar knowledge and optimize the structure according to the principle of cognitive network optimization.

To summarize the core points of the CNL algorithm are:
- Try to use existing knowledge to interpret data and execute growth, and require that the deviation does not exceed the standard.
- When the explanation is unsuccessful, follow the principle of **"deviations and errors mean that new knowledge is discovered"** and execute **"create new knowledge based on data"**.
- In the process of creation and application, remove unreasonable knowledge and leave the most reasonable knowledge according to constraints such as basic structure definition, cognitive network optimization principles, domain prior knowledge.
- More reasonable knowledge was selected through statistical counting, and adjust parameters such as conditional probability.
- Merge and optimize knowledge.

The CNL learning algorithm also has these characteristics:
- **Small sample learning**: **"Single sample learning structure, multi-sample learning probability"**, the structure can be learned from a single sample, and multi-sample learning is to perform statistical calculations on the learned structural components and adjust parameters such as probability. For a large amount of knowledge with probability$\approx$1 (for example, cat has cat's paw, $P(A|B)\approx 1$, $P(B|A)\approx 1$), the knowledge learned by a single sample is accurate enough, and a large number of samples are not required at all.
- **Real-time learning**: learning and application are always carried out at the same time, the two will not be absolutely separated. In fact, always applications' requirements promote the learning of new knowledge.
- **Independent parameter learning**: The value of the concept reflects the division of the concept itself; the probability of the concept reflects the statistical frequency ratio of the concept, and the boundary between the two is clear. The probability model is based on the assumption of independent sampling. Each knowledge and conditional probability are independent of each other, and can be independently learned and adjusted independently. In contrast, many machine learning regard parameters as a overall black-box structure, and the adjustment of parameters will entangle and affect each other.
- **Incremental learning**: Learning always interprets and learns new knowledge based on known knowledge. The ideal situation is to learn a small amount of knowledge each time and learn incrementally. For example: embodied in the above algorithm, i is a relatively low level and does not need to be the lowest level 1, which means that if the lower level has correct knowledge and can correctly interpret to the i level, the new knowledge only needs to be learned at level j=i+1. Other, if the new knowledge learned in layer j can be explained and grown by existing knowledge in layer k, there is no need to continue learning on layer k. In addition, the complete information of a tree network is gradually constructed through incremental learning. For example, the first time seeing a cat from the left, We built a tree network of cat and learned about its left ear, Left paw and other knowledge; the second time seeing a cat on the right side, we can use existing knowledge to add new knowledge about its right ear and right paw.

Obviously, these characteristics are also in line with the way humans learn knowledge.

Undoubtedly, implementing powerful unsupervised machine learning is an arduous task. The basic principles of machine learning oriented to dynamic cognitive structures are described above. The specific implementation requires a lot of detailed work.

### 3.3.10. Continuous calculation

Complex intelligent tasks need to combine a variety of different subtasks and algorithms to achieve. The algorithm system of the Dynamic Cognitive Network has a strong multi-task fusion ability. The cognitive network that fits the scene is shared by each subtask as the central structure, and exists in the entire life cycle of the overall task or even longer, and each subtask does not dock each other but separately docking with this central structure and cooperating with it for various processing. Various sub-tasks can be dynamically organized and performed continuously, and even adjusted and added tasks at any time. For example:

- In the space and material structure generated by image recognition, focus and enlarge the details, or extend the scope to carry on splicing recognition.
- Carry out generation image and creations tasks on the space and material structure generated by image recognition.
- Different tasks such as translation, dialogue, and sentiment analysis can be dynamically implanted in the dialogue process.
- Recall historical tasks and resume execution seamlessly.

The completeness of the cognitive network constructed in the process reflects the scope and depth of the target scene and directly determines the intelligence level of the task. For example: In a customer service scene, the processing goal is not limited to the words of both parties, but should be expanded to a complete cognitive network describing the entire scene. This cognitive network continues to grow in the process and integrates all entities, dialogue, reasoning, and thinking, so that the dialogue can be truly understood and handled accurately. At the same time, it also lays the foundation for the realization of flexible and comprehensive intelligent tasks: not only can the machine act as a customer service staff, but also can use the same method and the same knowledge to achieve scene understanding, information collection, perform translation, and act as a customer assistant.

Continuous calculation can not only realize powerful intelligent functions, but also effectively improve performance. For example: Contrary to the current AI system, human recognition of dynamic scenes is more efficient than the recognition of static images. The reason is that the recognition of each frame of image in a dynamic scene does not always start from scratch, because the cognitive network containing a large amount of collapsed information has been identified and constructed. Processing a new frame of image only requires a small amount of incremental processing on the already explicit cognitive network, which is extremely efficient. If the AI system can use the same method, the completeness, accuracy, and performance of its tasks can be expected to catch up and even surpass humans.

Obviously, a strong AI which is comparable to human intelligence must also be able to perform continuous tasks, not fragmented intelligence. The analysis of continuous computing further proves that the central structure that independent from computing and dynamic grows is the key to achieving strong AI! In the face of open domains, this central structure must be a cross-domain general cognitive structure similar to the DC Net. It is difficult to achieve the same goal with dedicated data models and fixed network structures for specific calculation tasks.

## 4. In depth discussion

In this chapter, we further analyzed and discussed that several key points of Dynamic Cognitive Network, and the similar logic structure and mechanism that might possibly exist in human brain.

### 4.1. Tree network is the basic structure of cognition

According to the previous analysis, the tree network, which combines both the problem decomposition capability of the tree and the comprehensive information representation capability of the network, turns out to be a basic structure preferable for cognitive representation. Any complex scene can be fitted by flexible composing and nesting of various tree networks.

On this basis, certain principles of the human brain operating mechanism can be inferred accordingly. From the abundant research findings about human brain structure, it can be seen that the human brain neurons are organized in a hierarchical structure, having connections on upwards, downwards, lateral, and various other directions. However, how these structures and information flows constitute a complete system of thinking is still unclear. Targeting at this problem, we put forward a hypothesis: human brain also adopts abundant of logic structures similar to the tree network to conduct cognitive representation and calculation.

Though the elements and mechanisms in human brain accurately corresponding to the concepts and relations in DC Net are still not clear, one thing for sure is that the human brain uses at least a group of neurons (e.g. vertical column) instead of a single neuron to represent a concept or relation. If we call such a group of neurons a neural unit, the structure of a certain tree network (e.g. face) can be represent as below:
- Use one neural unit to represent the high-level concept of the tree network—such as the face, while using other neural units to represent the low-level concepts of the tree network—such as eyes and nose.
- Use the top-bottom longitudinal connection to connect the high-level neural unit with the low-level neural units; meanwhile, use the bottom-top longitudinal connection to connect the low-level neural units with the high-level neural units as well.
- Neural units of a same hierarchical level contain a lot of lateral connections—for example, the lateral connections representing the adjacent relations of eyes and nose.

If the hypothesis that the top-bottom tree network structure is the main priori structure of the human brain's description and cognition to the world is tenable, its basic learning mechanism can be speculated as below: human brain has prepared a lot of prototypes of tree network structures, including various potential connections, thus to learn by establishing and strengthening correct connections whenever encountering new knowledge. And the goal of learning is to form tree network structures describing various specific knowledge.

The adjacent relation represented by short-distance lateral connections plays a very important role in the tree network structure. And as a matter of fact, this short-distance adjacent relation is a kind of basic priori knowledge. The human brain cognition tends to combine adjacent concepts together. This is true for both space and time. The adjacent relation in space corresponds to the static structure construction, and the adjacent relation in time corresponds to the causal relation construction. As for the learning, the adjacent relations among low-level concepts often trigger the building of lateral adjacent relation, and then, it can be inferred based on the priori structure hypothesis of the tree network that: the occurrence of the lateral relation indicates the existence of corresponding longitudinal relation and the whole tree network, as a result, the high-level concepts and longitudinal relation building could be triggered at the same time to constitute the while tree network structure all at once.

Therefore, the tree network is the basic unit for human brain to learn and apply knowledge. Specifically, knowledge learning is to build a group of connections relating to a tree network at the same time; and the knowledge application is also conducted by taking the tree network as a unit: each tree network tries to activate itself and compete with other tree networks, and those finally win the competition would be composed and nested and formed into correct scene representation.

### 4.2. Bidirectional connection of generate as main

The tree network is composed of concepts and bidirectional relations. As for bidirectional relation, though both of its unidirectional connections can be conducted with probability computing, they have substantially different effects on tree network. For the longitudinal relation connecting high-level concepts and low-level concepts, the top-bottom A=>B connection (also called the generated connection) is the main connection, while the bottom-top B=>A connection (also called the parsing connection) is the auxiliary connection. The reasons are as below:

- **The generated connection describes the essence of the tree networks:** tree network is always organized from top to bottom, It takes the A=>B connection from high-level concepts to low-level concepts and the parameters like conditional probability to represent the knowledge of the tree network itself. Such knowledge is non-related to other tree network. A tree network has multiple compatible A=>B connections. And these connections usually have comparatively large conditional probability. On the other hand, a low-level concept may be parsed into multiple high-level concepts and having multiple B=>A connections. However, this is not tree network structure because these B=>A connections mutually exclusive and each connection usually has comparatively small conditional probability.
- **The downward generating is far more than the upward parsing:** As mentioned above, a parsing task is usually composed of two aspects: one is to parse high-level concepts upwards according to part of the low-level concepts. And the other is to conduct generative parsing downwards according to the high-level concepts in order to infer all low-level concepts. One high-level concept corresponds to multiple low-level concepts—for example, the cat concept in the previous example is composed of 100 components. And according to the aforementioned principle of probability passing, the upward parsing of 1 component would trigger the downward generative parsing of the high-level concept "cat" to the rest 99 components; furthermore, it only needs a little part of the low-level concepts to cause high-level concept probability collapse，leaving the follow-up work achieved by downward generating from high-level concept. For example, 5 low-level concepts can be applied to achieve high-level concept collapse, while the rest of 95 low-level concepts are all generated downward from the high-level concept.
- **The downward generation has higher efficiency**: Suppose the high-level concept A has

collapse probability 1, The upward parsing of low-level concept B has collapse probability 1 then it should be the correct target to downward generate 100 components to complete the entire tree network. In this case, every step of the growth is effective. But for upward parsing of low-level concept B, even if B has collapse probability 1, it has only one correct connection among N connections with possible high-level concepts, while the rest N-1 connections are incorrect and ineffective. Therefore, even for parsing tasks, the top-bottom generating calculation is adopted more frequently. At the initial phase, certain amount of upward parsing calculation would be conducted from low-level concepts. Once the high-level concept with certain probability is inferred, it would as soon as possible turn to "top-bottom" generation to achieve the overall parsing task.

Therefore, the top-bottom generating connection is the main connection describing tree network, and it is also the main body of calculation, while the bottom-top parsing connection is an auxiliary structure. As for the growth of the whole tree network, most of the work is done by top-bottom generating, having only a small part of work is bottom-top parsing.

Similarly, we proposed hypothesis and analyzed the structure and mechanism in human brain achieving relations similar to bidirectional relation. We believe that the human brain also adopts a large number of bidirectional relations similar to those in Dynamic Cognitive Networks. But since the physiological mechanism determines that the information transfer of human brain neuron basically uses unidirectional information channel, the information is always transmitted from the axon of a neuron to the dendrite of another neuron (though researches have proved the existence of back-propagating action potentials towards the cell body along axon[9], we believe that it is not a main mechanism). Therefore, to realize bidirectional relation of two neural units, two independent unidirectional information channels are needed. As a matter of fact, the information channel is featured in very complicated structure. Ignoring the complicated details, we can roughly summarize as: one branch of axon of neural unit A connects directly or indirectly to dendrite of neural unit B, and the branch of axon of neural unit B connects directly or indirectly to dendrite of neural unit A. the connection strength is approximately corresponding to the conditional probability of relation.

In contrast, the fact that DC Net uses only one relation to carry the bidirectional conditional probability between two concepts indicates the more competitive aspect of the computer over the human brain. Once making breakthrough of the singular point of intelligence, computer would show greater potential than human brain in many aspects.

The bottom-top connection in human brain is called the feedforward connection traditionally, while the top-bottom connection is called the feedback connection. This definition should only focus on the partial function product of parsing external perception information into the internal cognition. And from the perspective of this view, the feedforward connection is considered to be the main body of information transmission and calculation, and feedback connection only only plays auxiliary adjustment role. Currently, multilayer neural network is also achieved based on this theory, it describes only the unidirectional feedforward connection and weight, lacking description about the more important feedback connection and parameters (back-propagation algorithm only adjusts weights along reverse direction of feedforward connection) .

It can be known from the above analysis that: the top-bottom A=>B connection which is called the feedback connection is the main body playing more important role, and the B=>A connection which is called feedforward connection is the auxiliary part. The fact that human brain has more feedback connections than feedforward connections could be the proof of the above conclusion.

Meanwhile, abundant of research findings also show that, the scenes saw by human brain are mostly generated or "imagined" instead of "saw". This is consistent with the above viewpoint.

Moreover, the viewpoint that takes feedforward as the core is the conclusion of only the peripheral function targeting at preliminary information parsing. The more advanced functions of think, plan and action directly indicate that tree network top-bottom and lateral directions growth is the main calculation method of human brain. In fact, even without any feedforward input, these functions can run as well—for example dreaming and meditation.

If above hypothesis is tenable, it seems more rational to take A=>B connection as the "main connection" while taking B=>A connection as the "auxiliary connection" from the perspective of substantial effects on cognition.

Lateral relation and longitudinal relation have no substantial difference. For certain lateral relations, their two unidirectional connections can be deemed as one main connection and one auxiliary connection, while for some other lateral relations, their two unidirectional connections can be deemed as two main connections.

### 4.3. It simulates the world comprehensively, accurately and efficiently

We can believe that the ultimate target of intelligence is to apply the limited storage and computing resources to simulate the world as comprehensively, as accurately and as efficiently as possible.

The world is an integrated system that influences each other and evolves dynamically. The role of force, which is the basic factor that drives the operation of the world, always interacts with each other rather than exert unidirectional influence. A more accurate description of the world calls for a comprehensive consideration of various interactions so as to simulate and calculate the entire system.

Imagine a physical system that is constituted of various celestial bodies, the gravitation of any two celestial bodies will generate mutual influence to each other. even if we are only concerned with the state (position, motion) of one celestial body, **But, in order to calculate its state, the states of other celestial bodies and the entire system must be calculated as well!** Therefore, the following numerical simulation methods need to be used for solutions:

- Solving for one celestial body: its state is solved based on the gravitation of each celestial body towards one celestial body. This is a unidirectional multiple-variable function calculation whose foundation is the bidirectional influence of gravitation between any two celestial bodies.
- Solving for each celestial body: the above-mentioned unidirectional function is used to solve the state of each celestial body respectively, and when the solutions are combined, it is to solve the entire system.
- Continuous iterative calculation: it carries out the difference processing to the time to iteratively solve the state of the overall system of the next moment, and modifies it at any time according to the updated information.

This numerical simulation method with the aim of evolution of whole system is the peak of tool calculation of human inventions, which is applied to numerous fields to solve complex problems(such as: Climate prediction, Studying the evolution of the universe, Nuclear test simulation). The intelligent cognitive calculation of human brains and this kind of numerical simulated calculation are very alike! Compared with professional system targeting at specific task, the aim of intelligent cognitive calculation is more comprehensive simulation and calculation. Facing the complexity of the world and the limitedness of available resources, a balance is needed in terms of comprehensiveness, accuracy and efficiency so as to extend the model:

- Realizing the hierarchy system: it constructs the multi-hierarchy concept and structured system. It is more efficient to represent and calculate at the most appropriate level according to the differences of task, such as calculating each celestial body as a single variable without calculating each atom.
- Extending the relation system: since it is of excessively low efficiency to calculate the complex world based on basic forces, various relations (such as: have, adjacency, derivation, emotion…) are created to represent and calculate. Such relations can be seen as forces in the broad sense, which exert influence on the concepts on both ends. The various influences are all reflected in the mutual calculation of binding probability.
- Expanding the boundary dynamically: in the intelligent cognitive calculation, the scope boundary of target system is not fixed without changes, but dynamically expanded and adjusted in line with the needs.

We believe that the magic evolution of human brain forms this mechanism that systematically simulates the world, in which bidirectional connection channel provides mutual calculation to concepts and forms the foundation of system. Multiple mechanisms of M-P models make up function calculation modules kin to conditional probability and probability overlay. The tree-like net structure realizes the definition and description of each sub-system that can be combined and nested. Compared with the nervous system of primitive organism that directly proceeds the output reflex behavior simulated by input, the emergence of mechanism that can simulate scene system with cognitive structure is the important sign for human brains to possess advanced intelligence.

For another, the comprehensiveness and integrity of simulated world is limited, and efficiency must be given consideration to while the task targets are met. More effective concepts and relations need to be chosen to construct the system for calculation based on task targets and actual scenes, and other unnecessary factors are ignored for simplication.

The key is to ensure the correctness of this simplication. In an open world, problems will always occur to fixed simplified assumptions. The simplified assumption that is correct in one scene may be completely mistaken when it is placed in another scene. This is where the fundamental problem lies that assumptive theories such as independent and identical distribution are not suitable for intelligent computing in real scenes.

Correct simplication must be both dynamic and intelligent. In one situation, if the influence of one factor on the target of computing task is not large (represented by a small probability), it can be abandoned and simplified. In another situation, the same factor that may have a larger influence on task target cannot be ignored and must be incorporated into the system.

As a result, intelligent calculation is both scene fitting calculation with the whole system as target, while the target system itself is not fixed but dynamically created. **In essence, each intelligent calculation is creating a unique system (rather than a calculation of a fixed system)!** Once the system is correctly created, the calculation task is basically completed. In creation, the main basis for selection or simplification of different cognitive results is the estimation of deviations and probability, and new cognitive structure can be created (that is learn) in necessity.

That the specific implementation mechanism the human brains construct the system dynamically based on the on-demand selection principle is to be studied. But we believe that its logic effect can be explained through analogy with theories such as probability collapse, probability priority, probability suppression etc.

## 5. Conclusions

Human brains see the world in a structured way; intelligence is to describe and fit the world by constructing various cognitive structures. The realization of strong artificial intelligence cannot avoid the fine design of rich cognitive structures in each domain. Traditional system or weak artificial intelligence system mainly partially deal with lots of scenes (small knowledge, big data) by adopting a small amount of limited-domain knowledge. Stronger artificial intelligence systems first achieve comprehensive processing of each scene (big knowledge, small data) with massive wide-domain knowledge before expanding the complexity and quantity of target scene. Finally, systems with powerful computing capability will comprehensively process a large number of scenes (big knowledge, big data), which reflects the super capabilities merging the human brains' flexible intelligent processing capabilities and computers' powerful iterative computing capabilities.

To construct this wide-domain, complete and flexible structured system is the fundamental difficulty to the realization of strong artificial intelligence. However, the key to solving this difficulty is to construct a basic system that has generalization ability and combination ability for various structures first. The DC Net model proposed in this paper has come up with a feasible scheme, in which the two-dimensional multi-level structure is an important mechanism: the tree-like network structure on domain dimension reflects the ability of dynamic combination of many structures in wide domains to realize scene fitting; the derived network architecture on set dimension embodies generalization processing of abstract hierarchical relations between knowledge and data and provides structural support for computing integration of application and learning.

On the basis of this fundamental system, by continuously constructing and improving various cognitive structures on different domains, the intelligence of system will be enhanced continuously and effectively, which will be the main work in the future.

Probability is an indispensable component of structural description. Based on conditional probability and probability plus formula combined with bidirectional conditional probability description, network probability state, probability collapse theory and other theoretical methods, DC Net supports probabilistic description and processing of intelligent systems under open domain scenes and significant probability. Furthermore, with specific implementation, it discusses the feasibility of simplified probability overlay formula. The refined perfection of similar schemes and partial extension on the probability model are where the future important topics lie.

DC Net adopts network growth algorithm to unify various major intelligent computing. Such algorithm system follows the ideas of comprehensive representation and on-demand selection, constructs cognitive networks dynamically with omnidirectional network growth driven by probability, which reflects the intelligent computing capabilities such as arbitrary transformation of input and output, continuous calculation and Interpretability, and is more suitable to be the fundamental algorithm to achieve more powerful AI compared with unidirectional function calculation.

Cognitive network learning algorithm is machine learning algorithm with structured learning as primary and parameter learning as auxiliary. This algorithm takes data that cannot be fitted in the fitting process as signs of discovering new knowledge and carries out learning by transforming such data into new knowledge combined with the priori knowledge to complete scene fitting. This incremental learning method that explains with the known and learns the unknown accord with the cognitive pattern of mankind, which will finally achieve completely unsupervised learning. But this process needs to be realized step by step, thus the manual-oriented construction of basic priori knowledge and algorithm is still the fundamental work at the moment.

To describe the complex world requires plenty of rich structures, but the structures are to be optimized as far as possible. The cognitive network optimization principle proposed based on DC Net can be viewed as the application and extension of information theory to the open cognitive domain. An important task in the future will be factors such as task target, deviation calculation, probability calculation, structure size are combined together to form a comprehensive evaluation system, which quantizes basic cognitive methods like concept division, structure construct, structure optimization etc.

The world is an integrated system in which comprehensive interactive effect relations exist among things. Function representation calculation extracts partial relations and defines them as curing systems to calculate partial variables, which is a partial simulation of the world. Scene fitting calculation selects or creates more effective relations based on needs to construct systems dynamically and to calculate the whole system, which is a more comprehensive simulation of the world. This is the method that human intelligence adopts, and it also should be the principle that strong artificial intelligence needs to follow. It is the inevitable course that traditional tool computing goes through in its evolution into the future intelligent computing.

**References:**

[1] Peter W. Battaglia, Jessica B. Hamrick, Victor Bapst, Alvaro Sanchez-Gonzalez, Vinicius Zambaldi, Mateusz Malinowski, Andrea Tacchetti, David Raposo, Adam Santoro, Ryan Faulkner, Caglar Gulcehre, Francis Song, Andrew Ballard, Justin Gilmer, George Dahl, Ashish Vaswani, Kelsey Allen, Charles Nash, Victoria Langston, Chris Dyer, Nicolas Heess, Daan Wierstra, Pushmeet Kohli, Matt Botvinick, Oriol Vinyals, Yujia Li, Razvan Pascanu. Relational inductive biases, deep learning, and graph networks. [2018]. https://arxiv.org/pdf/1806.01261.pdf

[2] Gary Marcus. The Next Decade in AI: Four Steps Towards Robust Artificial Intelligence. [2020]. https://arxiv.org/abs/2002.06177

[3] Judea Pearl, Dana Mackenzie. The Book of Why. Allen Lane,2018.

[4] Gary Marcus. The Next Decade in AI: Four Steps Towards Robust Artificial Intelligence. [2020]. https://arxiv.org/abs/2002.06177

[5] Gary Marcus.bengio v marcus and the past present and future of neural network models of language. [2018]. https://medium.com/@GaryMarcus/bengio-v-marcus-and-the-past-present-and-future-of-neural-network-models-of-language-b4f795ff352b. https://arxiv.org/ftp/arxiv/papers/1801/1801.00631.pdf

[6] Geoffrey E. Hinton,Alex Krizhevsky,Sida D. Transforming Auto-Encoders. Artificial Neural Networks and Machine Learning,ICANN 2011,21st : 44-51.

[7] WangSara Sabour, Nicholas Frosst, Geoffrey E Hinton. Dynamic Routing Between Capsules. [2017]. https://arxiv.org/abs/1710.09829

[8] Gary Marcus, Ernest Davis. Rebooting AI: Building Artificial Intelligence We Can Trust. Pantheon,2019.

[9] Nicola Kuczewski, Cristophe Porcher, Volkmar Lessmann, Igor Medina, Jean-Luc Gaiarsa. Back-propagating action potential. Communicative & Integrative Biology, 2008, 1:2: 153-155.

**Note appended**: The original version of this paper is written in Chinese, and there may be deviation in the translation process. Therefore, the Chinese original version is attached for reference. If there is any difference between the two versions, the Chinese original version shall prevail.

# 以场景拟合为中心的 AI 与动态认知网络

陈峰[1]

**摘要**

  论文简要分析了 AI 主流技术的优势和问题，提出了：要实现更强大的 AI，必须改变端到端的函数式计算而采用以场景拟合为中心的技术体系。并论述了名为动态认知网络模型(DC Net)的具体方案。
  论文论述了:用概念化元素构成的富连接的异构动态认知网络对综合领域的知识和数据进行统一表达；设计了两维度多层级的网络结构对组合、泛化等 AI 核心处理进行统一的实现；分析了开放域、封闭域以及显著性概率、非显著性概率等不同场景下计算机系统的实现差异，指出了在开放域和显著性概率场景下的实现是 AI 的关键，并设计了结合双向条件概率、概率传递和叠加、概率坍缩等理论的认知概率模型；设计了目标和概率驱动的全向网络匹配-生长算法体系对解析、生成、推理、查询、学习等计算进行一体化实现；提出了认知网络最优化原理，并设计了结构学习为主参数学习为辅的认知网络学习算法(CNL)的基本框架。
  论文也对 DC Net 模型和人类智能在实现上的逻辑相似之处进行了对比分析。

**关键字：** 动态认知网络；场景拟合；全向网络匹配-生长；派生网络；树形网络；认知概率模型；认知网络学习；认知网络最优化；双向条件概率；

## 1. 引言

  近年来，以深度学习为代表的一系列人工智能技术取得了显著进展。然而，人们都认识到实现强人工智能仍需要长期的努力，并在不同的技术路线上进行着积极的探索。
  在本文中，我们对人工智能的一些关键问题进行了分析，并提出了解决问题的要点。

  我们认为，现今，数据和硬件性能已经不是 AI 木桶上最短的木板，简单堆砌数据量和硬件计算能力对智能水平的提升将极为有限。关键是要在基础模型和算法理论上取得突破，其中，"组合泛化一定是 AI 实现人类能力的首要任务，结构化表示和计算是实现这一目标的关键[1]"。
  我们提出并强烈建议的核心理念是：**要实现更强大的 AI，不能只依托以固定输入输出为目标的端到端的函数式计算，而应该主要采用符合认知本质的以场景拟合为中心的计算。**为此，需要设计一种全新的具有综合领域场景描述能力的认知结构，并以对该结构进行动态生长的方式来进行场景拟合并实现各种智能任务。
  基于这种理念，我们赞同"对经典人工智能的三个关注点—知识、内部模型和推理需要重新深入思考，希望用现代的技术手段以新的方式解决这些问题[2]"，并做了大量的研究工

---

[1] 陈峰(1973-)，男，独立 AI 研究者，研究方向：AI 基础理论，自然语言处理，email：chenfeng@nengsike.com

作。

显然，构建核心的基础知识体系比起构建外围的数据知识更为重要，前者是解释后者的基础。这种知识应该是概念化和可解释的通用知识，具有全向灵活应用的能力，而不是黑盒式的或只适用于单一任务处理的专用知识。

在构建知识之前更基本的问题是承载这些知识的内部模型如何定义？这个最经典的 AI 难题现在必须得到解决，这是其它工作的基础。我们认为，基于网络的知识表示是正确的方向[3]，但除此以外还有很多同样重要的原则需要明确并实现。其中：遵循概念化的基本原则，单一的基本结构定义，跨领域的表达能力，灵活组合无限应用的能力，实时动态变化的能力，合理的概率模型，学习和应用一体化等都是必须具备的重要特点。

我们也赞同，因果关系的知识表达比起相关性表达更为重要[4]。但同时要看到，因果关系和相关性关系的划分是相对性的，并且因果关系只是描述世界所需要的众多关系中的一部分。我们认为：应该构建更完备的关系体系并在不同场景下灵活选择使用，并且遵循本质关系优先于相关性关系的原则[1]。

推理是对知识进行动态应用的重要能力。但同样也要看到，推理只是完整的智能算法体系的一部分，这套算法体系应该紧密围绕上述知识体系用统一的方式实现解析、生成、推理、查询、学习等算法并将它们融合在一起。

总之，更有效的知识是智能系统的关键，新一代的智能系统必须在设计更强大的知识表达结构、构建更有效的知识体系、和实现对知识更充分而灵活运用的算法体系这三个关键问题上取得重要的突破，才能具有可解释性、稳定性、通用性等方面的能力，向更强的人工智能演进[5]。

我们还认为，现在还没有到达可以全面地追求自动机器学习的阶段。自动机器学习是极具吸引力也是终极的目标，但必须在一个强大的基础模型上才能学到有效的通用知识。现阶段，这个基础模型还主要需要人工设计才能构建起来，然后将可以实现比现有方法更为强大的自动机器学习—这种学习一定是以结构学习为主，而不限于对权重等参数进行调整，并且可以实现无监督学习和实时增量式学习。

## 2. 基本分析

### 2.1.现状分析

以下先对几种主流技术的主要优点和问题进行分析，然后探讨更为合理的解决方案。

### 2.1.1. 符号主义方法

符号主义方法的正确性和优点主要体现在：
- 用经过提炼归纳的概念化符号来描述知识这个原则毫无疑问是正确的,符合人类认知思维和知识处理的本质。概念化的知识具有可解释、易维护、使用灵活的特点，能充分体现知识应有的价值，人类积累的大量知识也主要通过高度概念化的符号来表达和传承。

但一直以来，符号主义方法在实现强人工智能方面进展较慢，体现了其有待完善的问题：
- 现有符号系统的基本结构的描述能力不够。代表性的知识图谱体现出了用网络结构来表达知识的正确方向，但结构仍然过于简单，不够完备和灵活，难以实现更完整的认知体系。

---

1 完备的关系体系包括空间、因果等各种关系。本质性和相关性的划分是相对而非绝对的，本质性关系比相关性关系更直接，一个相关性的关系通常可以分解为多个更本质的关系。

- 已有的知识体系通常重定义缺解释。偏重于管理大量数据性知识，缺少对基础知识和常识的深度表达，无法支撑多层级的智能计算。
- 很多知识系统的设计和表达形式过于简化和固化，失去了智能计算必须具有的动态性和灵活性。
- 缺少合理的概率体系，各种符号系统的设计倾向于处理确定性的知识，对于更为广泛的具有不确定的知识没有构建起完整的描述和计算体系。
- 对于分类、知识、数据、实例、规则、泛化、推理、应用、学习等基本概念以及它们之间的相互关系分别采用不同的表达和实现手段，不能归纳到一个体系下进行处理。
- 缺少和符号化知识体系紧密结合的通用智能计算模型。

### 2.1.2. 连接主义方法

连接主义[1]的典型代表方法—深度学习近年来在一些领域成效显著，主要体现了以下正确方向：

- 网络结构的描述能力巨大，知识表达需要具有足够的连接密度的网络来实现。
- 运用权重这样的连续值来实现不确定性的描述和计算，证明了不确定性处理对于智能系统的重要性。
- 展现了自动机器学习的巨大价值和潜力。

同样，现有的连接主义和深度学习技术也存在多方面的问题，在可解释性、稳定性、灵活性等方面更存在总所周知的缺陷。其关键原因如下：

- 连接主义的内部网络元素非概念化，这是其缺乏解释性等能力的根本原因。这种网络中蕴涵的知识体现为针对特定输入输出进行拟合的高度专用的结构和参数，不具备概念化的知识应有的通用描述能力，难以实现对知识的灵活运用—例如组织高级的形式化推理计算。
- 因为元素非概念化，所以采用了实用主义的权重来处理不确定性，不能体现真正的概率体系具有的可解释性和独立调整能力。
- 仅依赖过于简化的结构来描述复杂的知识，例如仅依赖于一种 M-P 结构。实现 xor 这样的基本关系都需要采取复杂的多级结构来强行模拟并且不能彻底解决问题。更合理的方案是引入专门的 xor 关系来解决，连接主义的理论的源头—人脑就是这样做的，当然，这就意味着要摒弃限定单一结构而要采用多种关系类型的异构网络结构。
- 追求端到端的表面拟合而极度简化中间模型的设计，适合于单向函数式转换处理，而不能实现需要全向计算的灵活的智能处理。针对不同任务甚至细化的子任务都需要训练不同的网络，所以衍生了众多类型的网络却难以灵活地融合在一起解决真实场景面临的复杂问题。
- 上述基础上实现的机器学习难以学习结构知识，主要采用对权重等参数的调整来模拟结构(因此通常只能对可微的知识进行学习)。实际上，对于认知表达来说结构远重于参数，认知学习必须将结构学习作为基础目标。

---

1 本文中所指的连接主义方法、深度神经网络和深度学习通常具有这些特征：采用非概念化元素组织成内部多层级网络；不对每一个元素进行精确的类型和语义定义，主要通过权重等参数来体现彼此的差异；面向固定输入和输出的端到端拟合和计算；采用单一的结构模型例如 M-P 模型来表达和计算；训练和学习的手段主要是调整权重参数。

### 2.1.3. 概率图

概率图模型将概率理论和图论结合起来，应用到特定系统中解决不确定性问题，并秉持用因果关系来取代相关性关系的理念。这些理论和和方法都非常正确且重要，已经在很多领域里产生了巨大的价值。但要实现更强的人工智能还很不够，最根本的问题是：现有概率图的各种理论主要针对封闭场景和系统，不能适用于开放领域里的包含多种关系类型和动态结构的复杂认知系统，在这样的领域和场景下需要对概率模型的基本假设和描述计算体系进行全新的设计。

## 2.2. 目标系统

对上述关键问题进行分析的基础上，我们提出一种实现更强大的人工智能的方案。

实现强人工智能系统需要涉及很多领域的大量的工作，但就基本的软件结构和算法而言，目标系统应该由"智能系统"+"外围系统"构成。其中智能系统位于核心并实现面向开放领域的通用智能处理；外围系统则作为辅助实现各种封闭领域的专业工具处理，被智能系统调用和指挥。

显然，智能系统的实现是难点，如前所述，它应该采用以场景拟合为中心的计算。

传统的系统多数都表现为单向函数式计算，这种计算的特点是：

- 对一端的输入根据一个函数映射计算出另一端的输出。例如：在单射函数 $f(x,y)=x+y=>z$ 中，x 和 y 是已知和输入，z 是未知和输出，计算过程是对 x,y 计算出 z，x,y 对 z 单向影响，z 不对 x,y 产生反向影响，计算的目标是输出端而不是整个系统，系统只是完成计算的固定工具。
- 输入和输出在系统设计时就固定下来，计算过程中不会改变。
- 中间的数据和网络结构针对该函数的单向映射计算而设计，不能支持反向计算和描述整个系统。

现有的神经网络系统多数都是单向函数式计算，内部网络自身是单向的静态转换结构，数据从网络中流转计算后输出，网络自身不会动态构建并对场景进行拟合。

单向函数式计算实现工具计算任务表现优异，支撑起数十年来信息技术的飞速发展。

从表象上来看似乎人脑也是单向函数式计算—每一个神经细胞的信息传递基本是单向的。M-P 模型就是以此为理论基础而设计，衍生出的多层神经网络系统等相关技术在一些领域取得了显著的效果。这更产生了一种假象：似乎数据的单向流动就能产生智能。

但事实并非如此，单向函数式计算缺乏全面性和灵活性，难以实现更强大的智能计算。

再来看方程式的计算，它比函数式计算更为灵活，如果将前述的函数转换为方程式 $x+y=z$，那么可以看到：

- 方程式本身并不区分自变量和因变量，所有变量都是平等的，因此不限定已知未知和计算的方向，具体计算时既可以根据 x,y 计算 z，也可以根据 x,z 计算 y，或者根据 y,z 计算出 x。
- 本质上，一个函数式就是一个方程式的多个变化形态中的一个。方程式是一种描述，函数式是一种计算。**"描述重于计算，计算为描述服务"**。

场景拟合为中心的计算是一种系统化的结构计算，它具有方程式计算的本质性和灵活性，并且拥有更强大的描述和计算能力：

- 将所有变量构成的结构视为一个完整系统，不仅描述变量间的关系也描述变量和系

统间的关系。可以理解为：方程式的结构本身只是描述变量之间关系的工具，而场景拟合计算的结构自身也是一个目标变量(根变量)。
- **计算的目标是整个系统！**计算不限于根据已知变量直接计算未知变量，更主要是根据已知变量计算出整个系统并根据整个系统计算未知变量。例如一个 4 元素的场景拟合结构可以定义为 S(x,y,z)，其中 x,y,z 分别都和 S 存在着直接关系。可以根据多个或者仅一个变量对 S 以及其它变量求解，所有计算的可行性和结果评估完全根据概率来确定。
- **计算的方法是根据需要选择不同结构进行动态组合来构建对场景进行拟合的系统。**

要实现这种场景拟合的能力，需要设计一种全新的模型，即动态认知网络模型。

我们认为，智能的本质就是场景拟合(场景就是世界的一个片段)，也就是认知。

**认知**：人脑的认知是用一组神经元的特定状态表达的逻辑结构来拟合目标场景；计算机的认知则是构建一种数据结构模拟这种人脑的逻辑结构来拟合目标场景。这种逻辑结构可以称为认知结构，具体体现为一种网络结构，因此又称为认知网络结构。这种结构是一种全面而本质的描述结构(类似方程式)而不是单向的转换计算结构(类似函数式)，它独立于计算且重于计算，各种计算和输入输出都是对同一个认知结构的不同运用。

以场景拟合为中心的计算以构建起对目标场景进行充分拟合的认知网络为目标,其中输入输出也是构成目标场景的认知网络部件。因此，从输入到输出的计算总是分为三个步骤：
- 获得输入信息—即目标认知网络的已知部件；
- 根据输入信息推算并动态构建完整的认知网络进行场景拟合；
- 从构建完成的认知网络中提取需要的信息部件输出；

先举一个简单的例子：对"猫"构建了一个认知网络[猫：猫拥有一个头、猫拥有 4 只猫爪子、猫拥有一条尾巴…]，这是一个本质的描述网络，它自身并没有固定的输入和输出，直接询问"猫的输入输出是什么？"没有意义。只有在具体计算时才设定一部分信息作为已知而另一部分信息作为未知进行计算求解：在一个时刻从正面看到猫的头，则让头作为输入来计算构建整个猫的本质认知结构，这个结构构建完成后，不但可以实现识别和分类任务，还可以实现推理出不可见的爪子、尾巴等任务；另一个时刻可能从背面看到猫的尾巴，则让尾巴作为输入来计算构建同一个结构，并同样地实现各种目标任务。

比起端到端的"直接"计算，这种总是经过场景构建过程的"间接"计算反而是从根本上解决问题的终极方案，可以实现更强大的智能，尤其体现在以下方面：

**1、全面性**：在完整场景拟合的基础上就可以统一地实现各种领域的认知表达和不同计算任务。例如对于图像识别来说,完成了空间场景的构建就同时实现了图像分类、语义分割、实体分割等不同任务，而不需要分解为不同的系统来分别实现。并且，全面的场景拟合不限于猫这样的事物和空间场景，还可以将空间、时间、因果、思维、交流等各个领域的认知信息都构建为相同结构的认知网络来进行全面的场景拟合，这种拟合既包括对场景中已发生的事件和状态进行还原，也包括对场景的过去进行推算和对未来进行预测甚至操纵。这样就不再限于只能执行单一预设任务，而具备了全面的智能计算的能力。

**2、灵活性**：复杂场景可以由不同领域的场景组合构成，具体实现是按需选择认知网络模块来动态组合为新的认知网络，这种组合在运行时进行，根据任务的需要和计算过程状态实时调整，体现出强大的融合能力。固定结构的模型和系统很难实现这种能力。

**3、连续性**：围绕一个独立于计算且持久存在的认知网络结构为中心，可以实现多种任务的连续进行。

**4、解释性**：形成的认知网络结构是完全可解释的，甚至对于中间计算推理的过程都可以表达和解释。

**5、本质性**：这种计算更符合人类智能的本质。我们认为：**人脑的主要工作模式就是场景拟合！**虽然在局部观察到的神经细胞之间的单向信息流动看起来是单向函数式计算，**但从宏观的视角来看众多单向的信息流动(并结合 xor 等机制)结合起来的整体效果就是构建正确的场景拟合结构！**相对而言，M-P 模型只是表达了整个过程的局部而不是全貌。

这里列举一个示例来说明以场景拟合为中心的计算具有跨领域信息和任务整合的能力，并且这些能力通过共享同一套认知知识和算法来统一实现。

假设汤姆和杰瑞相遇了，并发生了以下对话：

*汤姆说："你好。"*

*杰瑞问："你好，你是法国人吗？"*

*汤姆回答："不，我是美国人。"*

*杰瑞说："太好了，我也是。"*

这个例子的顶层是一个[对话]场景，里边嵌套包含了[因果]、[询问]、[回答]、[国家]、[人]等不同类型和大小的概念和场景，它们的外在表现(例如影像、声音、语言)同样也是场景的一部分，也是认知网络的构成部件。

但是每个任务执行之初能得到的已知信息只是完整场景的一部分，并且在每次任务中都是不同部分，任务的核心总是根据这些**不同**的已知信息去推算补全**同一个**完整场景。

首先来看一个理解任务，假设机器人作为在场的第三方看到了汤姆和杰瑞对话的整个过程并进行理解。它能直接获取到的是影像、声音等信息，任务就是根据这些已知的部分信息来补全整个场景，具体过程涉及：根据图像识别补全出汤姆和杰瑞等实体；根据视频识别补全出[汤姆对杰瑞说话]等动作；通过语音识别得到说话的文字并进行语义理解得到说话内容；以及构建整个过程的因果关系……。而一旦能正确补全整个场景，各种特定的输出就变得非常简单：例如对于询问"杰瑞在做什么？"、"汤姆是哪国人？"，就可以用问题在构建的场景中匹配信息并提取正确的回答；也可以进一步的推理，例如询问"杰瑞是哪国人？"，这时需要对场景进行扩展，因为推理也是场景的一种组成结构，可以增加推理部件来扩大场景直到包含"杰瑞是美国人"的信息并提取输出。

再看一个对话任务，让机器人扮演汤姆这个角色进行对话任务。这个不同的任务中，只是设定的已知和未知发生了变化，整个场景并没有变化，所依据的认知知识也没有变化，根据部分信息去补全整个完整的场景的核心原理更不会变化。

更具体地分析一条信息,假设认知知识库里存在[[A 询问 B 问题] 引起 [B 回答 A 答案]]这样一条推导知识,根据这条知识就可以将[杰瑞问："你好,你是法国人吗？"]和[汤姆回答："不，我是美国人。"]这两个动作场景通过中间的推导关系组合构成更大的场景[[杰瑞问："你好,你是法国人吗？"]] 引起 [杰瑞回答："不，我是美国人。"]]。机器人作为第三方理解对话时，两个动作都是已知的，只需要补出中间的未知部分[引起]；而机器人扮演汤姆进行对话任务时，根据[杰瑞问："你好,你是法国人吗？"]这个已经发生的动作计算汤姆的反应，需要做的仍然是依据已知补全场景，差别只是：只有[问]是已知，因此只能根据这一条已知来补全[引起]和[答]这两个未知部分。

这里的重点是：每次的任务和所处的观察角色不同，因此输入和输出不同，**但计算依据的知识相同，构建的场景相同**。对于上述理解任务，已知的有[问]和[答]两个部件，所以求解结果会比较单一且概率较高，类似于根据 x,y 求解 S 和 z；而对于后述对话任务，已知只有[问]一个部件，求解的结果较多且每一个结果的概率较低，类似于根据 x 求解 S 和 y,z；甚至也可以设定[汤姆回答："不，我是美国人。"]作为已知求解前边的未知动作和关系，这

时，根据同样的知识，推算出[杰瑞问汤姆："你是 x 国人吗？"]这一个结果具有相对较高的概率。

再来看一个翻译任务，假如汤姆和杰瑞说的语言不同而要让机器人介入为双方提供翻译，这个任务中机器人仍然会遵循相同的知识和算法来拟合相同的场景。

总之，因为场景里包含了全部信息，文章分类、情感分析、语义搜索、信息抽取、文章摘要和文章生成等各种任务都可以围绕同一个场景拟合来实现，不需要分解成不同的任务并训练。也无法分解！因为所有信息是相互纠缠的。

另外，从上述例子中还应注意以下要点：
- 动态结构和静态结构相同处理，推算还未发生的信息(例如上述杰瑞对汤姆的问题进行回答)和推算已经发生但不能直接获取的信息(例如看到猫的头然后推算看不到的猫的尾巴)并没有本质区别。
- 在因果结构上，顺着时间向后的推算和向前的推算也没有本质区别。
- 主观信息客观化，对人类思维和交流的处理和对客观信息的处理没有本质区别。
- 行动和理解没有本质区别，自己执行行动和理解别人的行动遵循同样的知识。
- **总之，在场景拟合为中心的系统里，各种不同的计算任务没有本质差异，差异主要体现在概率等基本方面！**
- 场景拟合面向综合领域而不局限于自然语言等特定领域。在真实场景中，对汤姆和杰瑞说出的话语用自然语言处理，对说话动作用影像处理；在本论文中，因为无法插入视频所以对说话动作也采用自然语言来表达和处理。无论哪种方式，构建的目标都是相同的场景和内部认知结构，这个结构可以同时包含图像、影像和自然语言等不同的外部形式部件。

以场景拟合为中心的智能系统的设计和实现原则如下：

### 2.2.1. 统一的结构定义

智能系统的基础是一套由结合内置概率模型的概念化元素构成的网络结构。

**概念化元素**：符号是用来对信息进行描述和计算的形式，神经网络的隐藏层的节点和连接本质上也是一种符号。符号主义和连接主义的本质区别体现在对符号的定义和实现有不同的原则，具体来说，分别采用了概念化和非概念化的符号。

概念化的符号经过了归纳、规整、正交化的优化处理，非概念化的符号是一种模糊、冗余、非正交化的表达。

举一个例子来比喻，如图 1 所示，要表达二维平面上的一个点的信息，用两个正交的维度 X 和 Y 来表达可以视为概念化的符号表达；而用未经优化和正交处理的多个维度 $V_1$~$V_n$ 来表达则是非概念化的符号表达。

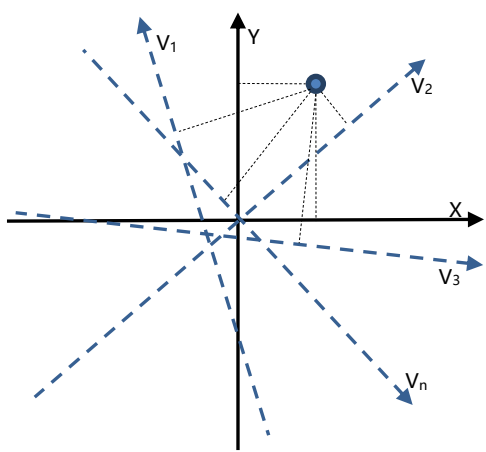

X、Y：概念化符号　　V₁~Vₙ：非概念化符号

图 1：概念化符号和非概念化符号对比示意图

显然，概念化体系对目标的描述更为清晰和高效，非概念化的体系即使能完成和前者等效的任务，但在灵活性和性能方面也要差很多。另外还有关键的可解释性问题：人类的认知思维和语言是高度概念化的，对含糊不清的信息难以做出人类能理解的命名和解释。假如 X 和 Y 对应于人的身高和体重，就可以给它们赋予"身高"和"体重"的语言命名来进行表达和交流，而 $V_1$~$V_n$ 都混杂着身高和体重的信息，无法进行有意义的命名，也就谈不上解释和交流。

因此，概念化的符号体系是系统追求的目标。即使可以将使用非概念化符号来初始化构建系统并自动学习训练作为一种方法手段，但最终的目标也应该向概念化系统进化来实现更好的优化（机器学习的一个重点就是用压缩、裁剪等网络简化处理来尝试达到类似目标）。如果一种机器学习方法足够优秀，自动学习并优化后形成的结果具有和概念化系统相同的能力并且具有相差无几的性能，有理由相信它的内部已经是概念化的系统[6]。

现在问题变成了：如何来构建概念化的结构体系？显然，如果自动机器学习能做到，那当然就是机器来做，如果还做不到，就只能先采取人工设计来做。我们认为现阶段可见的自动机器学习技术还不具备学习获得能全面地描述世界的概念化模型的能力，因此，这个工作还需要更多借助人工方式来实现。合适的技术框架可以将两者进行结合：人工构建为主机器学习为辅助，两者共享同一个知识库。机器学习做大量数据的相关性分析来构建待确定的雏形知识；人工构建为机器学习设定先验约束并对学习到的知识进行确认和修正。

因此，有别于主要采用非概念化符号构建并采用权重参数来调整的神经网络，新的智能系统应该是概念化的符号体系，每一个节点和关系都是对知识有着更精确、高效且可以解释的概念化符号。

*(对于人脑来说，也许在微观上每个神经元并不直接对应概念化的符号，但是有理由相信，这些微观结构形成的宏观效果是概念化的。计算机的基本原理和人脑有很大差异，所以没有必要完全模拟人脑的微观细节，直接模拟其概念化的宏观效果将更为高效。因此，本文提出的模型并不称为神经网络，而称为认知网络。)*

**网络结构**：网络结构具有灵活和强大的信息描述能力。秉持符号主义、连接主义、概率图、图神经网络等不同理念的模型和系统，都体现为某种网络结构。

新的智能系统的基本结构应该是由概念化的节点和关系构成的多层级的异构网络结构，具有足够的连接密度和结构复杂度(以这个标准来衡量，现有的知识谱图的结构过于简单)，以实现对丰富的知识进行深度、完整、灵活的表达。其中：异质结构要求网络的每一个元素都可以具有不同的类型和语义；这种多层级结构是按需精细组织的动态结构，和固定的多层级结构存在显著的不同。

**概率模型**：人类对世界的认知紧密结合着对不确定性的描述和计算，处理不确定性问题是 AI 系统的基本要求，可以作为区分 AI 系统和非 AI 系统的关键特征之一。智能系统应该

假设：不确定性概率(0~1)是绝大部分信息都具有的基本性质，确定性概率(0 or 1)只是不确定性概率的特定取值。

回顾已有系统，没有结合一个合理的概率模型是很多系统难以实现强大智能的重要原因之一。符号主义、连接主义和概率图等体系都证明了不确定性计算在 AI 领域的重要性并有不同的实现，只是这些实现方法仍然存在着问题需要改进。

一些系统(例如 prolog 语言、很多知识库系统)选择性地只对确定性的知识表达和处理，而对更普遍的不确定性的知识进行忽略或者强制去除不确定性，使得知识表达和计算不具有普遍实用性且具有很大的初始误差和累积误差，导致了无法解决的知识冲突难题[1]和缺乏可扩展性等问题。

另一些系统采用[all]、[exist]等量词来定性地表达知识的不确定性，应该统一为量化的概率来进行描述和计算。

如前分析，多数神经网络的内部结构是非概念化的，所以采用权重等机制而不是概率体系来解决不确定性问题，不具有概念化的概率体系对于认知表达和计算的完整描述能力，这也是系统难以解释、难以精确调试和调整等问题的重要原因。

概率图模型没有和富语义的异构认知结构结合，其概率模型的基本设计是面对封闭域、结构类型简单的静态网络系统，不适用于面对开放域、采用复杂类型结构的动态认知网络的智能系统。

新的智能系统必须解决上述问题，设计一个和认知网络相匹配且更适合于认知表达和计算的概率模型。

**统一结构**：这个基本结构具有高度的统一性。它既是描述结构，也是计算结构和记忆结构。

### 2.2.2. 构建认知知识库

在前述的统一结构模型基础上，针对不同领域构建认知知识库，应达到以下目标：

**全领域描述**：对于不同领域的认知知识(例如：事物、事件、动作、属性、关系、因果、思维、形式、图像、影像、自然语言、甚至计算过程)都用前述的统一结构来表达，可以将它们融合为一体来实现对复杂的世界和场景进行全面的拟合。

**本质描述**：认知知识应该描述目标场景的本质结构，减少为不同的任务分别构建和训练彼此孤立的知识。

**数据化表达**：各种知识尽量采用数据化表达，减少硬编码等固定形式的表达。

**知识和数据同构**：对于知识和数据采用相同的表达方式和计算处理，区别仅体现在它们所处的抽象层级。

**全面表达、选择使用**：尽量构建更全面的跨领域知识实现认知描述，并在具体场景下根据实际需要选择最有效的部分进行使用。这是人类智能的一个重要特点和优势，也是面向开放域的人工智能系统应该遵循的基础准则，可以从根本上解决独立同分布假设和固定参数结构的模型存在的问题。

### 2.2.3. 动态认知网络算法体系

在前述的模型结构的基础上实现动态认知网络算法体系，应该具有以下特点：

**动态网络生长**：总是围绕认知网络为中心组织各种算法处理，**计算的过程和结果都体现**

---

[1] 知识冲突本质上是把不确定性的知识强制变成确定性而引发，还原每个知识的不确定性是解决冲突问题的基础。

为同一个认知网络。这个认知网络随不同的算法处理进行创建、生长、变化，存在于整个任务的全过程甚至更久。

**算法融合**：将各种解析、分类、生成、查询、推理、甚至学习等算法融合为一体，它们都体现为同样的动态网络生长并共享同一套认知知识库，各种算法可以实时动态组合，并且可以实现多领域融合的增量式的连续计算。

**面向结构的机器学习**：机器学习获得的新知识必须和已有知识具有相同的结构，且遵循学习结构优于学习参数的原则，结构是定性而参数是定量，学习算法首先能学习到新的认知网络结构，在此基础上根据需要再针对结构的参数进行调整优化。

学习结构比学习参数更为困难。很多机器学习都是预置了某种固定结构后调整参数，这对于这种预设结构恰好能满足的单一任务会有效，而对于更复杂的智能计算任务无法通过调整参数来解决预设结构的描述能力不足的根本问题。

学习结构主要有两个困难：一是学习的目标结构定义问题，目标结构不明确自然无从学起。二是知识的可微问题，现在主流的机器学习方法通常要求知识是可微的。

新的智能系统要从根本上解决这两个问题：首先，智能系统的核心理念就是构建起统一的概念化认知结构，这就为机器学习限定了具体的目标结构；其次，结构学习的本质就是学习离散化的知识，离散值比起连续值更为基础且两者可以融合，智能系统将把离散化的结构学习和连续值的参数学习结合起来形成完整的认知学习。

## 3.动态认知网络

根据前边对目标智能系统的基本原则的分析，我们给出了具体的实现方案。

这个实现方案的基本模型称为"动态认知网络"或者"结构网络"，简称"DC Net"或者"S Net"。它面向开放领域，将各种认知知识概念化为认知网络元素，并融合以双向条件概率为基础的认知概率模型，组织为两维度多层级的认知网络结构来对场景进行描述和拟合，采用任务目标和概率驱动的全向网络匹配-生长算法体系来进行各种计算处理。

以下对DC Net模型从结构模型、概率模型和算法体系等几个方面进行具体说明。

### 3.1.结构模型

**基本元素**：作为一种网络结构，DC Net也由节点和边构成，一个DC Net可以视为一个三元组：

S=(C,R,P)

C：概念集合，每个概念体现为网络的一个节点。
R：关系集合，每条关系体现为网络的一条边。
P：参数集合，包括全局参数和概念与关系的私有参数。

概念和关系也统称为元素。

### 3.1.1. 概念

**概念**：DC Net是多层级的认知模型，通过对概念的逐级泛化和组合来表达对世界的认知，可以概括为"一切都是概念"。

概念的定义和描述具有全面性和基础性两方面的特点。

全面性：面向综合领域的认知进行概念化表达。对各种基础概念、参数概念、实体概念、关系概念、事件概念、因果概念、场景概念等都纳入到同一个概念体系里，并且对它们进行融合以实现全面的认知表达和计算。

基础性：注重优先构建最基础的概念(例如：数量、程度、集合、空间、时间、存在、比较、关系、推导等)，用它们构建起核心的语义和计算体系，以支撑综合领域的各种具体概

念的实现。

如何构建详细的概念和关系体系是一个专门的课题，不作为本文的重点。

### 3.1.2. 关系

**关系**：关系是特殊的概念。一条关系可以连接两个概念或者两条关系，两个元素分别称为关系的 A 端和 B 端。

在 DC Net 中，关系是双向的，一条关系可以看着两条相反方向的连接的结合：一条是从 A 到 B 的连接，称为 A=>B 连接；另一条是从 B 到 A 的连接，称为为 B=>A 连接。

关系比概念更为基础，概念和网络的语义及计算规则都主要由关系来承载。一个孤立的概念本身没有任何意义，每个概念都通过和其它概念的关系来实现对自身的解释，概念之间互为解释。

**关系的参数**：关系具有不同的参数。例如：表达空间结构的[拥有部件关系],[相邻关系]拥有[角度],[距离]等参数。

**关系派生**：DC Net 里一个重要设计是，关系和概念一样也是多层级派生的体系，关系和关系连接的概念一起同步派生并实现了整个网络的派生。

**基本关系**：处于派生体系最顶层的关系称为基本关系，不同的基本关系具有不同的语义、参数和计算规则。

下边列举最主要的一些基本关系进行说明，它们被归为集合维关系和领域维关系。

**1、集合维关系**

集合维关系表达了同类型概念的集合关系以及派生概念对基概念的语义继承关系。

**属于关系**：属于关系表示为[A 属于 B]或者 A⊆B。A 和 B 是关系的两端，可以是两个概念或者两个关系。A 称为派生概念(或派生关系)，B 称为基概念(或基关系)。

属于关系的条件概率满足 $P(B|A)=1, 0<P(A|B)<=1$，这用统一的条件概率量化地表达了"所有的派生概念都是基概念，部分的基概念是派生概念"的语义。

属于关系具有传递性，即：如果 A⊆B 且 B⊆C，那么 A⊆C。

一个概念可以具有多个基概念，并可以动态地为概念增加和删除基概念。

**等价关系**：等价关系表示为[A 等价 B]或者 A=B。

等价关系是属于关系的特例，其条件概率满足 $P(A|B)=1, P(B|A)=1$。本文中所说的属于关系都包含等价关系。

等价关系也具有传递性，并具有对称性。

**值概念的属于和等价**：对于连续值进行离散化为值概念，并用属于关系和等价关系表达值概念之间的集合关系。

例如：25 ⊆ (20,30) = (20,30) ⊆ (10,50) ⊆ (0,100) = (0,100)

值概念的属于关系和等价关系通过实时计算来体现，无需形式化表达。

集合维上的关系以及派生体系非常重要，是支撑整个认知描述和计算体系的基础。派生体系具有以下一些特点：

**派生重载**：派生概念(派生关系)继承基概念(基关联)的所有语义和参数，并可以对它们进行重载。重载的新值不能脱离基概念定义的范围而应该是更具体的取值。

**概念和实例归一化**：DC Net 很重要的理念是并不区分概念和实例的基本定义。本质上，对一切概念进行描述的目的都是为了映射特定的外延，实例也是概念—映射的外延数量等于 1 的概念。

为了强调这一点，分析这样一个派生链：太阳⊆恒星⊆天体⊆事物....。其中恒星、天体、事物显然是类，因为其映射的外延数大于 1。对于太阳应该定义为对象还是类则存在不同方案：传统的观点通常把太阳和另外三者区别对待，定义为对象实例而不是类，并对结构和算法采取不同的实现方法；但在 DC Net 里，认为这几者没有区别，它们都是概念，仅仅只是太阳映射的外延数量为 1 而已，而外延数量为 1 本身是难以保证永恒正确的信息，根据这样的信息对概念进行绝对划分会产生很多问题。

因此，DC Net 中一切都是概念，不对知识、数据、类、对象、实例等进行基本类型区分，并用同一种属于关系来实现概念之间的泛化关系—包括知识和数据、基类和派生类、类和对象实例之间的关系，不需要额外的实例化关系。用面向对象的术语可以称为将对象和类进行了统一——"实例即概念，对象即类"，这对于实现一个高度一致性的认知模型非常关键。

**变量绑定的形式化**：在需要时可以将变量绑定(变量赋值)也进行形式化处理，例如对"汤姆.年龄=15"的变量绑定操作可以表达为[汤姆.年龄 等价 15]关系。将单向的过程化操作转换为双向关系还可以对包括时间、版本等过程的因果场景进行描述和跟踪，例如用[[汤姆.年龄 等价 15] 拥有时间 2010 年]、 [[汤姆.年龄 等价 16] 拥有时间 2011 年]来描述两个关系。

**2、领域维关系**

领域维关系的语义比较丰富，并且可以根据需要进行扩展。以下列举部分典型的领域维基本关系进行简要表述，它们划分为纵向关系和横向关系两大类。

- **纵向关系**

纵向关系的两端概念具有不同的高低层级，典型的纵向关系举例如下：

**拥有部件关系**：表达为[A 拥有部件 B]，A 是高层概念，B 是低层概念，表达了 A 由 B 组合构成。

拥有部件关系的条件概率取值满足 0<P(A|B)<=1,0< P(B|A)<=1。

例如：[人 拥有 脸], [脸 拥有 眼睛], [脸 拥有 鼻子], [问 拥有 问方], [问 拥有 听方]。

**拥有部分关系**：表达为[A 拥有部分 B]，是拥有部件关系的特例，A 和 B 是同类型概念，A 的颗粒度比 B 大，体现了对概念进行不同颗粒度的分解和组合。

拥有部分关系实现对概念进行不同精度的表达，根据需要可以对大颗粒度的高层级概念进行粗略表达，也可以展开下级小颗粒度的低层概念进行更精细的表达。

例如：[曲线 拥有部分 曲线][面 拥有部分 面] [时间 拥有部分 时间]。

**拥有属性关系**：表达为[A 拥有属性 B]，A 是高层概念，B 是低层概念即属性。

拥有属性关系和拥有部件关系相似，只是属性概念比部件概念更为基本，体现为一种特征参数。

例如：[眼睛 拥有 颜色], [人 拥有 数量], [脸 拥有 角度], [问 拥有 时间] ，[[脸拥有眼睛] 拥有 角度], [相邻关系 拥有 距离]。

**拥有形式关系**：表达为[A 拥有形式 B]，A 是高层概念，B 是低层概念即形式概念。

在 DC Net 中，形式也是一种概念，和普通概念没有本质区别，只是在整个认知描述体系里处于相对外围和低层的位置，通常可以被感官或传感器直接感知和测量。形式概念的定义和具体领域相关，例如：自然语言处理里的语言形式(字符串、语言角色)、图像处理领域里的图像形式。

例如：[概念 拥有 语言形式]，[概念 拥有 图像形式]。
**拥有内容关系**：表达为[A 拥有内容 B]。A 是高层容器概念，B 是低层内容概念。
例如：[思维 拥有 内容]，[表达 拥有 内容]。
e.g., [think has content] , [represent has content].

- **横向关系**

横向关系的两端没有明显的高低层级，典型的横向关系举例如下：

**相邻关系**：表达为[A 相邻 B]，描述平级概念 A 和 B 之间的位置关系。
相邻关系拥有[角度][距离]等参数。
例如： [眼睛 相邻 鼻子]，[鼻子 相邻 嘴]。
**比较关系**：表达可量化的平级概念之间的比较关系，可以视为一种特殊的相邻关系。
比较关系拥有[比较差值]参数。
例如： [5>3]，[2001 年 早于 2002 年]。
**转换关系**：描述两个树形网的转换，是一种树形网关系。树形网关系的元素不仅包含关系自身和两端节点，还包含两端节点分别展开的树形网结构以及中间的附加关系。
例如：[加法 转换 减法]，[买 转换 卖]，[问题 转换 答案]。
**因果关系**：描述具有时间先后顺序的事件的转换关系。
因果关系的的条件概率满足 0<P(A|B)<=1,0< P(B|A)<=1，两个方向都可以进行直接推理计算。
例如： [起飞 引起 飞行]，[喜欢 引起 购买]，[问 引起 答]。
**变化关系**：表述同一个概念的状态随时间发生变化的因果关系，两端视为同一个概念的不同版本。
**移动关系**：同一个概念的位置随时间发生变化的因果关系，两端视为同一个概念的不同版本。
**等价关系**：等价关系既是集合维关系也是领域维关系,在领域维中体现为一种横向关系。两端可以是同一个概念的同一版本也可以是不同版本。
**互斥关系**：表示为[A xor B]，表达两个元素之间的互斥关系。
互斥关系的条件概率满足 P(A|B)=0, P(B|A)=0。

### 3.1.3. 两维度多层级结构(TDML)

DC Net 用概念和关系连接成认知网络，体现为集合维和领域维上的两维度多层级结构(TDML)。在集合维上以**派生网络结构**对认知模型的抽象性进行表达和计算，体现了**泛化**的能力；在领域维上以**树形网络结构**对各个领域的各种认知模型进行表达和计算，体现了**组合**的能力。两个多层级结构在语义内涵和计算规则上独立描述又紧密结合于一体。

**1、派生网络结构**
派生网络结构是集合维上的多层级结构。

定义：对于两个网络 $S_d$ 和 $S_b$，如果能在 $S_d$ 中找到一个子网络 $S_d'$，满足 $S_d'$ 和 $S_b$ 拓扑同构，且 $S_b$ 的每一个元素 $E_{bi}$ 和 $S_d'$ 中拓扑对应的元素 $E_{di}$ 满足 $E_{di} \subseteq E_{bi}$，这两个网络就形成派生网络结构，表达为 $S_d \subseteq S_b$，称 $S_d$ 是 $S_b$ 的派生网络，$S_b$ 是 $S_d$ 的基网络。

这个定义是单射而不是双射，要求基网络的所有元素在派生网络中都要有拓扑对应的元素，而反之不要求。这意味着派生网络在满足了基网络的基础上可以扩展或者派生重载自己

的结构和参数，从而实现灵活的网络组合扩展能力。反过来看，一个基网络代表着多个具有相同的基本结构而又扩展有不同细节的派生网络，体现了对认知网络实现泛化表达的能力。

**从基网络到派生网络的计算就是演绎和应用，从派生网络到基网络的计算就是归纳和学习。**

图 2-a 表达了一个二元关系的派生网络结构，即 $S_d(A,R,B) \subseteq S_b(A_b,R_b,B_b)$，图 2-b 表达了一个树形网的派生网络结构。

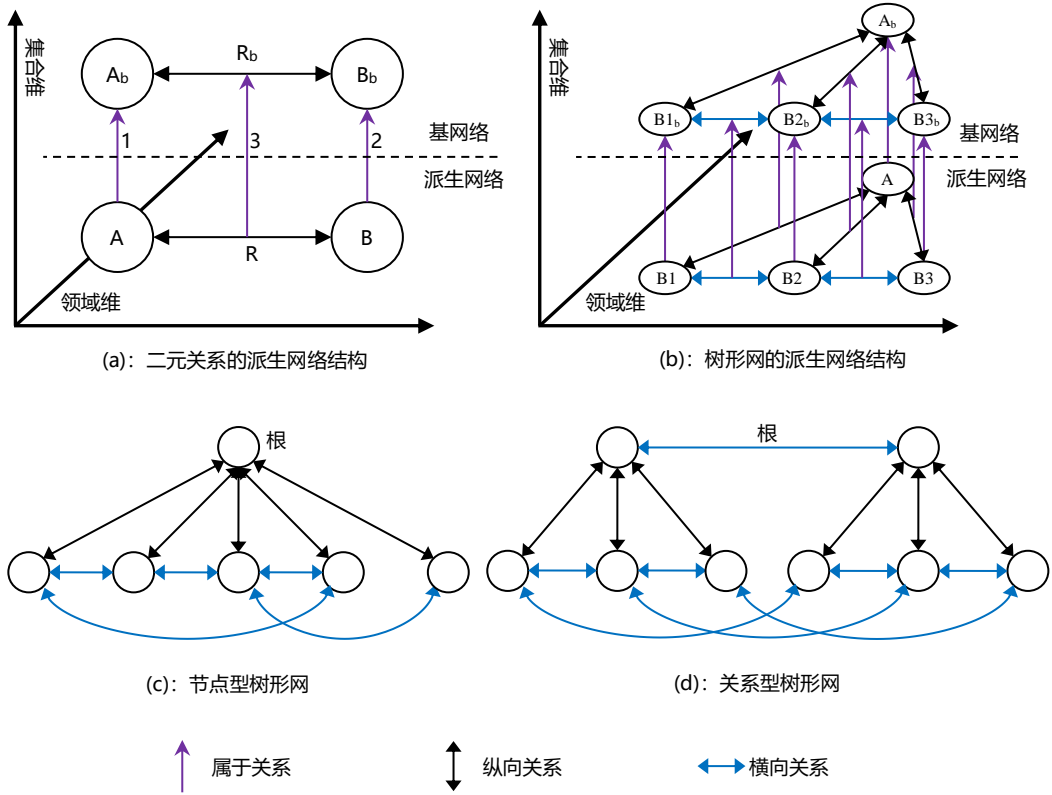

图 2：DC 网络基本结构示意图

**2、树形网结构**

树形网结构是领域维上的多层级结构。

定义：
1、一个树形网是一个连通网络。
2、有且只有一个根元素 Er，根元素可以是概念也可以是关系，在本树形网中是最高层的元素。
3、树形网里的关系根据其类型及其与根元素的连接关系可以划分为两类：
   a、树形网的纵向关系：多个纵向关系以从高往低的顺序连接且以根概念 Er 或者根关系的两端概念为最高层概念，这些纵向关系称为这个树形网的纵向关系。
   b、树形网的附加关系：其它的关系都称为树形网的附加关系，其中基本类型是横向关系的又称为树形网的横向关系。

图 2-c 表达了一个以概念为根元素的节点型树形网，而图 2-d 表达了一个以关系为根元素的关系型树形网。

对树形网的理解可以分主次两个方面来看待：

1、主体结构是以根元素为最高层元素并用多级纵向关系按从上到下展开的树形结构，体现了树的层级特征。

2、次要结构是在上述树结构的基础上增加了附加的关系，使整体结构从"纯"的树结构提升为网络结构。

树形网结构中的各个元素都是整个结构的组成部分，其中根元素可以作为句柄代表整个结构。

**树形网的嵌套**：树形网是认知网络的基本组织结构，任意一个 DC Net 都可以用预定义的各种树形网来组合构成。一个树形网可以收缩为根元素，根元素和外部元素的连接视为整个树形网和外部元素的连接。这样，树形网内部的关系结构对外部透明，并且可以以根元素为结合点对树形网进行组合嵌套。

基本树形网的定义应遵循最小化原则，每一个独立的树形网只定义自身不可缺少的组成部分。复杂的网络总是用基本的树形网组合嵌套来构成，这样可以把认知描述和计算的任务充分分解并独立解决。

对一个场景用多种树形网进行组合嵌套表达的示例如图 3。

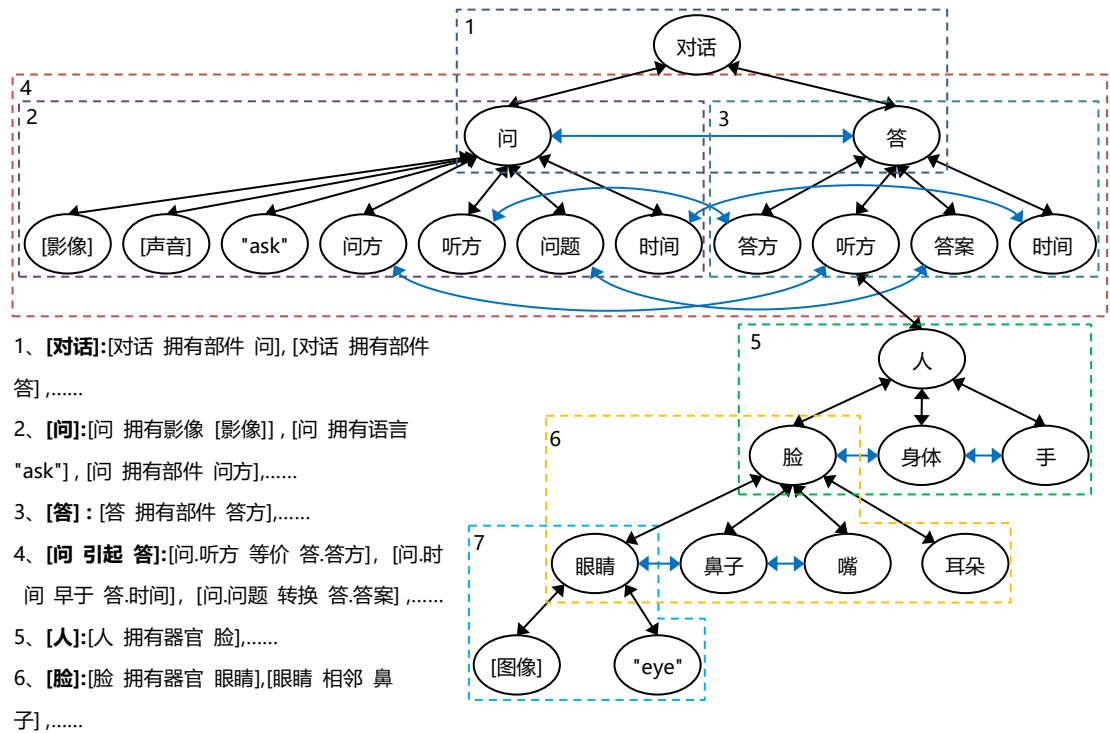

1、**[对话]**:[对话 拥有部件 问], [对话 拥有部件 答],......

2、**[问]**:[问 拥有影像 [影像]] , [问 拥有语言 "ask"] , [问 拥有部件 问方],......

3、**[答]** : [答 拥有部件 答方],......

4、**[问 引起 答]**:[问.听方 等价 答.答方], [问.时间 早于 答.时间], [问.问题 转换 答.答案],......

5、**[人]**:[人 拥有器官 脸],......

6、**[脸]**:[脸 拥有器官 眼睛],[眼睛 相邻 鼻子],......

图 3：树形网组合嵌套示意图

**树形网的意义分析**：单纯的树形结构和普通的网络结构都不能有效地表达认知信息。普通的网络结构缺少层级信息，并且难以进行问题分解；纯粹的树形结构则缺少完整地表达复杂模型的能力。树形网结构将树的层级结构和问题分解能力与网络的全面信息表达能力结合起来，将复杂的认知表达分解成相对简单的局部问题来独立解决，对 AI 的发展具有重要的意义。我们认为，人脑的运作也大量采用了类似于树形网的逻辑结构。

### 3、派生网络等价法则

派生网络结构和树形网络结构的结合产生了派生网络等价法则。描述如下：

假设：$S_d$ 和 $S_b$ 是两个树形网，其根元素分别是 $E_{dr}$ 和 $E_{br}$，$E_{ds}$ 和 $E_{bs}$ 是除了根元素以外

的其它元素。

那么：$E_{dr} \subseteq E_{br} \Leftrightarrow S_d \subseteq S_b \Leftrightarrow E_{ds} \subseteq E_{bs}$。

上述公式是双向推导的，其意义是：如果一个派生网络的根元素属于一个基网络的根元素，那么它的除根元素以外的其它元素就满足这个基网络的元素要求；反之亦然。

**部分满足推断**：上述公式表现了基网络的所有的元素都被派生网络满足就能充分地推算出派生网络对基网络的满足。但实际情况下并不是派生网络的所有元素都能都获知，经常需要只根据已知的部分元素来推算，这种推算的结果体现为一种隶属度，只要该隶属度达到一定的阈值就可以判断整个树形网也大概率会满足，整个树形网和未知的元素就可以全部推算和补充出来。这种量化计算在后边概率模型里具体叙述。

派生网络等价法则是人类对世界进行认知所遵循的最基本原理之一，面向对象、演绎归纳等思维方法都是这种原理的体现。DC Net 理论对此进行了网络形式的定义并结合概率计算形成数学量化的体系，作为认知描述和计算的基础。

### 3.2.概率模型

在介绍 DC Net 的概率模型前，先根据可用知识的范围和计算结果的概率两个方面对不同的计算场景进行对比分析。

**封闭域场景**：设定较封闭的边界条件来将可用的知识和信息限定到较小的范围，采用固化结构，从而简化计算。

**开放域场景**：减少边界条件限定来扩展可用的知识和信息的范围，采用动态结构，追求更全面地运用信息的计算。

**非显著性概率场景**：根据已知的模型、知识和信息，对未知目标计算的结果的概率较低而不具有显著性取值(例如:$P<0.9$)，计算的目标是选取相对最优结果来作为最终结果。

**显著性概率场景**：根据已知的模型、知识和信息，对未知目标计算的结果的概率较高而具有显著性取值(例如:$P>0.9$)，计算的目标可以追求达到显著性阈值的绝对最优结果作为最终结果。

根据开放域和封闭域对计算场景进行划分非常重要，从传统系统和弱人工智能向强人工智能进化的关键点是要将假设条件从封闭域向开放域转变。这个过程中需要解决由此产生的困难，但也会获得额外的优势。

对于封闭域系统来说，通过客观或者主观地限定已知条件和知识到封闭范围可以使计算变得简单，这将产生两种系统：

**系统 A**：在封闭域和显著性概率的场景里，任务目标本身比较简单清晰，根据封闭域限定的信息已经足够进行准确的计算。这就是众多传统计算机系统所面对的工作，这些系统已经很成熟且很好地运转，效率远远超越了人类。

**系统 B**：在封闭域但非显著性概率的场景里，限定的信息不足以推算出绝对正确的结果，只能选择出"相对较优"的结果，缺乏可靠性和稳定性。这是很多弱人工智能系统所处的现状。

对于开放域的系统来说，增加了更多可用的信息使得系统复杂度增加，但另一方面，如果能充分运用增加的这些信息就能获得更优甚至绝对最优的结果。具体地也可以划分为两种系统：

**系统 C**：在开放域场景下，获取了更多信息后可以实现显著性概率计算。

举一个例子来说明：人类从图像里识别出一个人通常会具有极大的置信度和稳定性，即得到了显著性的概率($P \approx 1$)。本质上是因为充分运用了图像中的大量信息—鼻子、眼睛、皮

肤、衣服、环境…，这些都是识别人的有效的信息，这些信息是过量的—超过了计算获得显著性概率所需要的信息量，实际上，即使遮挡住人的一半减去一半信息，仍然能得到同样的显著性概率结果。

如果能获得显著性的概率结果，那么可靠性、稳定性以及可解释性将不是问题。计算获得一个概率>99%的显著性结果，比起在75%和25%这两个结果里选取最优的75%，差异不是用99%和75%相差的22%来衡量，而是1%和25%之间25倍的差异。计算上也会更简单，如果得到显著性的结果就不需要对不同结果进行穷举和比较处理。

目前在这种场景下的系统的可靠性和稳定性不够的根本问题就是不能充分利用开放域下丰富的信息的优势。当然，这在理论上是能够解决的，也是必须要突破的工作重点。并且，一旦能做到充分利用丰富的信息，就可以更主动地扩展获取信息的方式和范围来得到更丰富的信息进一步提升系统的可靠性，这在当今的技术条件下并不是问题。

**系统 D**：在开放域场景下，充分运用了所有能获取的信息后对于任务目标仍然不能计算得到显著性概率的结果。这种情况不仅是人工智能系统需要关注的问题，同时也是人类自己也在不断遇到并尝试解决的永恒问题。无论是人工智能还是人类智能自身，都并不能用一个简单方法一劳永逸地解决，而是根据情况采取多方法、多步骤逐渐解决。大体上遵循的原则是：先把能够归纳为系统 C 的问题分解出来解决，尽可能确定能确定的信息，然后用已确定的信息来包围和限定不确定的系统 D，使后者的范围变小和难度降低，再逐步解决。

再重新审视下系统 B，可以分为两种情况：一种情况是对信息的封闭性限制是受客观条件的限制而难以改变，那么认为属于开放领域但确实无法获取足够信息，可以归于系统 D 的范畴。另一种情况是因为目前技术能力不够而造成的伪限定甚至是主动自我设限，那么就应该提升技术来解除这种限定转换成为系统 C 来有效解决。

根据上述分析，现在的关键是要对于系统 C 进行很好的解决实现，这将是迈向强人工智能方向上的重要突破。

DC Net 的概率模型是紧密围绕 DC 认知网络结构设计的概率表达和计算模型，面向开放域、显著性概率、异构和动态网络结构的场景，称为认知概率模型。实现目标是充分地运用在开放域中可以获取的大量知识和信息，对任务目标计算出接近显著性概率的结果。

根据上述假设条件和设计目标，认知概率模型具有以下特点：

**统一概率评估体系**：DC Net 将跨领域的多种结构融合为一体，对于所有认知概念和结构都用同一套概率体系来描述和计算。

**概率计算动态变化**：因为面向开放域可以获取大量知识和信息，可以根据需要进行选择使用，概率计算针对的不是固定结构、参数和权重的静态模型，而是实时变化的动态模型。计算过程和结果不遵循静态一致性而遵循动态一致性。这点和现有的多数概率理论和方法有显著的不同，实际上更接近人脑的计算方式。

**以概率加法为主**：根据面对开放场景和按需选用的假设，认知概率模型的概率计算以概率加法为主。

以下详细描述认知概率模型的实现。

### 3.2.1. 概率表达

**双向条件概率**：DC Net 采用双向条件概率模型，每一条关系都具有两个条件概率，具体定义和取值由关系的语义决定：

P(A|B)：在关系的 B 端概念存在的条件下，A 端概念存在的概率。

P(B|A)：在关系的 A 端概念存在的条件下，B 端概念存在的概率。

离散值的条件概率表达为一个概率值(0~1)。连续值的条件概率则用概率分布函数表达并用对应的隶属函数来计算隶属度，例如，[成人拥有身高]的条件概率可以定义为高斯分布函数 N(μ=1.7m, σ=0.1m)，根据一个身高值 x 计算 P(A|B)采用高斯型隶属函数 $f(x,\sigma,u) = e^{-\frac{(x-\mu)^2}{2\sigma^2}}$。

部分条件概率示例如表 1：

表 1：条件概率示例表

| 基本关系 | 基关系 | 关系 | P(B|A) | P(A|B) | 说明 |
|---|---|---|---|---|---|
| 属于 | 属于 | 苹果属于水果 | 1 | 0.00001 | |
| 等价 | 等价 | 苹果等价苹果 | 1 | 1 | |
| 拥有部件 | 脸拥有器官 | 脸拥有鼻子 | 1 | 1 | |
| 拥有部件 | 脸拥有器官 | 脸拥有眼睛 | 1 | 1 | |
| 拥有部件 | 脸拥有眼睛 | 欧洲人的脸拥有黑眼睛 | 0.3 | 0.05 | |
| 拥有形式 | 拥有形状 | 脸拥有椭圆形 | 1 | 0.00001 | |
| 拥有形式 | 拥有语言形式 | 脸拥有英文形式"face" | 0.9 | 0.6 | |
| 拥有形式 | 拥有语言形式 | 苹果拥有英文形式"apple" | 0.9 | 0.3 | "apple"可以解析为[水果],[公司]... |
| 拥有属性 | 拥有属性 | 人拥有身高 | N(1.0,0.5) | 0.00001 | |
| 拥有属性 | 人拥有身高 | 成年人拥有身高 | N(1.7,0.1) | 0.00001 | |
| 横向关系 | 转换关系 | 加法转换为减法 | 1 | 1 | |
| 横向关系 | 因果关系 | 问引起答 | 0.95 | 1 | |

**输入概率和结果概率**：DC Net 中的每一个概念都拥有概率，对一个网络进行计算时需要对每个概念记录两个概率。根据初始条件或者外部计算直接设定的概率称为输入概率；根据输入概率结合条件概率可以计算出各个概念的结果概率，简称为概率。

**概率重载**：在对关系重载时，可以为派生关系设定不同于基关系的条件概率值。

例如：基关系[脸拥有眼睛]的概率是 P(B|A)=1,P(A|B)=1，派生关系[欧洲人的脸拥有黑眼睛]的概率是 P(B|A)=0.3,P(A|B)= 0.05。根据这两条关系知识，用 B 端是黑眼睛的已知信息可以推断出 A 端是脸的概率为 1，是欧洲人的脸的概率为 0.05。

### 3.2.2. 概率状态

和传统的概率图模型不同，认知概率模型并不对一个认知网络计算一个概率值—例如联合概率或者势函数值，而是用每个元素的概率来描述认知网络的概率状态。对 DC Net 进行概率计算就是对每个元素的概率进行持续计算，这种计算的过程和结束没有明显的界限，可以随时结束也可以随时重启。理想的计算结果是每一个元素都达到概率坍缩，这等价于整个认知网络得到了完全确定的结果。

显然，认知网络概率状态更符合人类的认知模式。以图像识别为例，人类识别一幅图像不是对整张图像给出一个概率值，而是针对其中不同部分分别给出不同的识别概率。

某些计算任务确实需要对整个网络给出一个评估值，这个值也不应该是联合概率，而可以采用根据任务目标设定的某种评价指标结合概率计算的加权平均值。

### 3.2.3. 概率计算

**条件概率计算**：

假设关系 $R_{AB}$ 在两个方向上的条件概率是 P(A|B) 和 P(B|A)，那么：

已知 B 的概率 P(B)计算 A 的概率：

*公式 1：P(A)=P(B)\*P(A/B);*

已知 A 的概率 P(A)计算 B 的概率：

*公式 2：P(B)=P(A)\*P(B/A);*

**概率叠加公式：**

条件概率的定义遵循独立采样假设，计算一个概念 X 的概率可以根据 n 个存在直接关系的概念($Y_1, Y_2, ..., Y_n$)结合条件概率计算出 n 个概率并用概率加法进行叠加。

二元概率叠加按如下公式计算：

*公式 3：$P(X)=P(X_1 \cup X_2)=P(X_1)+ P(X_2)-P(X_1)P(X_2)$*

撤消一个二元概率叠加按如下公式计算：

*公式 4：$P(X_1)=(P(X)-P(X_2))/(1-P(X_2))$*

n 元概率叠加按如下公式计算：

*公式 5：$P(X)=P(X_1 \cup X_2 \cup ... X_n)$*

$$= \sum_{i=1}^{n} P(X_i) - \sum_{1 \leq i \leq j \leq n} P(X_i X_j) + \sum_{1 \leq i \leq j \leq k \leq n} P(X_i X_j X_k) - \cdots + (-1)^{n-1} P(X_1 X_2 ... X_n)$$

上述公式中：

$P(X_i)= P(Y_i)R(XY_i)P(X/Y_i)$

$R(XY_i)=0~1$：关系隶属度。例如：[脸拥有眼睛]的关系包含[角度][距离]等参数，根据这些参数在派生关系上的实际值和在基关系上的设定值计算关系隶属度。

因为 $P(X_1 \cup X_2 \cup X_3)= P(X_1 \cup X_2) \cup P(X_3) = P(X_1 \cup X_2) \cup P(X_3)= P(X_1) \cup P(X_2 \cup X_3)$，n 元概率叠加可以分解为任意顺序的二元概率叠加，便于实现多向计算和并行处理。

**概率传递和叠加算法(PPS)：** 在一个概念 Y 上增加了输入概率 P(Y)后，P(Y)以 Y 作为起点通过相关关系的条件概率向相邻的概念 $X_i$ 进行概率传递和叠加，使各个 $X_i$ 的概率得到提升。因为条件概率<=1，所以这种传递原则上是逐渐衰减的，传递到一定范围后会衰减到可以忽略而终止。

具体算法采用广度优先的网络遍历算法，执行一次 PPS 算法就完成 P(Y)对各个 $X_i$ 的概率传递和叠加，算法流程如下：

1、初始条件：概念 Y 增加了输入概率增量 P(Y)，标记 Y 为[已经遍历]，以 Y 为起点发起计算。
2、查询出一端为 Y 的所有关系 $R_{Xi-Y}$，这些关系的一端为 Y，另一端为 $X_i$。
3、对每一条关系 $R_{Xi-Y}$ 和对端概念 $X_i$ 进行如下处理：
　a、如果满足任意一条递归终止条件则返回；
　b、否则执行：计算 P(Y)对 $X_i$ 产生的概率 $P(X_i)= P(Y)R(X_iY)P(X_i|Y)$并按照公式 3 对 $X_i$ 的概率进行叠加；标记该概念 $X_i$ 为[已经遍历]，并设置 $Y=X_i$, $P(Y)=P(X_i)$后跳转到 2 进行递归处理。

递归终止条件：
1、如果对端概念 $X_i$ 标记为[已经遍历]。
2、对端概念 $X_i$ 的概率已经坍缩。
3、本次计算出的概率 $P(X_i)$的值很小，表示传递的概率影响已衰减到可以忽略不计。
4、达到了计算任务设定的截止边界。
5、根据任务要求和关系语义规定的其它递归终止条件。

**多路径和环**：DC Net 是全向描述的网络，并不排除环的存在。但在每一次 PPS 计算中，起点概念 Y 对一个目标概念 $X_i$ 的概率影响只允许通过一条路径进行计算，不会出现环。在存在多条路径时，不保证对同一个目标概念选择不同的路径计算的概率结果完全一致，但如果网络设计和概率取值合理，最终结果应该趋于一致。路径的选择规则可以设定为：选择对目标概念计算出的概率值为最高的路径进行计算。

对多个输入概率 $P(Y_i)$ 逐个调用 PPS 算法来对目标概念 X 进行概率叠加。因此，概率传递和叠加分为两个方面：

1、一个概念 $Y_i$ 的输入概率对一个目标概念 X 的概率影响通过多级条件概率乘法进行传递。

2、多个概念 $Y_i$ 的输入概率对一个目标概念 X 的概率影响体现为对上述概率乘法结果进行叠加。

**扩展讨论：**

按照公式 5 进行概率叠加计算具有清晰的数学解释，计算结果在理论上较为精确，但计算处理较为复杂。实际计算可以采用简化模型，一种可行的方案如下：

将公式 5 进行简化，只保留前边的概率分项部分而去除后边的联合概率部分，并为每个分项设置一个补偿系数 $k_i$ 进行补偿处理，就成为简化的概率叠加公式：

*公式 6*：$P(X)= \sum_{i=1}^{n} P(X_i) = \sum_{i=1}^{n} k_i P(Y_i) R(XY_i) P(X|Y_i)$。$0<k_i<=1$。

简化处理后，概率的取值区间从(0~1)变成了(0~>1)，概率坍缩阈值(参见后边)可以设定为 P>=1。

补偿系数 $K_i$ 的取值根据实际的模型和关系定义确定(通常取 0.5~1)，这个系数起到补偿的作用，使计算出的 P(X) 在(0~1)区间时和正常的概率叠加公式计算的理论值尽量接近并避免 P(X) 过快达到坍缩。而一旦达到了概率坍缩后，根据后述的坍缩规则，概率将不会继续增加，后续处理就和正常的概率叠加处理完全一致了。

举一个例子：假设鼻子、眼睛、嘴的已知概率都是 1 且它们对于脸的条件概率都接近 1，那么只需要其中任意一个参数就足够使脸的概率=1。在加入补偿系数=0.5 以后，每个参数计算出的概率都减小到 0.5，就需要两个参数的叠加才能使脸的概率=1，这在参数过量和显著性概率的假设条件下不存在根本问题。

公式 6 的简化处理比起公式 5 降低了理论上的精确性，但表达和计算处理简单，在很多场景下被广泛运用。

可以看到，公式 6 和 M-P 模型中的基本公式 $\sum \omega_i \cdot x_i + b$ 中的主体部份 $\sum \omega_i \cdot x_i$ 相似。我们认为：这表明类似的多项式是能有效地进行不确定性计算的基础公式，比概率图模型中通常采用的复杂联合概率计算更适用于开放域场景。

另一方面，和 M-P 模型中的权重和阈值相比，认知概率模型中的参数和计算结果都是取值为(0~1)的概率，具有清晰的可解释性和独立调整能力，计算处理只依据概率理论来进行，不需要对不同的激活函数和偏置项进行选择和训练。此外，将根据关系的参数计算的关系隶属度 $R(XY_i)$ 纳入概率传递和叠加使整个体系更完整。

神经科学研究揭示了人脑神经元的基本工作模式之一接近于多项式计算和激活模型，这也是 M-P 模型的主要理论依据。我们认为：这种观点总体上是合理的，人脑偏重于寻找更有效的信息来计算出一个显著性概率的结果，而尽量避免仅依赖不充分的信息进行"精确"的概率计算并在多个非显著性概率的结果中做出选择，采用类似上述简化的概率叠加模型的工作模式就能有效满足需要。

### 3.2.4. 概率坍缩

在开放域和显著性概率场景假设下，结合概率叠加和全向计算的特点，设计了概率坍缩的原理和方法。

按照场景拟合的理念，一切计算都是对未知计算使之逐步成为已知的过程。正确的目标网络中的每个概念的真实概率都是 1，但是在计算开始时只有部分信息是已知的，体现为一些概念的输入概率(P<1)，按照上述概率算法用已知的输入概率对各个概念的概率进行计算就是把未知逐步变为已知。

我们规定，计算过程中如果一个概念的概率达到预设的显著性阈值(例如：P>0.9)，就认为得到了这个概念的真实概率 1，从而触发对这个概念的坍缩处理。称坍缩后的概念处于"坍缩态"，尚未坍缩的概念处于"叠加态"。

对一个概念 X 的坍缩处理按如下顺序执行：
1、撤销在坍缩之前对 X 产生了影响的 PPS 计算以避免重复叠加，具体算法参照公式 4；
2、设置 X 的输入概率=1，并重新发起 PPS 计算；

一个概念的概率达到坍缩将对后续计算产生影响：这个概率不需要也不可能继续增加，后续的概率叠加进行到该概念将被截止而不需要继续传递；更重要的是这个概念成为绝对的已知，改变了整个网络的已知和未知状态分布，引起对后续计算流程的重新规整和简化，甚至转换不同的计算目标和方法。

例如：根据嘴和鼻子计算使脸的概率达到了坍缩，后续就不需要再使用眼睛、耳朵等对脸继续进行无意义的计算！不仅如此，脸成为绝对的已知后反过来对眼睛、耳朵等进行推算，即使它们的图像并不清楚甚至完全不可见—例如被墨镜遮住。

结合了概率坍缩后，整个认知网络的概率计算表现为如下过程：

- 一个初始的 DC Net 中的各个概念处于概率=0 的状态。
- 随着已知信息和输入概率的加入，通过 PPS 计算使各个概念的概率逐步提升，这时各个概念处于叠加态。
- 概率继续提升，某个概念率先达到坍缩态，一个概念的坍缩将把整个叠加态网络分割成多个子网络来分别求解，复杂度就下降了。
- 同时，一个概念的坍缩会增加较大的输入概率，使其它概念更快达到坍缩并推动整个网络全面坍缩—**"一点坍缩，连带坍缩"**。

对比传统一些系统的处理上的区别：

一些系统完全忽略概率，相当于初始就把所有的概念都强行坍缩为了 0 或者 1 两种状态。对于本来就面向确定性领域的系统来说没有问题，而面向具有大量不确定性的系统，相当于一开始就引入了概率误差，后边的每一步计算都同样纳入更多的误差，最终结果自然不会理想。

一些系统引入了对不确定性(0~1)的处理，但其假设条件和设计原理没有纳入概率坍缩的理论和方法，或者算法体系上不能支持按需选用和多向计算，使计算过程中整个系统一直处于概率叠加态不能局部坍缩简化，只能到最后环节才选择相对最优结果进行坍缩，过程中不确定性网络的规模和计算量变得很庞大，性能很低而结果却未必更为精确(不确定性的传递和计算本身有累积误差，**正确的局部坍缩其实会在中间过程消除这种误差！**)，而且信息越多网络越庞大这个问题就越突出。

概率坍缩也是人脑的基本思维方式，人类观察和解释世界随时会遇到未知和不确定性的

信息，需要尽快去确定甚至操纵能优先明确的信息，一旦将部分信息明确下来成为已知，就可以转变关注点和计算推理的流向，以已知再计算其它未知，这种条件和计算的转换持续不断地进行，就可以对复杂的世界进行有效的处理。如果对于该明确的信息不能尽快明确，面对着不确定信息越来越多的"混沌"体系，就做不了任何事情。

### 3.2.5. 概率扩展实现

前述的基本的概率处理模式可以实现大部分的认知概率描述和计算。如果需要，在局部可以用更精确的扩展的概率计算(可以结合 and、or 等逻辑计算)来替代基本的概率计算，这种局部的计算精度优化并不影响整体概率体系。

## 3.3.算法体系

以下对 DC Net 的核心算法体系进行论述。

### 3.3.1. 算法综述

DC Net 的核心算法体系围绕认知网络结构为中心，以任务目标和概率状态共同驱动认知网络的匹配和生长来实现各种任务目标，可以称为**全向网络匹配-生长算法体系(OMG)**，其基本原理和主要特点如下：

**网络生长**：按照场景拟合理论，每一个计算任务的核心都是构建认知网络来拟合特定的目标场景，构建过程体现为认知网络的持续生长过程。

**动态组合**：网络生长的具体实现方法是用认知网络组件进行动态组合成为更大的网络，这种组合可以无限嵌套并跨领域融合，具有很强的灵活性。

**按需选用**：知识库里拥有大量认知网络组件，实际计算根据需要选择出合适的组件来使用。

**算法融合**：包括解析、分类、生成、查询、推理、预测、行动、学习等各种算法都融合为统一的网络生长算法。

**全向生长**：在认知网络的定义和描述里并不设定输入输出和计算方向，只有在具体计算时才设定网络的一些部分作为已知另一些部分作为未知进行计算，**已知和未知的划分也是相对的，完全通过概率来区分！**因此整个计算过程和网络生长可以向任意方向进行并且实时调整，双向条件概率等模型设计对此提供了基础支持。

整体算法可以用以下形式描述：

$S_k' + S_t' = P_s(T, S_k, S_t, S_I)$

$S_I = P_I(I)$，$O = P_O(S_k' + S_t')$

$P_s$：DC Net 内部计算。

$T$：任务信息，包括任务目标和相关参数。

$S_k$：已有的知识网络。

$S_k'$：新的知识网络。

$S_t$：已有的实例网络。

$S_t'$：新的实例网络。

$S_I$：概念化的输入信息。

$I$：输入信息。

$P_I$：输入适配处理。

$O$：输出信息。

$P_O$：输出适配处理。

整体算法流程如下：
1、输入处理：对外围输入信息 I 通过输入适配算法 $P_I$ 概念化为 DC Net 结构的概念 $S_I$；
2、$P_s$ 计算：将 $S_I$ 和已有的 $S_t$ 和 $S_k$ 结合，按照任务 T 的要求进行计算，产生新的实例网络 $S_t'$ 和新的知识网络 $S_k'$。
3、输出处理：从生成的 $S_k'$ 和 $S_t'$ 中提取所需部分通过输出适配算法 $P_O$ 转换为任务要求的输出信息 O；

$P_s$ 计算的基本流程如下：
1. 初始条件：输入 $S_I$ 是 DC Net 片段的集合，每一个元素都具有输入概率，元素之间允许存在冲突和歧义。$S_t$ 是已有的实例网络，可以存在多个 $S_t$，每个 $S_t$ 分别进行生长处理。作为优化，概率更高的 $S_t$ 将优先生长，并可以抑制概率低的 $S_t$ 的生长。
2. 匹配：从 $S_I$ 中选取一个网络片段 $S_{Ii}$ 并结合 $S_t$ 的部分元素组成网络片段 $S_d$ 在知识库 $S_k$ 中进行匹配，如果成功将返回一个或多个知识网络 $S_b$，$S_d$ 对 $S_b$ 匹配计算的隶属度应达到激活阈值(激活阈值<坍缩阈值，根据不同任务而设置)。
3. 生长：以基网络 $S_b$ 为模板创建派生网络 $S_D$ 并加入到 $S_t$ 中进行生长，$S_d$ 将被吸收到 $S_D$ 中。
4. 处理：过程中进行概率叠加、概率坍缩、网络缩减等处理，如果需要也会对知识网络 $S_k$ 进行扩展或修改，即进行学习。
5. 迭代：重复以上步骤，直到 $S_I$ 的元素 $S_{Ii}$ 全部计算完成。
6. 结束：如果整体任务结束，则从吸纳了全部 $S_I$ 的 $S_t$ 中选择最优结果进行输出处理；否则等待新的 $S_I$ 输入进行下一轮计算。按照认知概率理论，优先选择所有元素都达到坍缩状态的绝对最优结果，如果不存在绝对最优结果，也可以对各个 $S_t$ 进行比较并选择出相对最优结果。

### 3.3.2. 匹配

网络匹配是很基础的算法，基本原理是依据派生网络等价法则和概率叠加公式，计算一个或者一组网络片段 $S_d$ 对于一个基网络 $S_b$ 的隶属度。

**完整匹配**：参见派生网络等价法则，如果 $S_d$ 是一个 $S_b$ 的完整的派生网络即 $S_d \subseteq S_b$，那么 $S_d$ 对 $S_b$ 完全匹配即隶属度 P=1。一些任务要求必须采用完整匹配，例如问题查询。

**部分匹配**：根据已知信息不完全原理，通常已知的 $S_d$ 并不是 $S_b$ 的完整的派生网络而只是派生网络的子网络，那么就执行 $S_d$ 对 $S_b$ 的部分匹配来计算隶属度 P，具体算法如下：
- **单概念的匹配**：概念可视为一个最小的树形网，计算一个概念 $S_d$ 对另一个概念 $S_b$ 的隶属度 P，分为两种情况：
  1、对于离散概念，如果 $S_d \subseteq S_b$，则 P=1，否则 P=0。
  2、对于连续值概念，根据基概念 $S_b$ 中记录的概率分布函数用对应的隶属函数计算 $P=N(S_d, S_b)$。
- **树形网的匹配**：一个树形网的匹配计算分为两个步骤：
  1、结构匹配：以基网络 $S_b$ 作为基础，假设 $S_d$ 是 $S_b$ 的一个派生网络 $S_D$ 的子网络，依据派生网络结构的定义进行结构匹配。如果派生结构匹配成功则执行下一步，否则表示不匹配而返回 0。
  2、隶属度计算：依据上述的结构匹配结果，将 $S_d$ 的每个元素 $S_{di}$ 作为已知信息分别代入基网络 $S_b$ 的对应元素 $S_{bi}$ 并分别计算各个元素的隶属度 $P_i$，然后在 $S_b$ 上用 $P_i$ 作为输入概率调用 PPS 算法进行概率叠加，最终得到 $S_b$ 的根元素的概率就是 $S_d$

对 $S_b$ 的隶属度 P。
- **嵌套树形网的匹配**：采用递归算法对每一级树形网逐级匹配，最终得到最顶层树形网的根元素的概率就是整个树形网的隶属度 P。

### 3.3.3. 生长

生长算法是以知识库里的基网络作为模板，通过创建、组合、增加元素等方式对实例网络进行生长并进行概率叠加。

首先定义三种基础生长算法：

**1、单概念生长**：以基网络里的基概念 $S_b$ 为模板复制一个派生概念 $S_d$，并设置属于关系 $S_d \subseteq S_b$。$S_d$ 的 ID 标识具有新的唯一取值，其它参数和 $S_b$ 的参数相同。

**2、双向生长**：在两个派生概念 $S_{di}$ 和 $S_{dj}$ 之间增加关系来连接它们，同时也对它们所在的网络进行连接生长和概率叠加。步骤是：以指定的基关系 $R_{bi-bj}$ 为模板复制一个派生关系 $R_{di-dj}$ 并设置属于关系 $R_{di-dj} \subseteq R_{bi-bj}$，设置 $R_{di-dj}$ 的两端为 $S_{di}$ 和 $S_{dj}$ 来连接它们，最后对之前影响到 $S_{di}$ 和 $S_{dj}$ 的 PPS 计算先撤销后重做来完成两个网络相互的概率传递和叠加。

**3、单向生长**：和双向生长类似，但此时只有 $S_{di}$ 存在而 $S_{dj}$ 不存在。步骤是：先执行单概念生长完成 $S_{dj}$ 的生长，然后和上述双向生长一样执行 $S_{di}$ 对 $S_{dj}$ 的连接生长和概率叠加。

**树形网的生长**：以 $S_b$ 作为基网络并以 $S_d$ 为子网络的派生网络 $S_D$ 的生长分为两个步骤：

1、对 $S_d$ 的每个元素和 $S_b$ 中对应的元素设置属于关系，完成这些已存在元素的生长。

2、以 $S_b$ 作为模板查找出 $S_D$ 中还未创建的元素 $S_{dn}$，对它们逐个创建来完成对 $S_D$ 的生长，每一步生长根据实际情况选择上述三种基础生长算法之一来进行。

生长的结果是：形成了从 $S_b$ 派生的实例网络 $S_D$，$S_D$ 里包含已知元素 $S_d$ 和新生长的元素 $S_{dn}$，$S_D$ 和 $S_b$ 建立起完整的派生网络关系，$S_D$ 的每个元素都有了新的概率。

**注意：**

1、作为优化，如果新生长的 $S_{dn}$ 里的一些元素的概率低于激活阈值，就不立即创建它们而推后处理。另外，根据任务的特点也可以限制不必要的元素的生长。具体参见后边缩减算法。

2、如果需要新建的目标对象已经存在，就直接进行概率叠加而不重复创建。

3、生长算法可以从网络的任意位置发起并可以向任意方向进行。具体生长方向主要受两个因素影响：首先受任务目标的约定，这里不详细叙述；其次是由概率驱动进行启发式生长，通常选择最高的输入概率优先生长。

4、达到坍缩状态的概念就成为了完全明确的结果，将抑制和它冲突的概念的生长。

**解析、推理、生成等算法本质上都是同一个全向网络生长算法！** 都是根据一部分已知概念来补全整个网络。区别只是初始设定的已知和未知不同而导致起始计算方向不同，在计算过程中随着已知和未知状态的变化计算方向也会动态变化，网络生长不限于单向而是全向进行，各种算法还可以组合起来共同完成任务。以下进行具体解释。

**解析**：根据低层概念推算高层概念的网络生长，也称为自下而上的解析。

例如：
- 在自然语言处理中，根据语言字符串生成内部认知网络。
- 在图像处理中，根据超像素生成鼻子、眼睛、脸。
- 在图像处理中，根据鼻子或者眼睛生成脸。

**生成式解析**：根据高层概念推算低层概念的网络生长，也称为自上而下的解析。

解析任务的目标同样是完成整个网络，只是已知信息通常是一些低层概念。根据全向生长原则，解析任务的执行过程中高层概念的概率也会逐步提升成为已知并发起自上而下的生长来生成低层概念，这种"生成"的整体目标仍然是完成解析任务。

**横向生长**：横向生长是基于相邻关系、转换关系、因果关系等横向关系进行的生长，和上述纵向的解析和生成式解析计算没有本质差别，并且也是双向的生长。

例如：
- 根据眼睛和鼻子的相邻关系进行相互生长。
- 根据加法和减法的转换关系进行相互生长。
- 根据事件的因果关系进行相互生长。
- 根据移动关系进行移动运算。

**生成**：生成算法是自上而下和横向的单向生长，通常由语言生成、图像生成等任务触发。和生成式解析不同的是，生成算法中高层概念是已知而低层概念完全未知(P=0)，从上而下的生长是单向的"纯"生成。

生成算法的具体实现是根据已知的高层概念结合向下和横向的连接来生成低层概念，结果由关系的条件概率等参数以及任务设定的其它条件来决定。因为低层没有已知概念作为约束，单向生成的可能结果会比较多。例如：画一棵树，100个人会有100种不同画法，都同样有效。

例如：
- 在自然语言处理中，根据高层级认知网络扩展生成低层级认知网络。
- 在自然语言处理中，根据内部认知网络生成语言字符串。
- 在图像处理中，根据大的概念生成小的部件，例如：根据脸生成鼻子、眼睛。
- 在图像处理中，根据脸、鼻子、眼睛等概念生成超像素。
- 在行动规划中，根据大的任务目标生成行动计划。

理论上，DC Net的结构和算法可以支撑完整的生成任务，但在具体实现中，更优化的方案是：智能系统负责较高层级的场景和数量较少的大粒度概念的生成，低层的细节生成和渲染处理则由专业系统来完成。这体现了前述的"智能系统+外围系统"方案的意义。

**补隐藏和省略**：各种任务处理里都有不具备直接可见形式的隐藏对象，例如图像处理里有大量被遮住的部分，自然语言处理里同样有大量的省略。对这些对象进行补全同样体现为生长算法，都是依据已知概念来推算需要补全的未知概念。同样，这种推算也可以在多个方向上进行。

**纠错**：纠错和补省略没有本质区别，只是目标对象拥有错误的已知信息，因此进行纠错需要依据的正确已知概念更为强势(概率更高)，才能触发采用计算的结果作为正确结果并强制废弃掉错误的已知信息。

几种典型的生长算法的示意图参见图4。

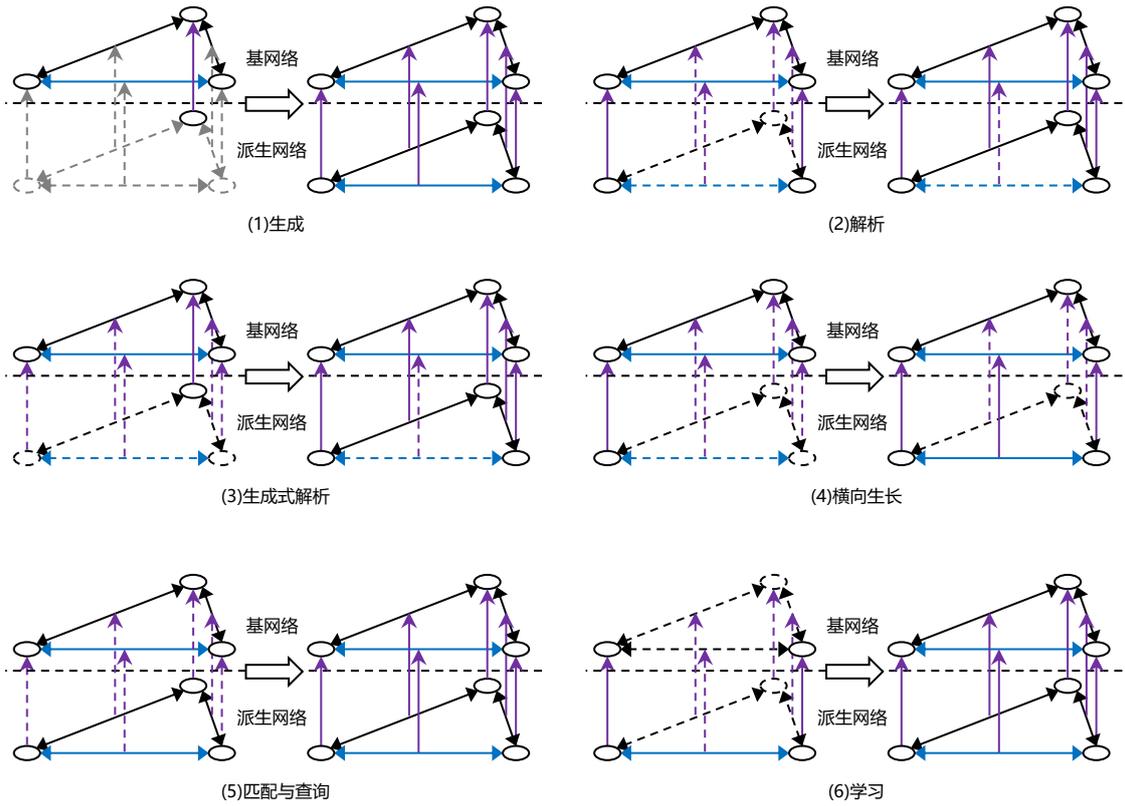

实线：已经生长的元素；虚线：客观存在并且未生长的元素；灰色虚线：客观不存在的元素；

黑色：概念和纵向关系；紫色：属于关系；兰色：横向关系；

图 4：典型算法示意图

以下举一个生长算法的简单例子。具体任务是对图 5 的一张脸进行识别，为了简单起见，只涉及一个层级的计算并只采用了自下而上的解析和自上而下的生成式解析处理。

初始条件：
- 前一阶段已经根据[眼睛拥有图像]等关系(绿色连线)计算得出了眼睛、鼻子、嘴、耳朵、脸、鸡蛋、耳朵、杯柄等的概率，作为本次计算的输入概率。脸和鸡蛋，耳朵和杯柄是相互冲突的歧义解释，存在互斥关系(红色连线)。
- 本轮计算依据[脸拥有鼻子][脸拥有眼睛][脸拥有嘴][脸拥有耳朵][杯子拥有杯柄]等关系，它们的双向条件概率都设置为：P(B|A)=1, P(A|B)=1。
- 假设根据距离、角度等参数对各关系计算的关系隶属度都满足 R(AB)=1。
- 坍缩阈值设定为 0.9。

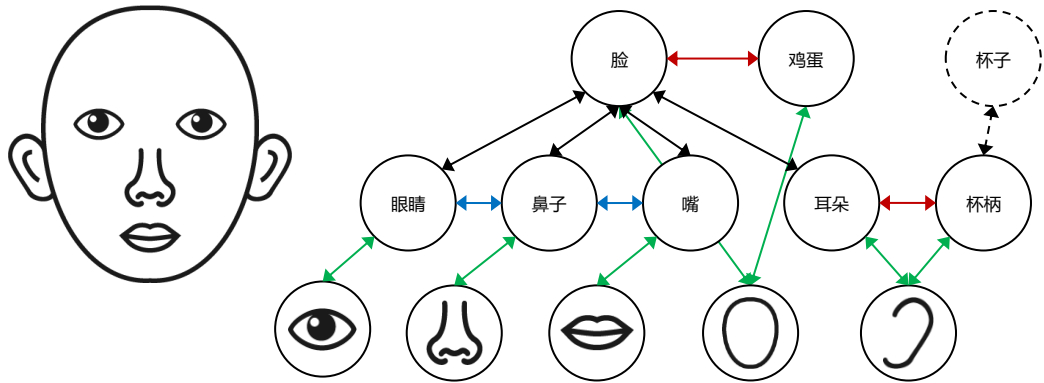

| 步骤 | 眼睛 | 鼻子 | 嘴 | 耳朵 | 脸 | 鸡蛋 | 杯柄 | 杯子 |
|---|---|---|---|---|---|---|---|---|
| 0、初始条件 | **0.6**,0.6 | **0.5**,0.5 | **0.4**,0.4 | **0.1**,0.1 | **0.3**,0.3 | **0.5**,0.5 | **0.4**,0.4 | 未创建 |
| 1、眼睛发起计算 | **0.6**,0.6 | **0.5**,0.8 | **0.4**,0.76 | **0.1**,0.64 | **0.3**,0.72 | **0.5**,0.5 | **0.4**,0.4 | 未创建 |
| 2、鼻子发起计算 | **0.6**,0.8 | **0.5**,0.8 | **0.4**,0.88 | **0.1**,0.82 | **0.3**,0.86 | **0.5**,0.5 | **0.4**,0.4 | 未创建 |
| 3、嘴发起计算 | **0.6**,0.88 | **0.5**,0.88 | **0.4**,0.88 | **0.1**,0.892 | **0.3**,0.916 | **0.5**,0.5 | **0.4**,0.4 | 未创建 |
| 4、脸坍缩的计算 | **0.5**,1.0 | **0.5**,1.0 | **0.4**,1.0 | **0.1**,1.0 | 1.0(坍缩) | **0.5**,0.5 | **0.4**,0.4 | 未创建 |
| 5、结果 | **1.0**(坍缩) | **1.0**(坍缩) | **1.0**(坍缩) | **1.0**(坍缩) | 1.0(坍缩) | **0.5**,0.5 | **0.4**,0.4 | 未创建 |

正常数字：结果概率； 粗体数字：已执行 PPS 的输入概率； 粗体带下划线数字：未执行 PPS 的输入概率；

图 5：网络生长计算示例

按照上述初始条件执行的计算步骤表达于图 5 的表格中，计算过程解释如下：

- 第 1 步、因为眼睛的输入概率 0.6 最高，所以由眼睛首先发起计算：以眼睛的输入概率 0.6 结合关系[脸拥有眼睛]对脸进行解析，传递给脸的概率=0.6×1=0.6，叠加后脸的概率=0.3+0.6-0.3×0.6=0.72；然后用传递给脸的概率 0.6 根据[脸拥有鼻子]的关系继续对鼻子进行生成式解析计算，传递叠加后鼻子的概率=0.5+0.6-0.5×0.6=0.8；同理，推算出嘴的概率=0.76、耳朵的概率=0.64。
- 第 2 步、现在鼻子的输入概率最高，由鼻子发起计算，使脸、眼睛、嘴、耳朵的概率增加。
- 第 3 步、由嘴发起计算，使脸、眼睛、鼻子、耳朵的概率提升。其中脸的概率达到了 0.916，成为了坍缩态。
- 第 4 步、对脸进行概率坍缩处理，将脸的输入概率设置为 1.0 后对眼睛、鼻子、嘴、耳朵进行概率传递叠加计算，因为条件概率都设定为 1.0，所以使眼睛、鼻子、嘴、耳朵的概率都达到了 1.0，成为了坍缩态。
- 第 5 步、鸡蛋、杯柄和已经坍缩的脸和耳朵存在互斥关系，被概率抑制而无需计算，其它对象全部达到坍缩，计算结束，坍缩的对象构成的网络成为最终结果。
- 根据按需选择原理，每次计算涉及的概念、关系和计算路径并不需要保证唯一。这个示例中的计算路径只采用纵向关系，另一次计算可能会采用不同的计算路径包括横向关系。无论如何，如果认知知识的定义和概率设置是合理的，不同的计算路径得到的最终结果也将趋于稳定。

根据这个例子对算法的主要特点做进一步的解释和分析：

**相互推算：** 每一次单向计算都根据一个概念对其它概念进行 PPS 计算增加后者的概率，所有概念相互进行计算。例如脸和鼻子两者进行直接的相互推算，而鼻子和嘴通过脸进行间接的相互推算。对各个概念的推算合并起来的最终目标是推算并完成整个网络。

**全向生长**：示例任务的总体目标是解析，但具体的计算和生长并不限于从下往上进行，也会从上往下进行。在第 2 步脸达到坍缩前，眼睛根据[脸拥有眼睛]关系的 P(A|B)对脸进行了自下而上的解析，而脸增加的概率同时也通过[脸拥有鼻子]等关系的 P(B|A)对鼻子等进行自上而下的生成式解析；在第 4 步，脸达到坍缩后用因坍缩而额外增加的输入概率发起对眼睛、鼻子、耳朵的生成式解析；此外，也可以根据横向关系进行横向的双向生长。

**概率驱动**：全向生长的具体实现是根据任务目标和概率状态来选择最优先的关系和方向，并动态调整。保证每一步计算更为有效，效果不佳甚至没有意义的计算被推后处理，最终可能被完全抑制和取消。

**概率坍缩**：让各个概念的概率迅速增加并达到概率坍缩既是计算的全局目标也是局部的目标。作为局部目标可以使计算任务得到阶段性的确定结果而得以简化。例中第 4 步，脸达到了概率坍缩后导致鼻子等也达到概率坍缩。

**概率抑制**：脸所在的椭圆形状解释为鸡蛋的概率比脸更高，耳朵所在的弧形形状解释为杯柄的概率比耳朵更高。但是眼睛等概念的优先生长将鸡蛋和杯柄的生长推后处理，而脸和耳朵的生长顺利并达到了坍缩态完全抑制了鸡蛋和杯柄的生长机会，根据杯柄创建杯子的潜在计算也就没有执行，避免了无效的计算。另一方面，即使鸡蛋、杯柄先得到了生长的机会，但是没有更多信息的支持从而不能持续生长达到坍缩，那么脸的生长仍然将得到执行并最终坍缩。如果认知知识定义是合理的，即使中间计算过程有变化但最终结果仍将是稳定的。

**多层级计算**：这个例子只展现了脸和眼睛等概念的两级计算，扩展到多层级计算合生长的原理也完全相同。向上可以对脸根据[人拥有脸][问拥有问方]等关系来逐级推算并生长出人、问等更高层的概念，向下也可以从眼睛推算并生长出睫毛、瞳孔等更低级的概念。总之，根据不同的已知条件和任务目标，整个认知网络会选择不同认知组件组合起来向不同方向进行生长，所有的生长都依据相同的概率计算模型。

传统智能算法通常都是单向计算，并且将解析、生成等计算划分为完全不同的计算任务，对模式识别通常定义为自下而上的单向解析计算。近来，对自上而下的解释的重要性也开始得到重视[7]，两个方向的解释都要兼顾的观点也被提了出来[8]。DC Net 里的全向生长则是更完整且具有具体实现的方案，现在，我们有了一个将多种任务统一起来的全向生长模型[1]。

我们相信人脑的思维也采用类似的方式：计算方向总是根据已知条件和中间计算结果的概率变化不断调整优化，并且抑制掉大量概率较低的不必要计算。人脑对于图像识别等有如此的高效性(目前显然比 AI 高效)原因就体现在这些关键点上。

**按需选用**：例子中，根据耳朵拥有的 0.1 的输入概率结合[脸拥有耳朵]关系对脸的计算并没有执行，因为脸根据其它信息已经达到概率坍缩，只需要根据 P(B|A)让脸从上往下对耳朵进行生成式解析，使本身输入概率仅为 0.1 的耳朵完成概率坍缩。概率驱动的全向生长方法实现了"全面表达、选择使用"这一重要原则。

这里对"全面表达、选择使用"进行一个更深入的讨论。假设描述猫这个概念用了数十个概念：头、腹部、背部、左耳、右耳、左眼、右眼、鼻子、嘴、尾巴…。但根据认知概率模型实际上只需要得到 4、5 个部件的已知信息就足以使猫达到概率坍缩，所以每次识别猫的计算只需要使用部分部件。

并且也只能使用部分部件，因为不可能同时获得猫的全部信息，并且不同时刻能获得的信息都不相同。因此，我们认为：按照独立同分布和参数共享的假设训练一套权重并用固定的网络来应付所有情况并不合理，会产生大量无效计算且计算结果并不稳定。

DC Net 的认知概率模型和生长算法适用于这种按需选用的认知计算：

---

1 事实上，全向生长模型将判别模型和生成模型也进行了统一。

- 条件概率基于独立采样假设，概率叠加适用于选择部分信息进行计算；
- 动态生长算法保证需要的知识才会被创建和计算，这是很重要的优化；
- 概率坍缩、概率抑制等机制减少了无效的计算，实现了更充分的优化；

人类智能本身也遵循这种模式，人脑记忆了大量的知识模块，它们具有不同的视角并适用于不同场景，在不同情况下人脑会选用不同的知识模块来解决特定的问题：不仅可以根据不同的身体部件来识别猫；也可以根据听到的声音来识别猫；还可以根据脚印来推测猫；甚至运用[向别人询问"你抱着的是猫吗？"]这样的知识模块来解决识别猫的任务。

当然，虽然在每个具体任务中只需要使用部分知识就足以解决问题，但仍然要尽量完备地构建整个知识体系，以保证在不同场景下高效地选择正确的知识。

"知识越多歧义越多"的悖论也是人工智能的经典难题。这里给出一个解释：每一条知识都有适用范围，没有一条知识的概率绝对恒等于1，对知识设定绝对适用和绝对概率的假设就会出现大量冲突。在不同场景下选用最合适的知识来解决问题是人类智能的关键，也是强人工智能的正确方向。这要求每条知识要有适用性场景和合理的概率取值，并用明确的结构形式表达出来，在这样的体系下，系统的知识越多就越智能—正如人脑的表现。

### 3.3.4. 缩减

网络生长不能无限制地进行。生长的目的是为了完成任务，任务目标和实际条件都可以限定网络的生长边界，对于已经完成生长的网络也可以裁剪删除进行简化。

**截止**：截止就是生长到一个边界而终止，不同方向上的生长都可以被截止，具体分为两种情况：一种是生长结果已经能满足任务目标而进行截止；另一种是已知信息不足以支撑继续生长而截止，例如图像的精度限定了能识别的最小概念(当然，如果相机具有变焦放大的能力也可以增量生长提升细节—就像人眼一样。显然，这也必须围绕动态认知网络结构为中心才能实现。)。

**裁剪**：对已经生长的部分网络进行裁剪以节省空间。基本原则是保留更重要的信息而裁剪相对不重要的信息，通常，更大颗粒的信息以及任务目标紧密相关的信息更重要。例如：识别出了猫以后，裁剪掉识别猫时创建的头、尾等低层概念而只保留根概念。

### 3.3.5. 推理和迭代

将多级的网络生长和缩减结合起来就可以实现多级推理和迭代计算。

在 DC Net 中，向上、向下、横向等不同方向都可以进行多级生长，所有的生长都是广义的推理，通常时间相关的多级横向关系生长具有较长的生长序列，可以称为狭义的推理。各种推理都是双向的，唯一的影响因素就是概率，概率大的方向的推理更有效而更优先，概率很小的方向就无法推理。

迭代通常是变化、移动等横向关系的生长，这些关系的两端的树形网结构完全相同，只体现出时间、版本等参数的变化。如果每一次生长都只保留最新状态的树形网而裁剪掉旧的树形网，就体现为不保留历史版本的"纯"迭代。

### 3.3.6. 网络流转

从宏观层面来看，各种算法组合起来形成了"网络流转"的综合效果。

- 完成一个树形网：每一个基本计算单元都是完成一个树形网。首先这个树形网被构建起来并处于概率叠加状态，然后结合各种信息提升概率并完成树形网补齐，成功以后树形网将完成生长并达到概率坍缩。
- 任务转移到新的树形网：从已完成的树形网出发，根据任务的需要可以从根节点向上生长更高层的树形网，也可以从内部节点向下生长更细节的树形网，或者横向生

长出其它的树形网。
- 删除不需要的树形网：如果历史信息不再需要，就可以收缩或者完全删除无用的树形网。

观察整个过程：在任务目标和计算结果的驱动下，不断形成新的网络；网络不断生长；网络从叠加态转为坍缩态；无用的网络被缩减，这样形成了连续的网络流转计算，拟合了目标场景的动态变化。

我们认为，这样的模式和人脑的思维方式是比较接近的。树形网的构建对应着人脑里一组神经元的激活，树形网的删除对应着神经元恢复正常状态，网络流转就对应着神经元激活状态的不断转移—即连续的思考。

### 3.3.7. 查询

在 DC Net 里，实现信息查询任务可以划分为两部分：

**查询匹配**：查询的基本处理是以问题任务产生的查询模板作为基网络去知识网络库中执行网络匹配算法，匹配成功的结果就输出为查询结果。这种匹配通常要求完全匹配而不是部分匹配，即查询模板包含的所有参数都要得到匹配满足。

**转换推理**：复杂的查询任务需要对信息进行推理甚至计算等处理，具体就是选择各种生长算法来进行多级推理，这种转换推理的每一个步骤都可以进行展现和解释。

### 3.3.8. 外围实现

不同层级的概念和关系的语义实现分为内部实现和外围实现两种方式。

**内部实现**：DC Net 各种基本概念和关系的语义形成自描述的体系，内部较高层级的概念和关系进行相互描述，并通过内置的基本算法来实现相互计算。在系统成熟稳定后，就只需要通过手动或者自动学习来增加和调整认知网络结构实现系统的扩展，而不再需要改变程序的算法和代码。

**外围实现**：实现最基础的语义需要专门的外围模型和外围算法，这些模型和算法的数量是有限的。

内部实现和外围实现需要进行对接，后者为前者提供最基本的语义实现支持，两者之间的界限根据技术的发展而调整优化，举例分析如下：

**自然语言处理**：自然语言处理并不简单视为串对串的转换处理，而体现为两个部分：主体部分是对内部认知网络的生长处理，这部分其实和具体语言无关，就是各领域的通用认知计算；语言处理只是少量的次要工作,体现内部认知网络和语言形式之间的解析和生成转换，其中，字符串和语言角色等语言形式也都是 DC Net 的组成元素。因此，自然语言处理主要由 DC Net 的内部算法实现，外围实现较为简单。

**空间场景和图像处理**：空间场景与图像处理是很基础性的工作。其中，按照 DC Net 的原则设计一个可伸缩的高效三维空间认知模型是最重要的工作之一。

图像识别和场景重建的基本实现框架如下：先由外围系统采用 SLIC 和卷积等基础算法对图像进行处理形成带有特征信息的超像素作为初级元素，然后再通过网络生长形成完整的二维和三维空间场景。

不要求外围系统生成的初级元素绝对正确(概率坍缩态)，允许它们存在重叠和歧义(概率

叠加态)。这种歧义由后续的网络生长来解决，具体是按照全向生长和概率优先原理，选择更为合理的初级元素逐级生成最终场景，错误的初级元素就自然地被舍弃。自然语言处理中词语一级的分词和歧义也是同样的原理解决，DC Net 的生长算法可以避免单向流水线式的处理算法中前一阶段的错误传递到后一阶段被放大扩散的难题。

**行动处理**：行动也是一种生成和输出。对网络生长算法生成的行动部件进行输出就是执行行动。要点是需要对行动部件分解细化到外围行动系统能接受和实现的颗粒度。

### 3.3.9. 机器学习

首先对学习进行一个基本定义，其中的要点是知识和数据这两者的关系。

DC Net 在集合维上也是多层级结构，其中更高层的网络更抽象，更偏于知识，更低层的网络更具体，更偏于数据。

应用任务要求高层的知识网络已经存在并将其作为依据和模板，对派生的实例网络进行生长处理，在集合维里体现为从上而下的生长。

学习任务是在不存在正确的知识网络情况下，根据基本先验知识直接创建实例网络和知识网络，并对知识网络进行归纳和优化，在集合维里体现为从下而上的生长。

认知知识包含两个方面：主要方面是概念和关系组成的网络结构；次要方面是结构里包含的条件概率等参数。显然，**"结构重于参数"**，学习任务首先要产生有效的认知结构，其次再对认知结构的参数进行调整优化。

机器学习算法也体现为网络生长算法，并且和应用任务紧密融合，基本原理是：
1、执行应用任务：即根据知识网络构建实例网络进行场景拟合。
2、通过网络匹配等算法计算实例网络和场景的误差来判断结果的正确性，误差低于标准的实例网络视为正确结果，误差超标则视为错误结果。
3、如果无法根据已有知识网络构建出正确的实例网络，那么就认为是缺少必要的知识网络，需要尝试创建新的知识网络来解决—触发学习算法。

在任务过程中，如果产生了多个有效的实例网络，那么应该选用哪一个呢？这引出了一个重要的基础理论。

**认知网络最优化原理**：在使用认知网络进行信息表达时，在保证误差达标的前提下总是试图采用最优化的网络结构进行信息表达。最优化的网络可以粗略地定义为动态结构最小的网络，而网络结构的大小可以用网络包含的概念和关系的总数来度量。

可以认为，人脑结构的进化很大程度上遵循了这个最优化原理。人脑虽然很强大，但其容量(神经元和连接的数量)终究是有限的，要处理和记忆近似于无限的信息就要进行简化处理，采用尽量小的结构来记忆尽量多的有用信息。这种简化处理会丢失部分信息，但只要保证关键信息基本正确和完整就可以满足要求。

DC Net 中，树形网和派生网结构就是体现这个原理的具体体现。树形网体现了对单个概念的组合简化，派生网体现了对多个相似概念的泛化简化。

认知网络最优化原理涉及很多内容，这里不进行深入分析，仅以前述例子简单说明：
一只猫总是由头、身体、爪子、尾巴等部件组成，因此，用[头+身体+爪子+尾巴]这个网络就可以完整地描述信息,如果只考虑信息的完整性那么"猫"这个概念根本不需要存在。但这个网络的大小是 4，每一次描述类似的信息都需要重复这个 4 个元素的网络而造成信息存储和传递的资源浪费。因此构建了一个猫的概念和树形网[猫：拥有头、拥有身体、拥有

爪子、拥有尾巴]，这个网络的大小是5，定义这个网络消耗了额外的空间，但以后遇到类似信息就可以只使用"猫"这个大小为1的概念来表达而节省空间，头、身体等部件不需要显式表达，可以根据树形网里定义的知识和概率推算出来，信息并没有丢失。

优化表达以引入了误差为代价，"猫"的定义只是一个树形网，为了起到优化效果就不会记忆每一只猫的完整信息而是记录平均的信息，例如记忆了[猫：拥有平均体重3公斤]，在表达每一只猫时，如果它的实际体重和平均体重的误差在允许范围则没有问题，如果误差超标例如出现了一只50公斤重的猫，就违背了误差要达标的首要原则，需要进行处理，也就是要进行学习和创建新知识，例如：派生创建出一个"巨猫"新品种。

人类的语言交流在保证信息有效传递基础上进行优化是语言学的基本原理之一，认知网络最优化原理表明内在的认知描述同样遵循类似的优化原理，并且更为基础和本质。

认知网络最优化原理是认知网络结构的设计原则之一，并作为机器学习的重要先验知识和目标约束。基于这个原理实现的认知网络学习和端到端的拟合学习有显著的不同：
- **端到端的拟合学习**：设定一个相对固定的初始结构；采用反向传播等算法来调整参数使计算出的残差不断变小；学习以追求末端的残差最小为目标；
- **认知网络学习**：并不设定一个固定结构，而是学习不同的结构来拟合场景；误差不作为拟合目标而作为每一步计算的基本原则，即规定误差不允许超标；学习以追求最优的结构为目标。

其中，误差的具体定义和计算和特定领域和任务相关。例如：对于自然语言处理来说，误差主要体现为实际字符串和目标字符串的偏差；而对于图像识别来说，形状、尺寸、纹理等是进行误差计算的主要指标。各种误差最终都转换为隶属度来进行统一评估并和概率计算体系相融合。

前边提到了结构学习面临的主要困难体现为目标结构如何定义以及要求知识可微的问题。在 DC Net 中，这些困难可以在基本理论上得到解决：
1、定义了树形网等基本结构，学习到的知识要符合基本结构，这本身就是有效的先验知识和约束条件。
2、认知网络最优化原理为学习到的认知结构提供了重要的评估依据和拟合方向。
3、结构学习并不要求可微，可微的参数学习只是结构学习的扩展。
4、可以结合具体领域的各种先验知识—包括人工输入的知识。
将上述原则整合起来就为实现面向结构的认知网络机器学习奠定了基础。

**认知网络学习算法(CNL)**：基于上述理论设计出认知网络学习算法的基本框架，算法假设已知知识和待学习的知识混合一起，用已知来学习未知。典型的实现步骤如下：
- 学习算法以动态认知网络为中心并在领域维和集合维上同时进行，分别围绕树形网和派生网络结构实现学习。
- 1、初始状态：在领域维上的低层 i 层上已经有了前一级处理提供的多个实例概念 $S_{di}$(例如：图像处理里的超像素)。此时无需保证这些实例概念绝对正确(概率=1)，而是尽量获取所有可能的实例概念，它们各自具有一定的概率。
- 2、执行生长：按照应用任务的要求执行网络生长，如果生长成功，也就是在 j 层上匹配了基网络 $S_{bj}$ 并派生创建了实例网络 $S_{dj}$ 进行生长并吸收了 $S_{di}$，表示已有知识可以解决遇到的数据，无需学习新的知识。
- 3、创建假设知识：如果上述生长失败，原因就是 j 层缺少应有的基网络，所以尝试创建新的基网络 $S_{bj}$：具体方法是选择部分 $S_{di}$ 并结合必要的先验知识(图像识别

里相邻关系是非常重要的先验知识)来创建新的知识树形网 $S_{bj}$ 和实例树形网 $S_{dj}$，它们包含必要的纵向关系和横向关系。最终可能产生多个 $S_{bj}$，它们都是假设的新知识。

- 4、初始化参数：首先标记每个假设的新知识为[新知识]；然后对其参数和概率进行初始化赋值，默认的设置是用每个实例概念 $S_{di}$ 的参数对 $S_{bj}$ 的对应概念的参数赋值，而条件概率都初始化为 1.0，统计数量初始化为 1。如果有适用的先验知识，则根据这些先验知识来进行参数和概率计算并赋值。
- 5、继续生长：将假设的新知识网络 $S_{bj}$ 和其派生的实例网络 $S_{dj}$ 加入体系中，然后跳转到 2 继续执行网络生长，直到所有 $S_{di}$ 都被吸收生长到一个或多个完整的实例网络 $S_d$ 中。
- 6、结果筛选：从所有实例网络 $S_d$ 中选取整体达到坍缩状态的网络为最终结果。如果有多个 $S_d$ 都坍缩则按照认知网络最优化原理选取结构较小的 $S_d$ 作为最终结果。最终结果确定后其包含的新知识网络也就确定了。
- 7、统计累积：学习创建的新知识加入到体系就立即可以应用，后续更多任务的执行都会触发对这些知识的应用，并记录每条新知识被成功应用的统计数量，根据统计数量可以计算出条件概率等量化参数，条件概率较大的新知识最终确定为正确的知识保留，双向条件概率都很小的知识则被丢弃。
- 8、合并相似：在集合维上将新学到的知识和已有的知识进行比对，对于相似的知识进行合并，遵循认知网络最优化原理进行结构优化。

概括一下 CNL 算法的核心要点是：
- 用已有知识解释数据并执行生长，要求误差不超标。
- 解释不成功时，遵循**"误差和错误意味着发现了新知识"**的原理，执行**"依照数据创建新知识"**。
- 根据基本结构定义、认知网络最优化原理、领域先验知识等约束在创建和应用过程中去除不合理的知识，留下最合理的知识。
- 通过统计计数进一步筛选更合理的知识，并调整条件概率等参数。
- 对知识进行合并优化。

CNL 学习算法还具有这些特点：
- **小样本学习**："单样本学习结构，多样本学习概率"，根据单样本就可以学习到结构，多样本学习则是对学习到的结构部件进行统计计算并调整概率等参数。对于大量概率≈1 的知识(例如：猫拥有猫爪子，$P(A|B)≈1$，$P(B|A)≈1$)来说，单样本学习到的知识就足够准确，根本不需要大量样本。
- **实时学习**：学习和和应用总是同时进行，两者不会绝对分开。实际上，总是应用的需要推动新知识的学习。
- **独立参数学习**：概念的值体现了对概念自身的划分；概念的概率则体现了概念出现的统计频次比，两者的界限清晰。概率模型基于独立采样假设，各条知识和条件概率彼此无关，可以独立学习并独立调整。相对而言，很多机器学习将参数视为黑盒式的整体结构，对参数的调整会相互纠缠和影响。
- **增量式学习**：学习总是根据已知知识来解释和学习新知识，理想的情况是每次学习少量知识，增量式地学习。例如：体现在上述算法中，i 是相对低层而不需要是最低的第 1 层，意味着如果低层有正确的知识能正确解释到第 i 层，新知识就只需要在 j=i+1 层上开始学习。另外，学习到 j 层的新知识如果在第 k 层能被已有知识进

行解释生长，就不需要在 k 层上继续学习。并且，一个树形网的完整信息也是通过增量式学习来逐步构建的，例如：第 1 次见到猫是从左侧，我们构建了猫的树形网并学习到了关于它的左耳朵、左爪等知识；第 2 次再见到猫是右侧，则可以运用已有的知识增加关于它的右耳朵、右爪等新知识。

显然，这些特点也符合人类学习知识的方式。

当然，实现强大的无监督机器学习是一个艰巨的任务，以上描述了面向动态认知结构的机器学习的基本原理，具体实现还需要大量的细节工作。

### 3.3.10. 连续计算

复杂的智能任务需要结合多种不同的子任务和算法来实现。动态认知网络的算法体系具有强大的多任务融合能力，对场景进行拟合的认知网络作为中心结构被各个子任务共享，存在于整体任务的整个生命周期甚至更久，各个子任务不进行相互对接而分别和这个中心结构对接并协同对它进行各种处理，各种子任务可以动态地组织并连续地进行，甚至随时调整和追加任务。例如：

- 在图像识别产生的空间和物质结构上，聚焦放大细节，或者扩展范围进行拼接识别。
- 在图像识别产生的空间和物质结构上进行图像的生成和创作任务。
- 在对话过程中可以动态地植入等翻译、对话、情感分析等不同任务。
- 调出历史任务并无缝地恢复执行。

过程中构建的认知网络的完整程度体现了目标场景的范围和深度，直接决定了任务的智能水平。例如：在一个客服场景中，处理的目标并不限于双方的话语，而要扩大为描述整个场景的完整认知网络，这个认知网络在过程中持续生长并将所有实体、对话、推理、思维和过程都组织在一起，这样才能真正准确地理解和处理对话。同时也为实现灵活而全面的智能任务奠定了基础：不仅可以让机器扮演客服人员，还可以用相同方法和相同的知识统一地实现场景理解，信息收集，执行翻译，扮演客户助理等不同任务。

连续计算不仅能实现强大的智能化功能，也能有效地提升性能。举例来说：和目前的 AI 系统相反，人类对动态场景的识别比对静态图像的识别反而更高效，原因是在动态场景中对每一帧图像的识别并不总是从零开始，因为前边已经识别并构建起了包含大量已经坍缩信息的认知网络，处理新的一帧图像只需要在已经明确的认知网络上进行少量的增量式处理，这样就具有极高的效率。如果 AI 系统能运用同样的方法，那么它执行任务的完整性，准确性以及性能就有望追上甚至超越人类。

显然，能比拟人类智能的强人工智能必须同样能执行连续的任务，而不应该是割裂的碎片化智能。对连续计算的分析进一步证明：独立于计算并动态生长的中心结构是实现强人工智能的关键！面对开放领域，这个中心结构必然是类似 DC Net 这样的跨领域通用认知结构，针对特定计算任务的专用数据模型和固定网络结构很难实现同样的目标。

## 4. 深入讨论

本章对动态认知网络的几个要点进一步深入分析，并对人脑中可能存在的类似逻辑结构和机制进行探讨。

### 4.1. 树形网是认知的基本结构

按照前边分析，树形网结合了树的问题分解能力与网络的全面信息表达能力两个特点，

成为非常适合进行认知表达的基本结构,各种复杂场景都可以通过各种树形网的灵活组合和嵌套来进行拟合。

据此可以对人脑运作机制的一些原理进行推测。对人脑结构进行研究的大量成果中可以看到人脑神经元的组织呈现层级结构,并且具有向上、向下、横向等各个方向的连接,但这些结构及其信息流动如何构成完整的思维体系仍是一个疑问。我们提出一种假设：人脑也大量采用类似树形网这样的逻辑结构来进行认知表达和计算。

人脑中和 DC Net 中的概念和关系精确对应的元素和机制还不能明确,可以确定的是,人脑不会用一个神经元而至少要用一组神经元(例如垂直柱)来表达一个概念,如果把这样的一组神经元称为一个神经单元,那么表达一个树形网(例如脸)的结构如下：

- 用一个神经单元表示树形网的高层概念—例如脸；另一些神经单元表示树形网的低层概念—例如眼睛、鼻子。
- 用从上到下的纵向连接将高层神经单元连接到低层神经单元；也用从下而上的纵向连接将低层神经单元连接到高层神经单元。
- 同层级的神经单元之间也有很多横向连接—例如表达眼睛和鼻子之间的相邻关系的横向连接。

如果从上往下的树形网结构是人脑对世界进行描述认知的主要先验结构的假设成立,那么可以对其最基本的学习机制进行一个推测：人脑一开始就准备好了很多树形网结构的雏形,其中包含各种潜在的连接,在遇到新知识时就建立和加强正确的连接来完成学习,学习的目标就是形成描述各种具体知识的树形网结构。

树形网结构中,由短距离横向连接体现的相邻关系非常重要。这种近距离的相邻关系是最为基础的先验知识之一,人脑的认知倾向于将相邻的概念结合在一起,对于空间和时间都是如此,空间上的相邻关系对应静态结构的构建,时间上的相邻关系对应因果关系的构建。学习时,通常首先由低层概念之间的相邻关系触发横向相邻关系的构建,然后基于树形网的先验结构假设可以推断出：横向关系的出现就意味着存在对应的纵向关系和整个树形网,因而同时触发了高层概念和纵向关系的创建,一次性地构建起整个树形网结构。

因此,人脑进行学习和应用的基本单元都是树形网。学习知识是对一个树形网相关的一组连接同时进行构建；应用知识也是以树形网为单位进行,每一个树形网都尽力激活自己,树形网之间相互竞争,最终胜出的各个树形网组合嵌套形成正确的场景表达。

### 4.2. 生成为主体的双向连接

树形网由概念和双向关系构成,双向关系的两个单向连接都可以进行概率推算,但两者对树形网的作用存在着本质的不同,对于连接高层概念和底层概念的纵向关系来说,自上而下的 A=>B 连接(也可以称为生成连接)是主体连接,而自下而上的 B=>A 连接(也可以称为解析连接)是辅助连接,分析原因如下：

- **生成连接描述了树形网的本质**：树形网总是从上往下组织,从一个树形网的高层概念向低层概念的 A=>B 连接及条件概率等参数完整地表达了树形网自身的知识,这种知识和其它树形网无关,一个树形网的多条 A=>B 连接是相容的,并且通常都具有较大的条件概率。另一方面,一个低层概念可能解析为多个高层概念,存在着多条 B=>A 连接,但这种结构并不是树形网,这些 B=>A 连接相互互斥,每一条连接通常具有较小的条件概率。
- **向下的生成远多于向上的解析**：如前所述,一个解析任务通常分两个方面,一方面根据部分低层概念从下往上解析高层概念；另一方面根据高层概念向下进行生成式解析来推算所有低层概念。一个高层概念对应多个低层概念—例如前例中猫概念由 100 个部件构成,按照前述的概率传递原理,1 个部件向上的解析将引起高层的猫概念对其它 99 个部件的向下生成解析；更进一步,一部分低层概念就可以使高层

概念达到概率坍缩，后续更多的工作就完全由高层概念向下生成，例如用 5 个低层概念就能使高层概念达到坍缩，剩下 95 个低层概念都是根据高层概念向下生成。

- **向下的生成具有更高的效率**：假设高层概念 A 本身已经具有坍缩概率 1，那么向下生成 100 个部件完成整个树形网就是正确的目标，每一步生长都是有效的。而低层概念 B 向上解析时，即使 B 也具有坍缩概率 1，但是对应 N 个可能的高层概念时只有 1 条是正确的，其余 N-1 条都是错误而无效的。因此，即使对于解析任务也总是偏向更多采用自上而下的生成计算。在初期阶段先对低层概念进行一定数量的向上解析计算，一旦推算出具有一定概率的高层概念后，就会尽快转向从上到下的生成来完成整体解析任务。

因此，自上而下的生成连接是描述树形网的主体连接，也是计算的主体。自下而上的解析连接则是一种辅助结构。整个树形网的生长只有少部分工作从下而上解析，大部分工作都是从上而下生成出来。

同样，我们对人脑里实现类似双向关系的结构和机制进行假设和分析。我们认为，人脑里也大量采用类似动态认知网络里的双向关系，只是因为生理机制决定了人脑神经元的信息传递基本使用单向信息通道，信息总从一个神经元的轴突传送到另一个神经元的树突(虽然研究表明存在着沿轴突向神经元胞体的反向传播动作电位[9]，但我们认为它不是一种主要机制)，因此实现两个神经单元的双向关系需要采用两条独立的单向信息通道来实现，实际的信息通道结构非常复杂，忽略复杂的细节，我们可以粗略地认为：A 神经单元的轴突的一个分支直接或者间接地连接到 B 神经单元的树突，而 B 神经单元的轴突的一个分支直接或者间接地连接到 A 神经单元的树突，连接的强度大体上对应于关系的条件概率。

相比之下，DC Net 可以仅采用一条关系来承载两个概念之间的双向条件概率体现了计算机比起人脑更优势的一方面，一旦突破智能的奇点后，计算机将在很多方面体现出比人脑更巨大的潜力。

传统上，把人脑里自下而上的连接称为前馈连接，把自上而下的连接称为反馈连接。这种定义应该是仅关注于将外部感知信息解析为内部认知的局部功能的产物，这种观点认为前馈连接是信息传递和计算的主体，反馈连接仅进行辅助性的调整，目前多层神经网络也主要基于这个理念来实现，并且只描述单向的前馈连接和权重，而并没有描述更为重要的反馈连接和参数(反向传播算法就只是顺着前馈连接的反方向去调整权重)。

而根据上述分析我们知道了：被称为反馈连接的自上而下的 A=>B 连接才是更重要的主体，而被称为前馈连接的 B=>A 连接才是到辅助的作用。人脑中反馈连接的数量远比前馈连接的数量更多的事实可以证明这一点，大量的研究也都指出，人脑看到的场景更多是生成的或者说"想象"的，而不是"看"到的，这和上述的观点一致。

并且，前馈为主的观点只是针对初级信息解析这个外围功能研究的结论，思考、规划、行动等更高级的功能直接体现了树形网从高向低和横向的生长才是人脑的主要计算方式。实际上，在完全没有前馈输入的情况下，这些功能也可以自我运转—例如做梦和冥想。

如果上述假设是正确的，那么从对认知起到的实质作用来看，将 A=>B 连接称为"主体连接"并将 B=>A 称为"辅助连接"更为合理。

横向关系和纵向关系没有本质区别，有一些横向关系的两个单向连接可以看着一条主要连接和一条辅助连接，另一些横向关系则可以看着两条主要连接。

### 4.3. 全面准确高效地模拟世界

我们可以认为：智能的终极目标就是运用有限的存储和计算资源，尽可能全面、准确、高效地模拟世界。

世界是一个相互影响、动态演变的整体系统，驱动世界运行的基本因素—力的作用就总是相互影响而非单向影响。对世界更准确的描述需要综合考虑各种相互影响来对整个系统进

行模拟计算。

想象多个天体构成的一个物理系统，其中任意两个天体之间的引力都对彼此产生相互影响。即使我们只关心其中一个天体的状态(位置、运动)，**但是为了计算它的状态必须同时计算其它天体以及整个系统的状态！**所以需要如下数值模拟方法求解：

- 对一个天体求解：根据各个天体对一个天体的引力对它的状态求解。这是一个单向多变量函数计算，其基础是每两个天体之间引力的双向影响。
- 对各个天体求解：用上述单向函数对每一个天体的状态分别求解，合起来就是对整个系统求解。
- 连续迭代计算：对时间进行差分处理来迭代求解下一个时刻整个系统的状态，并随时根据更新的信息进行修正。

这种以整体系统演变为目标的数值模拟方法是人类发明的工具计算的巅峰之作，被应用到众多领域来解决复杂的问题(例如：气候预测，宇宙演化研究，核试验模拟)。人脑的智能认知计算和这种数值模拟计算非常相似！只是和针对特定任务的专业系统相比，智能认知计算的目标是更全面的模拟和计算，面对世界的复杂性和可用资源的有限性，需要在全面性、准确性和高效性几个方面进行平衡因而对模型进行了扩展：

- 实现了层级体系：构造了多层级的概念和结构体系，根据任务的不同选择在最合适的层级上表达和计算更为高效，例如将每个天体作为单一变量进行计算而无需对每一个原子进行计算。
- 扩展了关系体系：根据基本的力来对复杂的世界进行计算过于低效，所以创造了各种关系(拥有、相邻、推导、情感…)来进行表达和计算。这些关系都可以看着广义的力，对两端的概念产生相互影响，各种影响都体现为结合概率的相互计算。
- 边界动态扩展：智能认知计算中，目标系统的范围边界并不是固定不变，而是根据需要动态扩展和调整。

我们认为：人脑神奇地进化形成了这种对世界进行系统模拟的机制，其中双向连接通道提供了对概念进行相互计算并组合成系统的基础；M-P 模型等多种机制构成了近似于条件概率和概率叠加的函数计算组件；树形网结构实现了对可组合嵌套的各个子系统的定义和描述。比起原始生物的神经系统对受到输入刺激到输出的反射行为进行直接处理，这种能用认知结构模拟场景系统的机制的出现是人脑开始具有高级智能的重要标志。

另一方面，模拟世界的全面性和完整性是受限的，在满足任务目标的同时必须考虑效率。需要根据任务目标和实际场景选择更有效的概念和关系来构建系统进行计算，并忽略其它不必要的因素来进行简化。

重点是要保证这种简化的正确性。在开放的世界里，固定的简化假设总会出现问题，在一个场景下正确的简化假设放置在另一个场景下可能是完全错误的。这也就是独立同分布等假设理论不适用于真实场景下的智能计算的根本问题所在。

正确的简化必须是动态和智能的。在一个场合下，一个因素对于计算任务目标的影响不大(体现为概率很小)就可以进行舍弃简化；在另一个场合下，同样的一个因素可能对任务目标影响变得较大就不能忽略而必须纳入体系。

因此，智能计算既是以整体系统为目标的场景拟合计算，同时目标系统本身也不是固定的而是动态创建。**每一次智能计算本质上都是在构建一个独一无二的系统（而不是一个固定系统的一次计算）！**系统正确构建完成后计算任务也就基本完成了。构建时对不同认知结构进行选择或简化的主要依据就是对误差和概率的估算，并且在需要时可以创造新的认知结构(也就是学习)。

人脑根据按需选用原则进行动态系统构建的具体实现机制有待研究，但我们认为其逻辑

效果主要可以采用概率坍缩、概率优先、概率抑制等理论来进行类比解释。

## 5. 总结

  人脑用结构化的方式来看待世界，智能就是构建各种认知结构来对世界进行描述和拟合，实现强人工智能不能回避在各个领域对丰富的认知结构进行精细设计。传统系统或者弱人工智能系统主要用少量狭义域知识对大量场景进行部分处理(小知识、大数据)，更强的人工智能系统首先要实现用大量的广域知识对每一个场景进行全面的处理 (大知识、小数据)，然后再扩大目标场景的复杂度和数量，最终具有强大计算能力的系统将可以对大量场景并行地进行全面处理(大知识、大数据)，体现出将人脑灵活的智能处理能力和计算机强大的迭代计算能力融合在一起的超级能力。

  构建这个广域、完整而灵活的结构体系就是实现强人工智能的根本难题，解决该难题的关键是首先构建起具有对各种结构进行泛化和组合能力的基础体系。本文中提出的 DC Net 模型提供了一种可行的方案，其中两维度多层级结构是重要的机制：在领域维上的树形网结构体现出对广泛领域的多种结构进行动态组合来实现场景拟合的能力；在集合维上的派生网络结构则体现了对知识和数据的抽象层级关系进行泛化处理并为应用和学习合一的计算提供结构支持。

  在这个基础体系之上，对不同领域的各种认知结构不断地构建和完善就能持续有效地提升系统的智能性，这将是今后最主要的工作。

  概率是结构描述不可或缺的构成部分。DC Net 设计了以条件概率和概率加公式为主体并结合双向条件概率描述、网络概率状态、概率坍缩等理论方法的基础认知概率模型，以支撑开放域场景和显著性概率条件下的智能系统的概率描述和处理。并在具体实现上讨论了简化的概率叠加公式的可行性，对类似方案的细化完善以及在局部对概率模型进行扩展实现是今后的重要课题。

  DC Net 采用网络生长算法体系来统一实现各种主要的智能计算。这种算法体系遵循全面表达、按需选用的理念，以概率驱动的全向网络生长来动态构建认知网络，体现出任意变换输入输出、连续计算、可解释性等智能计算所需要的能力，比起单向函数计算更适合作为实现更强大的 AI 的基础算法。

  认知网络学习算法是以结构学习为主、参数学习为辅的机器学习算法。这种算法在场景拟合过程中将无法拟合的数据作为发现新知识的标志，通过将这些数据结合基本的先验知识转换为新知识并完成场景拟合来进行学习。这种用已知来解释和学习未知的增量式学习方法符合人类的认知模式，最终将实现完全的无监督学习。但这个过程需要逐步实现，以人工为主的方式来构建基本的先验知识结构和算法仍是当前的基础工作。

  描述复杂的世界需要足够丰富的结构，但也要尽可能对结构进行优化。在 DC Net 基础上提出的认知网络最优化原理可以视为是信息论向开放的认知领域的应用和扩展，今后的一个重要的课题是：将任务目标、误差计算、概率计算、结构大小等因素结合起来形成一个整体评估体系，对概念划分、结构构建、结构优化等基本认知方法进行量化。

  世界是一个整体系统，事物之间存在着全面的相互影响关系。函数式计算抽取部分关系定义为固化系统对部分变量进行计算，是对世界的部分模拟；场景拟合计算根据需要选取或

创造更有效的关系来动态构建系统并对整体系统进行计算，是对世界更全面的模拟，这是人类智能采用的方法，也应该是强人工智能需要遵循的原则，是从传统的工具计算进化到未来的智能计算的必由之路。

**参考文献：**

[1] Peter W. Battaglia, Jessica B. Hamrick, Victor Bapst, Alvaro Sanchez-Gonzalez, Vinicius Zambaldi, Mateusz Malinowski, Andrea Tacchetti, David Raposo, Adam Santoro, Ryan Faulkner, Caglar Gulcehre, Francis Song, Andrew Ballard, Justin Gilmer, George Dahl, Ashish Vaswani, Kelsey Allen, Charles Nash, Victoria Langston, Chris Dyer, Nicolas Heess, Daan Wierstra, Pushmeet Kohli, Matt Botvinick, Oriol Vinyals, Yujia Li, Razvan Pascanu. Relational inductive biases, deep learning, and graph networks. [2018]. https://arxiv.org/pdf/1806.01261.pdf

[2] Gary Marcus. The Next Decade in AI: Four Steps Towards Robust Artificial Intelligence. [2020]. https://arxiv.org/abs/2002.06177

[3] Judea Pearl, Dana Mackenzie. The Book of Why. Allen Lane,2018.

[4] Gary Marcus. The Next Decade in AI: Four Steps Towards Robust Artificial Intelligence. [2020]. https://arxiv.org/abs/2002.06177

[5] Gary Marcus.bengio v marcus and the past present and future of neural network models of language. [2018]. https://medium.com/@GaryMarcus/bengio-v-marcus-and-the-past-present-and-future-of-neural-network-models-of-language-b4f795ff352b. https://arxiv.org/ftp/arxiv/papers/1801/1801.00631.pdf

[6] Geoffrey E. Hinton,Alex Krizhevsky,Sida D. Transforming Auto-Encoders. Artificial Neural Networks and Machine Learning,ICANN 2011,21st : 44-51.

[7] WangSara Sabour, Nicholas Frosst, Geoffrey E Hinton. Dynamic Routing Between Capsules. [2017]. https://arxiv.org/abs/1710.09829

[8] Gary Marcus, Ernest Davis. Rebooting AI: Building Artificial Intelligence We Can Trust. Pantheon,2019.

[9] Nicola Kuczewski, Cristophe Porcher, Volkmar Lessmann, Igor Medina,Jean-Luc Gaiarsa. Back-propagating action potential. Communicative & Integrative Biology, 2008, 1:2: 153-155.